\documentclass[11pt]{article}
\usepackage[utf8]{inputenc}
\usepackage{CJKutf8}

\usepackage{microtype}
\usepackage{graphicx}
\usepackage{adjustbox}
\usepackage{subcaption}
\usepackage{booktabs}
\usepackage{microsoft-tech-report}
\usepackage{hyperref}
\usepackage{amsmath}
\usepackage{amssymb}
\usepackage{mathtools}

\usepackage{amsthm}
\usepackage{bm}
\usepackage{booktabs}
\usepackage{float}
\usepackage{multirow}
\usepackage{enumitem}
\usepackage[capitalize,noabbrev]{cleveref}
\usepackage{xspace}
\usepackage{titletoc}
\usepackage{placeins}  
\usepackage{tcolorbox} 
\usepackage{CJKutf8}
\tcbuselibrary{breakable,skins}
\usepackage[ruled,linesnumbered,noend,noline]{algorithm2e} 
\DontPrintSemicolon
\usepackage{listings}  
\usepackage{hyperref}
\usepackage{marvosym}
\usepackage{adjustbox}
\usepackage{xcolor}
\usepackage{colortbl}

\definecolor{bestblue}{HTML}{B3CDE3}  
\definecolor{secondblue}{HTML}{E0ECF4}

\newcommand{\best}[1]{\cellcolor{bestblue}\textbf{#1}}
\newcommand{\second}[1]{\cellcolor{secondblue}#1}
\tcbuselibrary{theorems,skins,breakable,listings}
\definecolor{promptbg}{HTML}{F8F8F8}
\definecolor{promptborder}{HTML}{1F4E79}
\newtcolorbox{promptbox}[1][]{
  colback=promptbg, colframe=promptborder, fonttitle=\bfseries\sffamily,
  breakable, enhanced, arc=3pt, boxrule=0.8pt,
  left=10pt, right=10pt, top=8pt, bottom=8pt,
  title={#1}
}

\lstdefinestyle{json}{
  basicstyle=\small\ttfamily,
  breaklines=true,
  breakatwhitespace=false,
  showstringspaces=false,
  columns=fullflexible,
  literate={"}{\textquotedbl}1,
}

\newif\ifappendixtoc
\appendixtoctrue   

\definecolor{color_blue}{HTML}{E7EFFA}
\definecolor{color_green}{HTML}{E6F8E0}
\definecolor{color_gray}{HTML}{ECECEC}
\definecolor{pearDark}{HTML}{2980B9}
\definecolor{theoremblue}{HTML}{EBF5FB}
\definecolor{theoremborder}{HTML}{2980B9}
\definecolor{propgreen}{HTML}{EAFAF1}
\definecolor{propborder}{HTML}{27AE60}
\definecolor{defyellow}{HTML}{FEF9E7}
\definecolor{defborder}{HTML}{F39C12}
\definecolor{remarkgray}{HTML}{F2F3F4}
\definecolor{remarkborder}{HTML}{7F8C8D}

\hypersetup{
  colorlinks=true,
  linkcolor=msftblue,
  citecolor=msftblue,
  urlcolor=msftblue,
  pdftitle={Mage-VL: A Codec-Native Streaming Multimodal Foundation Model}
}

\newtcbtheorem[auto counter]{theorem}{Theorem}{
  colback=theoremblue, colframe=theoremborder, fonttitle=\bfseries,
  breakable, enhanced, sharp corners, boxrule=0.6pt, left=6pt, right=6pt, top=6pt, bottom=6pt,
  crefname={Theorem}{Theorems}
}{thm}

\newtcbtheorem[use counter from=theorem]{proposition}{Proposition}{
  colback=propgreen, colframe=propborder, fonttitle=\bfseries,
  breakable, enhanced, sharp corners, boxrule=0.6pt, left=6pt, right=6pt, top=6pt, bottom=6pt,
  crefname={Proposition}{Propositions}
}{prop}

\newtcbtheorem[use counter from=theorem]{lemma}{Lemma}{
  colback=theoremblue, colframe=theoremborder, fonttitle=\bfseries,
  breakable, enhanced, sharp corners, boxrule=0.6pt, left=6pt, right=6pt, top=6pt, bottom=6pt,
  crefname={Lemma}{Lemmas}
}{lem}

\newtcbtheorem[use counter from=theorem]{corollary}{Corollary}{
  colback=propgreen, colframe=propborder, fonttitle=\bfseries,
  breakable, enhanced, sharp corners, boxrule=0.6pt, left=6pt, right=6pt, top=6pt, bottom=6pt,
  crefname={Corollary}{Corollaries}
}{cor}

\newtcbtheorem[use counter from=theorem]{definition}{Definition}{
  colback=defyellow, colframe=defborder, fonttitle=\bfseries,
  breakable, enhanced, sharp corners, boxrule=0.6pt, left=6pt, right=6pt, top=6pt, bottom=6pt,
  crefname={Definition}{Definitions}
}{def}

\newtcbtheorem[use counter from=theorem]{assumption}{Assumption}{
  colback=defyellow, colframe=defborder, fonttitle=\bfseries,
  breakable, enhanced, sharp corners, boxrule=0.6pt, left=6pt, right=6pt, top=6pt, bottom=6pt,
  crefname={Assumption}{Assumptions}
}{asm}

\newtcbtheorem[use counter from=theorem]{remark}{Remark}{
  colback=remarkgray, colframe=remarkborder, fonttitle=\bfseries,
  breakable, enhanced, sharp corners, boxrule=0.6pt, left=6pt, right=6pt, top=6pt, bottom=6pt,
  crefname={Remark}{Remarks}
}{rem}

\crefname{tcb@cnt@theorem}{Theorem}{Theorems}
\crefname{tcb@cnt@proposition}{Proposition}{Propositions}
\crefname{tcb@cnt@lemma}{Lemma}{Lemmas}
\crefname{tcb@cnt@corollary}{Corollary}{Corollaries}
\crefname{tcb@cnt@definition}{Definition}{Definitions}
\crefname{tcb@cnt@assumption}{Assumption}{Assumptions}
\crefname{tcb@cnt@remark}{Remark}{Remarks}
\crefname{algocf}{Algorithm}{Algorithms}

\newcommand{\magevl}{Mage-VL\xspace}

\newcommand{\magevit}{Mage-ViT\xspace}
\techreportshorttitle{Mage-VL}

\begin{document}
\thispagestyle{empty}

\noindent
\begin{minipage}[c]{0.5\linewidth}
\raggedright
\raisebox{-0.5\height}{\msftbrandmark}
\end{minipage}
\begin{minipage}[c]{0.49\linewidth}
\raggedleft
{\msftdatefont\small\color{msftgray}July, 2026}    
\end{minipage}\par
\vspace{0.35em}
\noindent{\color{msftline}\rule{\linewidth}{0.8pt}\par}

\vspace{1.0em}
\begin{center}
{{\msfttitlefont\fontsize{21}{25}\selectfont\color{msftdark}
Mage-VL: An Efficient Codec-Native Streaming Multimodal Foundation Model\par}}
\vspace{1.25em}

{\normalsize\rmfamily\color{msftdark}
\hyperref[sec:contributor]{Microsoft Mage Team}\par

}

\end{center}

\vspace{0.45em}
\begin{msfttitlebox}
\setlength{\parindent}{0cm}
\setlength{\parskip}{0cm}

\begin{abstract}
Standard Vision-Language Models (VLMs) suffer from Moravec's paradox: they excel at complex offline visual reasoning, but fail and suffer computational inefficiency on simple streaming perception tasks. We present Mage-VL, an efficient codec-native streaming foundation model for real-time multimodal understanding and interaction. At its core, our custom tokenizer, Mage-ViT, replaces uniform frame sampling by selectively encoding dynamic, entropy-rich regions using motion vectors and residual energy across sparse anchor (I) and predicted (P) frames. Operating at a $16\times16$ patch level, this reduces visual token consumption by over 75\% while preserving spatio-temporal context. Trained from scratch on approximately 560M unlabeled images and 100M unlabeled video frames, Mage-ViT matches or outperforms flagship encoders trained on billions of image-text pairs. We establish AI4AI data pipelines, encompassing prompt-code joint optimization for multimodal captioning and AI-driven performance diagnosis to guide training recipes. Furthermore, through a bio-inspired dual-system architecture—a lightweight System 1 event gate and a causal System 2 decoder—Mage-VL enables proactive streaming perception. Extensive evaluations show that Mage-VL-4B matches Qwen3-VL-4B on static tasks while achieving strong gains in video understanding and 2D/3D spatial reasoning with up to 3.5$\times$ wall-clock inference speedup, comprehensively surpassing the 15B Phi-4-reasoning-vision baseline. Beyond model artifacts, we deliver seven key empirical findings covering pre-training data efficiency, variable-resolution scaling, codec system acceleration, VideoQA SFT redundancy, motion-spatial synergy, AI4AI data pipelines, and Zero-Vision SFT for multimodal RL.
\end{abstract}

\vspace{0.14cm}
{\setlength{\parskip}{0.06cm}\small
{\msftmetalabel{Project Page}\href{https://microsoft.github.io/Mage}{https://microsoft.github.io/Mage}\par}
{\msftmetalabel{Code}\href{https://github.com/microsoft/Mage}{https://github.com/microsoft/Mage}\par}
{\msftmetalabel{Model}\href{https://huggingface.co/collections/microsoft/mage}{https://huggingface.co/collections/microsoft/mage}\par}
{\msftmetalabel{Date}July, 2026\par}                
}

\end{msfttitlebox}
\suppressfloats[t]  



\section{Introduction}
\label{sec:introduction}

\begin{figure}[t]
  \centering
  \includegraphics[width=\linewidth]{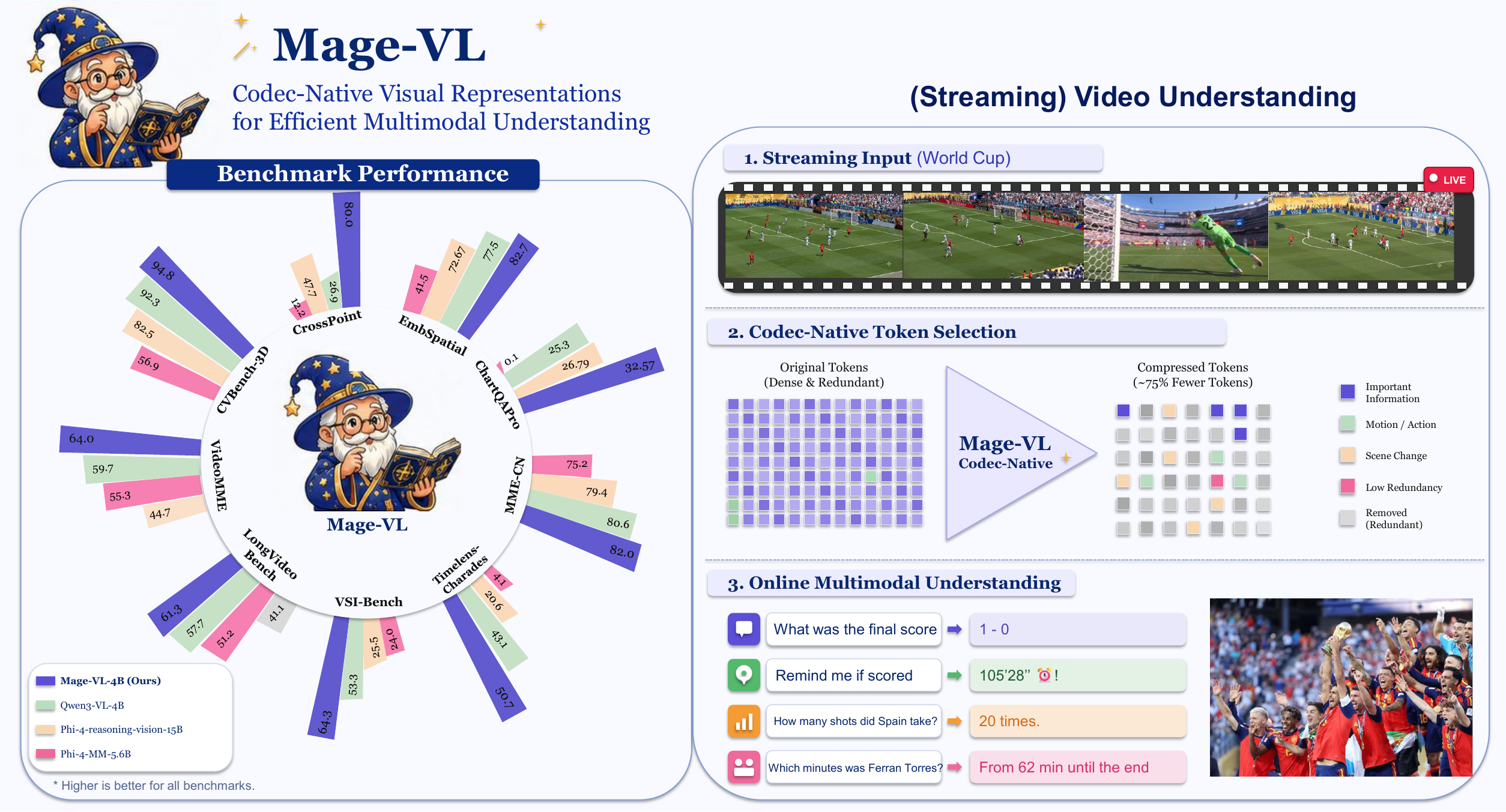}
  \caption{\textbf{Overview of Mage-VL.} \textbf{(Left)} Performance comparison of Mage-VL-4B against Qwen3-VL-4B, Phi-4-MM (5.6B), and Phi-4-reasoning-vision (15B) across spatial, video, and visual understanding benchmarks. \textbf{(Right)} Codec-native video processing workflow: given continuous streaming video inputs, Mage-VL uses motion and residual signals to select dynamic visual patches, reducing visual tokens by approximately 75\% while supporting real-time online video question answering.}
  \label{fig:teaser}
\end{figure}

Recent vision-language models (VLMs) have substantially advanced multimodal intelligence~\cite{anthropic2026claude5,deepmind2025gemini3pro,openai2026gpt55}. These models can decipher complex code~\cite{anthropic2026claude5}, read dense documents~\cite{deepmind2025gemini3pro}, and solve geometry problems~\cite{openai2026gpt55} with high proficiency. However, extending these capabilities to continuous video streams exposes a modern manifestation of \textbf{Moravec's paradox}~\cite{moravec1988mind}. Existing open-source VLMs—such as Qwen-VL~\cite{qwen3vl,bai2025qwen25vl}, InternVL~\cite{internvl3}, LLaVA-OneVision~\cite{li2024llavaonevision15}, Keye-VL~\cite{Keye-VL-1.5}, and Cambrian~\cite{cambrian}—perform well on complex reasoning over sparsely sampled keyframes, yet remain slow and computationally heavy when responding to dynamic visual events.

This paradox stems from a structural mismatch between continuous visual input and current model architectures. Physical perception is continuous and event-driven, whereas mainstream VLMs rely on vision encoders pre-trained on web-scale static images. Kimi-VL~\cite{kimi2025vl}, for instance, initializes its vision encoder from SigLIP2~\cite{tschannen2025siglip}, which is trained on over 10 billion image--text pairs. When processing videos, these models typically use uniform frame sampling at fixed resolutions. This strategy ignores spatio-temporal redundancy: static background regions are repeatedly encoded even when only a small portion of the scene changes between frames. As a result, computation scales rapidly with video duration.

Biological visual processing suggests a more efficient approach. Natural vision operates under strict resource constraints by filtering out redundant background signals at the sensory front-end and prioritizing dynamic motion~\cite{sterling2015principles,niven2008energy,gollisch2010eye}. Higher-level perception further relies on a dual-system process: a fast, low-latency pathway that determines when to react~\cite{goodale1992separate}, combined with a deliberate reasoning process for complex tasks~\cite{kahneman2011thinking}.

Inspired by this efficient dual-process design, we present \textbf{Mage-VL}, a codec-native streaming foundation model. Instead of uniform frame sampling, Mage-VL models video streams by adapting principles from classical video codecs. Built upon the OneVision Encoder architecture~\cite{ovencoder2026}, it uses temporal motion vectors and residual energy to selectively encode dynamic regions rich in visual information. Incoming video is divided into anchor frames (I-frames), which are fully processed, and predicted frames (P-frames), which retain only the patches affected by motion. Operating at a $16 \times 16$ patch level, this selective encoding significantly cuts visual token count while preserving temporal context.

Unlike standard VLMs that inherit vision encoders pre-trained on billions of static images, we train \textbf{Mage-ViT} from scratch using approximately 560 million unlabeled images and 100 million unlabeled video frames. Despite this small training volume, Mage-ViT performs on par with, and on several image and video benchmarks better than, established encoders trained on billions of image--text pairs (such as SigLIP2~\cite{tschannen2025siglip} and MoonViT~\cite{kimi2025vl}). This finding demonstrates that alignment with temporal video structures and data efficiency can matter more than sheer pre-training dataset scale. Furthermore, while Mage-ViT is trained using H.265, it exhibits strong robustness across codec families: when paired with neural codecs such as DCVC~\cite{li2021dcvc} at inference time, it maintains comparable performance while further reducing visual token overhead.

Building on Mage-ViT, we pre-train and fine-tune the Mage-VL foundation model through a progressive five-stage curriculum. While the earlier four stages inherit structural archetypes from LLaVA-OneVision2~\cite{llavaonevision2}, they differ fundamentally in data composition and strategic objectives. In \textbf{Stage 1}, we warm up the model with dense image captions and short-video captions using roughly four times the captioning data of LLaVA-OneVision2, endowing the language model with broad visual understanding capabilities. In \textbf{Stage 2}, we introduce image instruction data, temporal grounding tasks, spatial supervision, and additional video captions to establish instruction following and grounding capabilities. In \textbf{Stage 3}, we train on long-video captions to extend the model's perceptual context window. This teaches the model about event ordering and long-range dependencies. In \textbf{Stage 4}, we switch video supervision to codec video streams. Benefiting from our bio-inspired predictive patch mechanism, codec tokenization consumes approximately $1/8$ or fewer visual tokens compared to dense frame sampling, allowing us to train on video sequences that are eight times longer than those in Stage 3. In \textbf{Stage 5}, we perform streaming training: we convert the earlier video captions into a streaming format and train a proactive gate on top of the Mage-ViT to serve as System 1. This gate decides when the model should respond to incoming streaming inputs. Once triggered, the full model acts as System 2 to perform comprehensive reasoning and generate the response. The resulting Mage-VL supports continuous perception, event-triggered interaction, and conventional user-initiated question answering within a single model. Notably, Mage-VL achieves strong performance within the 4B parameter class, particularly on video understanding and spatial reasoning. It also substantially outperforms larger Phi-family baselines, including Phi-4-multimodal-instruct (5.6B) \cite{abdin2024phi4} and Phi-4-reasoning-vision (15B) \cite{aneja2026phi4reasoningvision15btechnicalreport}, across most evaluated dimensions, despite using considerably fewer parameters.


Beyond the model itself, this work provides systematic empirical studies and practical insights into efficient multimodal modeling, summarized in seven key findings: 
First, we show that web-scale pre-training is not essential for visual encoders; a custom backbone trained on merely 560M unlabeled images and 100M video frames can achieve performance competitive with encoders trained on multi-billion-scale image-text corpora~\cite{tschannen2025siglip}. 
Second, variable-resolution pre-training enables continuous, monotonic performance scaling with visual token budgets without resolution degradation. 
Third, codec-native tokenization establishes a superior spatio-temporal accuracy-efficiency frontier, achieving up to 3.5$\times$ wall-clock inference speedups over uniform frame sampling. 
Fourth, we show that explicit long VideoQA SFT is redundant; fine-tuning solely on dense video detailed captions alongside short video SFT is fully sufficient for strong zero-shot long VideoQA capabilities. 
Fifth, dynamic video training significantly enhances static 2D/3D spatial reasoning, validating the principle that motion and spatial structure are inherently synergistic~\cite{bergen1991plenoptic}. 
Sixth, we demonstrate an AI4AI data pipeline where agentic closed-loop feedback and prompt-code co-design systematically boost caption quality and downstream benchmark performance, inspiring SkillOpt-Lite~\cite{shen2026skillopt}. 
Seventh, we reveal that bypassing visual SFT in favor of pure-text reasoning SFT unlocks potent multimodal RL capabilities, offering a compute-efficient path to narrow the agentic capability gap.

Overall, our main contributions are as follows:
\begin{itemize}
    \item \textbf{Codec-Native Proactive Streaming Architecture:} We introduce Mage-VL, a unified streaming VLM that operates on continuous video via sparse codec-native updates. By encoding motion-salient patches across anchor and predicted frames, our architecture reduces visual token consumption to $1/8$ or less compared to dense frame sampling. Building on this dynamic representation, a lightweight System 1 gate monitors incoming visual streams to trigger a causal System 2 LLM decoder for low-latency, event-driven interaction.

    \item \textbf{Highly Efficient Foundation Models \& Open Release:} We present and open-source Mage-ViT and Mage-VL-4B. Notably, Mage-ViT is pre-trained from scratch on approximately 560M unlabeled images and 100M unlabeled video frames, yet achieves competitive performance compared to vision encoders pre-trained on billions of image-text pairs (e.g., SigLIP2). Driven by this efficient vision foundation, Mage-VL-4B achieves up to 3.5$\times$ wall-clock inference speedups, matches Qwen3-VL-4B on static tasks while outperforming it on video and spatial reasoning.

    \item \textbf{AI4AI and Empirical Insights:} We provide seven foundational empirical insights covering pre-training data efficiency, resolution scaling laws, codec-driven system acceleration, VideoQA SFT redundancy via dense captions, motion-spatial synergy, AI4AI closed-loop data pipeline optimization, and Zero-Vision SFT for multimodal RL. These findings challenge conventional multimodal scaling practices and provide practical recipes for efficient foundation model development.
\end{itemize}

\section{Related Work}
\label{sec:related}

\subsection{Visual Tokenization: From Image ViTs to Codec-ViT}
\label{sec:related_tokenization}

Most VLM visual encoders inherit the dense patch-grid representation introduced by Vision Transformers~\citep{dosovitskiy2021vit}, including CLIP~\citep{radford2021clip}, SigLIP and SigLIP2~\citep{zhai2023siglip,tschannen2025siglip}, DINOv2~\citep{oquab2024dinov2}, and the visual encoders of recent Qwen-VL models~\citep{bai2025qwen25vl,qwen3vl}. To reduce redundant tokens, prior work has explored adaptive pruning and merging, including DynamicViT~\citep{rao2021dynamicvit}, AdaViT~\citep{meng2022adavit}, ToMe~\citep{bolya2023tome}, FastV~\citep{chen2024fastv}, and LLaVA-PruMerge~\citep{shang2024prumerge}.

For video, TimeSformer~\citep{bertasius2021timesformer}, ViViT~\citep{arnab2021vivit}, Video Swin Transformer~\citep{liu2022videoswin}, and VideoMAE~\citep{tong2022videomae} extend visual modeling into the temporal dimension. Video VLMs further reduce token cost through temporal compression or adaptive selection, including LLaMA-VID~\citep{li2023llamavid}, Chat-UniVi~\citep{jin2024chatunivi}, SlowFast-LLaVA~\citep{xu2024slowfastllava}, MovieChat~\citep{song2024moviechat}, LongVU~\citep{shen2024longvu}, and VideoChat-Flash~\citep{videochatflash2025}. However, these methods generally compress features after dense frame encoding or select entire frames, leaving substantial spatio-temporal redundancy unresolved.

Codec-based representations provide a more direct way to exploit temporal redundancy. CoViAR~\citep{wu2018coviar} demonstrated compressed-domain video recognition using motion vectors and residuals, while Video-LaVIT~\citep{videolavit2024} introduced codec-derived motion information into multimodal modeling. OneVision Encoder~\citep{ovencoder2026} further develops codec-aligned patch sparsity, and LLaVA-OneVision-2~\citep{llavaonevision2} incorporates video-native representations into VLM training.

\magevit builds on this direction by representing videos with anchor frames and sparse predicted frames, allocating tokens only to temporally informative regions before expensive visual encoding. Unlike image-centric encoders, \magevit is trained from scratch on both image and video data, while its codec-native design supports efficient long-video processing and incremental streaming.

\subsection{Open Vision--Language Models}
\label{sec:related_vlm}

Modern VLMs typically connect a pretrained vision encoder to a decoder-only LLM through a lightweight projector. Early works such as Flamingo~\citep{alayrac2022flamingo} and BLIP-2~\citep{li2023blip2} explored efficient vision--language alignment, while LLaVA~\citep{liu2023llava}, MiniGPT-4~\citep{zhu2023minigpt4}, InstructBLIP~\citep{dai2023instructblip}, and LLaVA-1.5~\citep{liu2024llava15} established visual instruction tuning as a standard paradigm.

Recent open VLMs scale this recipe with stronger visual encoders, larger multimodal corpora, and multi-stage training. Representative models include Qwen-VL~\citep{bai2023qwenvl,bai2025qwen25vl,qwen3vl}, InternVL~\citep{internvl3,chen2024nnvl}, DeepSeek-VL~\citep{lu2024deepseekvl}, Pixtral~\citep{agrawal2024pixtral}, Molmo~\citep{deitke2024molmo}, Kimi-VL~\citep{team2026kimi}, LLaVA-OneVision~\citep{li2024llavaonevision15,llavaonevision2}, and Keye-VL~\citep{Keye-VL-1.5}. Compact models such as Phi-3.5-Vision~\citep{abdin2024phi3} and Phi-4-Multimodal~\citep{abdin2024phi4} provide strong smaller-scale baselines, while BAGEL~\citep{deng2025bagel}, Emu3.5~\citep{cui2025emu35nativemultimodalmodels}, Ovis-U1~\cite{wang2025ovis} further unify visual understanding and generation~\cite{zhang2025unified}.

Video VLMs largely extend this image-centric paradigm by aggregating features from sampled frames. VideoChat~\citep{li2023videochat}, Video-ChatGPT~\citep{maaz2023videochatgpt}, Video-LLaMA~\citep{zhang2023videollama}, and Video-LLaVA~\citep{lin2023videollava} introduced early video extensions, while LLaVA-OneVision~\citep{li2024llavaonevision} and LongVA~\citep{zhang2024longva} improve unified image--video modeling and long-context understanding.

Despite these advances, most video VLMs still rely on sampled frames followed by dense per-frame tokenization. \magevl instead trains a video-native visual encoder from scratch and directly consumes sparse codec-aligned visual streams, reducing temporal redundancy before it reaches the language model.

\subsection{Streaming and Online Video-Language Models}
\label{sec:related_streaming}

Most video VLMs operate offline, processing sampled frames as a static multi-image input before generating a response. Recent work instead explores continuous video interaction and proactive response timing. VideoLLM-online~\citep{chen2024videollmonline} interleaves streaming perception and language generation, MMDuet~\citep{wang2024mmduet} supports responses at arbitrary moments, and Dispider~\citep{qian2025dispider} decouples perception, decision, and reaction. StreamMind~\citep{ding2025streammind} further introduces event-gated response triggering, while JoyAI~\citep{joyai2026vlinteraction} enables real-time interaction through a multi-agent pipeline.

Another line of work focuses on memory and computational efficiency for long-running streams. Flash-VStream~\citep{zhang2024flashvstream}, StreamingVLM~\citep{xu2026streamingvlmrealtimeunderstandinginfinite}, and InternLM-XComposer2.5-OmniLive~\citep{zhang2024ixc25omnilive} maintain compressed or rolling visual states for long-term streaming understanding. Streaming supervision and evaluation have also been studied using temporally aligned data, such as LiveCC~\citep{chen2025livecclearningvideollm} and MatchTime~\citep{rao2024matchtimeautomaticsoccergame}, together with benchmarks including StreamingBench~\citep{lin2024streamingbench} and OVO-Bench~\citep{li2025ovobench}.

Unlike prior methods that mainly add streaming mechanisms to conventional frame-based encoders, \magevl makes streaming native to the visual front-end. It processes rolling codec-token windows and integrates proactive response triggering with language generation through a dual-system architecture.

\section{Mage-ViT}
\label{sec:magevit}

\begin{figure*}[t]
    \centering
    \includegraphics[width=\textwidth]{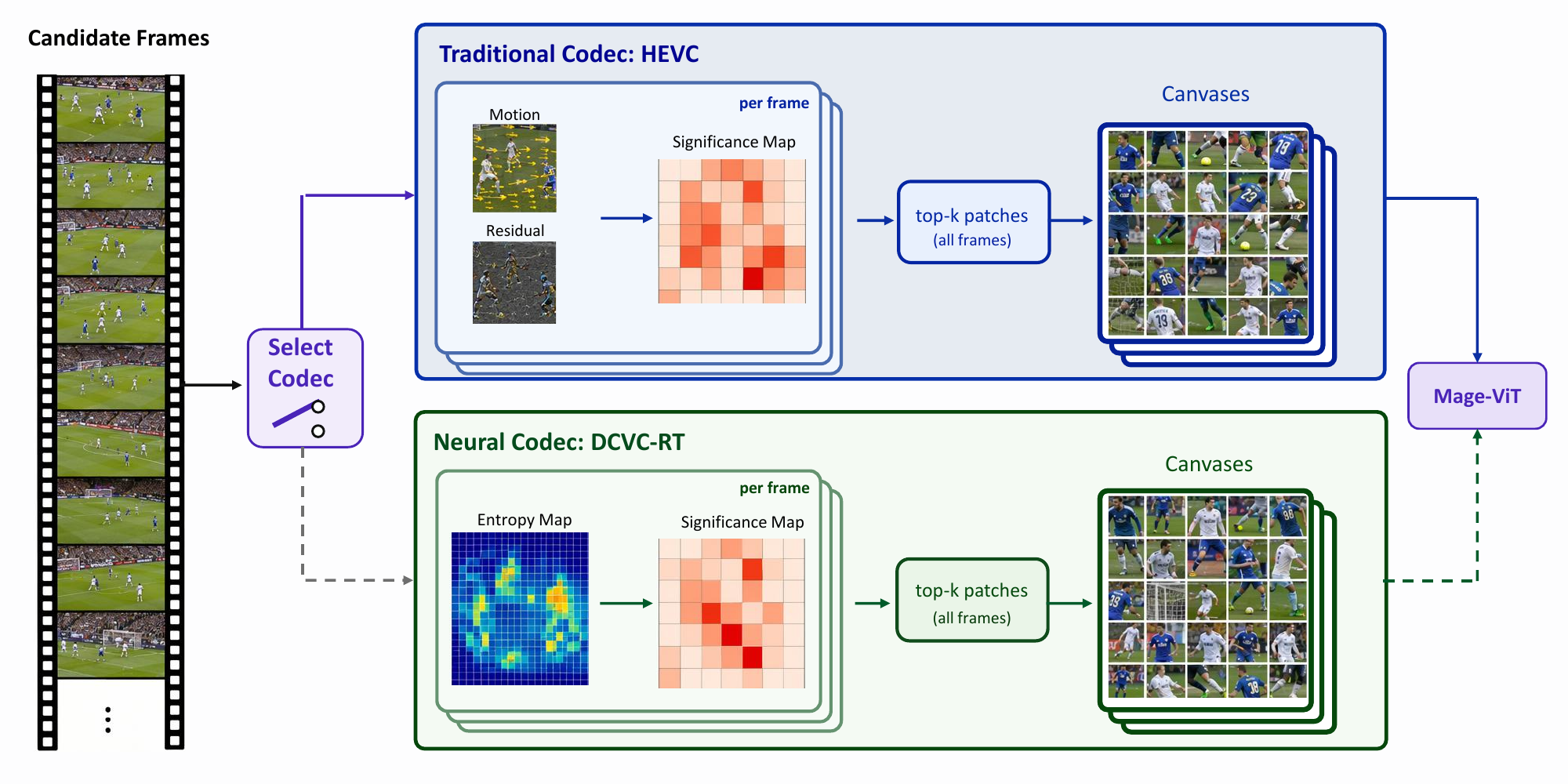}
    \caption{
    \textbf{Codec-driven patchifier of \magevit.}
    \magevit robustly supports both traditional codec HEVC/H.265 \cite{sullivan2012hevc} and neural codec DCVC-RT \cite{jia2025towards} to select patches from video frames according to information density. The selected patches are composed into canvas, which is fed to \magevit with corresponding spatial-temporal positional encoding information.
    }
    \label{fig:magevit_framework}
\end{figure*}

This section introduces \magevit, the visual encoder at the core of \magevl. \magevit is designed primarily for video, while single images are treated as a degenerate one-frame case. It follows the \emph{codec-aligned sparsity} principle of OneVision-Encoder~\citep{ovencoder2026}: visual tokens should be allocated according to where the codec spends bits, since codec bit allocation provides a natural proxy for spatio-temporal importance. \cref{sec:magevit_arch} describes the architecture, and \cref{sec:magevit_training} describes the pre-training data, objective and the two-stage recipe.

\subsection{Architecture}
\label{sec:magevit_arch}

\magevit adopts a Codec-ViT architecture consisting of a codec-driven
patchifier, a Vision Transformer trunk, and a shared 3D rotary
positional encoding.

\paragraph{Codec-driven patchifier.} \magevit first divides the input into $16{\times}16$ pixel patches and then performs codec-guided sparse selection. 
For multi-frame inputs, a per-patch importance tensor $S \in \mathbb{R}^{T\times H\times W}_{\geq 0}$ is estimated to represent how many bits or how much information the codec assigns to that spatio-temporal patch. We adopt the \emph{traditional codecs} HEVC/H.265~\citep{sullivan2012hevc} as the default codec in patch selection, where $S$ is defined as a weighted combination of the magnitude of motion vectors and the residual energy of P-frames. We note that our ViT can robustly support more diverse codecs such as the \emph{neural codecs} DCVC-RT~\citep{jia2025towards}. In this case, $S$ is obtained directly from the codec's learned probability model, where the negative log-likelihood of each patch corresponds to the estimated number of bits required for coding. We observe that the resulting bit-allocation map provides a reliable proxy for local motion and temporal variation. The codec-driven patchifier is illustrated in Fig.~\ref{fig:magevit_framework}.

Based on $S$, the patchifier keeps all I-frame patches, and selects the top-$k$ P-frame patches by importance $S$ within a token budget $B$. For a $64$-frame clip with $16\times16$ tokens in each frame, we select $B=4096$ tokens, corresponding to roughly $75\%$ token reduction. We additionally support two other video input modes: \textit{Chunk-wise patchification,} which follows conventional frame-centric paradigm to partition the input into $C$ temporal chunks and samples one frame per chunk; \textit{Collage patchification,} where one frame is first sampled from each temporal chunk and the sampled frames are vertically concatenated into a single tall 2D canvas.

\paragraph{ViT trunk.} \magevit uses a $24$-layer pre-norm Vision Transformer with hidden dimension $1024$, $16$ attention heads, and GELU MLPs at $4{\times}$ expansion ratio, implemented with Flash Attention 2 \cite{dao2023flashattention2} for both training and inference. The trunk processes variable-length visual token sequences produced by codec-driven sparse selection. Its outputs are passed through the multimodal projector and subsequently consumed by the language model in \magevl (\cref{sec:magevl}). During batched training, padding tokens are masked so that attention respects the valid sequence boundary of each sample.

\paragraph{Shared Positional encoding.} Position information for the ViT trunk is provided by a shared 3D rotary positional encoding~\citep{su2024rope} over the un-pruned spatio-temporal grid.  In this case, the spatial-temporal relations across patches are retained even when large fractions of the grid are dropped by the codec front-end.

\subsection{ViT Pre-training}
\label{sec:magevit_training}

\magevit is trained from scratch using a two-stage pre-training pipeline.

\paragraph{Data.} We filter the image and video dataset following the filter rules in OneVision Encoder~\citep{ovencoder2026}. For images, we select source images from LAION-400M~\citep{schuhmann2021laion}, COYO-700M~\citep{byeon2022coyo}, OBELICS~\citep{laurenccon2023obelics}, Zero250M~\citep{xie2023ccmb}, and ImageNet-21K~\citep{deng2009imagenet}. For videos, we use videos from HowTo100M~\citep{miech2019howto100m} and Panda-70M~\citep{chen2024panda70m}. It contains a diverse set of publicly available images and videos, spanning natural photographs, artworks, documents, charts, indoor scenes, web videos, robotic trajectories, instructional content, and high-motion clips.

\paragraph{Objective.} We train \magevit with a large-scale \emph{cluster-discrimination} objective over the visual representation space. The objective encourages semantically related samples to align with shared visual concept prototypes. Specifically, we extract MetaCLIP features from both image and short-video data, perform K-means clustering over the combined corpus, and assign each training sample to its nearest cluster centers. \magevit is then optimized using negative-sampling cluster discrimination against these visual prototypes. This objective encourages semantically related content, such as text regions, human faces, or traffic signs, to occupy nearby regions in the learned representation space.

\paragraph{Stage 1: Variable-resolution image pre-training.} 
We first pre-train \magevit on image data with resolutions ranging from $224$ to $448$ and aspect ratios from $1{:}2$ to $2{:}1$. Images of the same resolution are packed together to better utilize the target token budget, which we find improves optimization efficiency. This stage uses single-image spatial patchification. Since each sample contains only one frame, the temporal prediction pathway remains inactive. We use AdamW with an initial learning rate of $10^{-3}$, cosine decay with warm-up, gradient clipping at $1.0$, bf16 mixed precision, and a large effective batch size obtained through gradient accumulation. The negative-sampling ratio for cluster discrimination is set to $r=0.1$.

\paragraph{Stage 2: Joint image and video pre-training.} 
Starting from the Stage-1 checkpoint, we jointly train \magevit on image and video data. Image resolution varies from $224$ to $448$, while videos are trained at resolution $256$ with 64 frames per clip and a token budget of $4096$. The three patchification modes are mixed during training with approximate proportions of $50\%$ codec, $40\%$ chunk-wise, and $10\%$ collage. The learning rate is reduced to $5\times10^{-5}$. This stage activates codec-driven temporal sparsification and jointly optimizes spatial representation learning and spatio-temporal token selection. Training is conducted in bf16 mixed precision.

\section{Mage-VL}
\label{sec:magevl}

Building upon the codec-native visual encoder introduced in \cref{sec:magevit}, we develop \magevl, a unified multimodal model for image understanding, offline video reasoning, and proactive interaction over continuous video streams. The same model supports static images, videos of varying durations, event-triggered commentary, and conventional user-initiated question answering.

\Cref{sec:magevl_arch} presents the unified architecture across image, offline-video, and streaming modes. \Cref{sec:magevl_data} describes the image, video, and streaming supervision, while \cref{sec:magevl_training} details the progressive five-stage training curriculum.

\begin{figure*}[t]
    \centering
    \includegraphics[width=\textwidth]{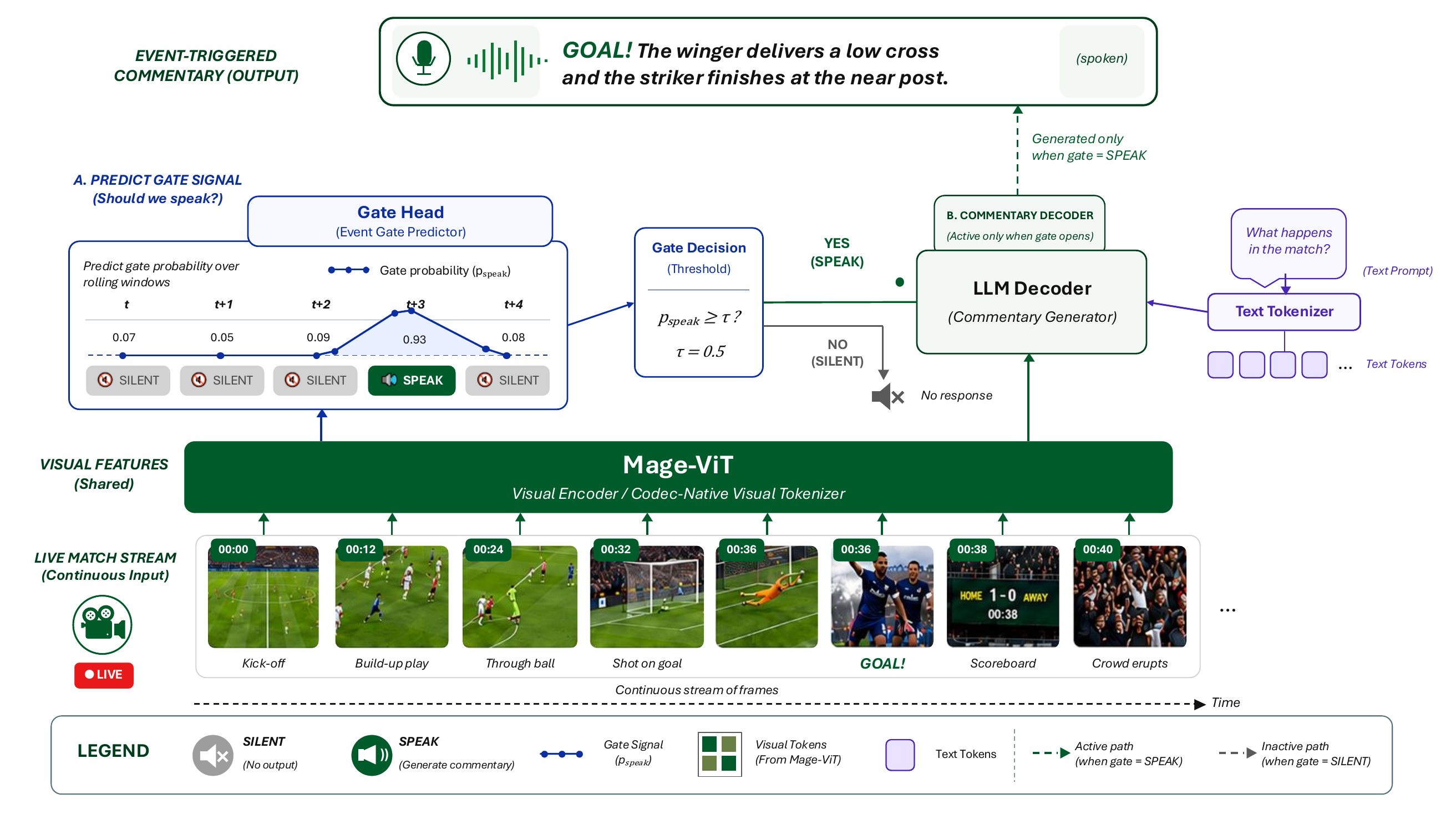}
    \caption{\textbf{Proactive streaming framework of \magevl.} \magevit incrementally encodes a continuous video stream into codec-native visual features shared by the event gate and causal language decoder. The gate evaluates each rolling visual window and predicts whether the current prefix contains a response-worthy event. The model remains silent when the gate is closed; when it opens, the decoder generates an event-conditioned response from the recent visual context and the given text prompt.}
    \label{fig:magevl_framework}
\end{figure*}

\subsection{Unified Architecture}
\label{sec:magevl_arch}

\paragraph{Codec-native visual encoder.}
\magevl uses \magevit as its shared visual encoder across image, video, and streaming inputs. A still image is represented as a single spatial token sequence. For video, \magevit produces temporally ordered codec-token windows composed of densely encoded anchor-frame patches and sparsely selected predicted-frame patches. Anchor frames preserve complete scene context, whereas predicted frames retain only patches associated with motion or visual changes. This codec-native representation avoids repeated computation over temporally static regions while preserving fine-grained spatial and temporal information. The resulting visual features are passed to the vision-language projector and causal language decoder.

\paragraph{Vision-language projector.}
A lightweight two-layer MLP maps \magevit features into the token-embedding space of the language model. Because \magevit applies shared 3D rotary positional encoding over the original spatio-temporal grid, retained tokens preserve their original spatial and temporal coordinates after codec-driven sparse selection.

\paragraph{Causal language decoder.}
The language backbone is initialized from Qwen3-4B-Instruct-2507~\cite{qwen3}. Projected visual tokens and text tokens are processed by the same causal decoder. For images, visual tokens form a single block preceding the text instruction. For videos, codec-token windows are concatenated in temporal order. This unified interface allows \magevl to process images, short videos, long videos, and ultra-long videos without modifying the language backbone or introducing modality-specific decoders.

\paragraph{Proactive streaming mechanism.}
As illustrated in Fig.~\ref{fig:magevl_framework}, a lightweight cognition gate continuously evaluates the visual features of incoming codec windows. Given the current streaming representation $\mathbf{h}_t$, the gate predicts $p_{\mathrm{speak}}=g(\mathbf{h}_t)$ and triggers generation when $p_{\mathrm{speak}}\geq\tau$, where $\tau$ is a fixed inference threshold. Background segments and non-response-worthy moments keep the gate closed, whereas relevant events activate language generation.

During streaming inference, newly arrived codec windows are incrementally processed by the perception pathway. The cognition gate evaluates the accumulated streaming memory, while response generation uses a local sliding window containing the most recent codec-token segments. A text query may be inserted at any time, whereas proactive responses are controlled by the cognition gate. This design supports continuous perception, event-triggered commentary, and user-initiated interaction within the same model.

\subsection{Data Construction}
\label{sec:magevl_data}
Our training corpus contains approximately \textbf{350M image--caption pairs}, \textbf{54M image-instruction samples}, \textbf{7.95M unique video--caption samples}, and \textbf{3.35M streaming samples}. Caption data establish broad visual-language and temporal grounding, whereas instruction data provide task-oriented reasoning, response formatting, and grounding behavior. The streaming corpus further supervises response timing over causal video prefixes. During video training, selected image-instruction samples are interleaved with video data to preserve static-image understanding and general instruction-following capabilities.

\subsubsection{Image Data}
\label{sec:image_data}

\begin{figure}[t]
    \centering
    \includegraphics[width=\textwidth]{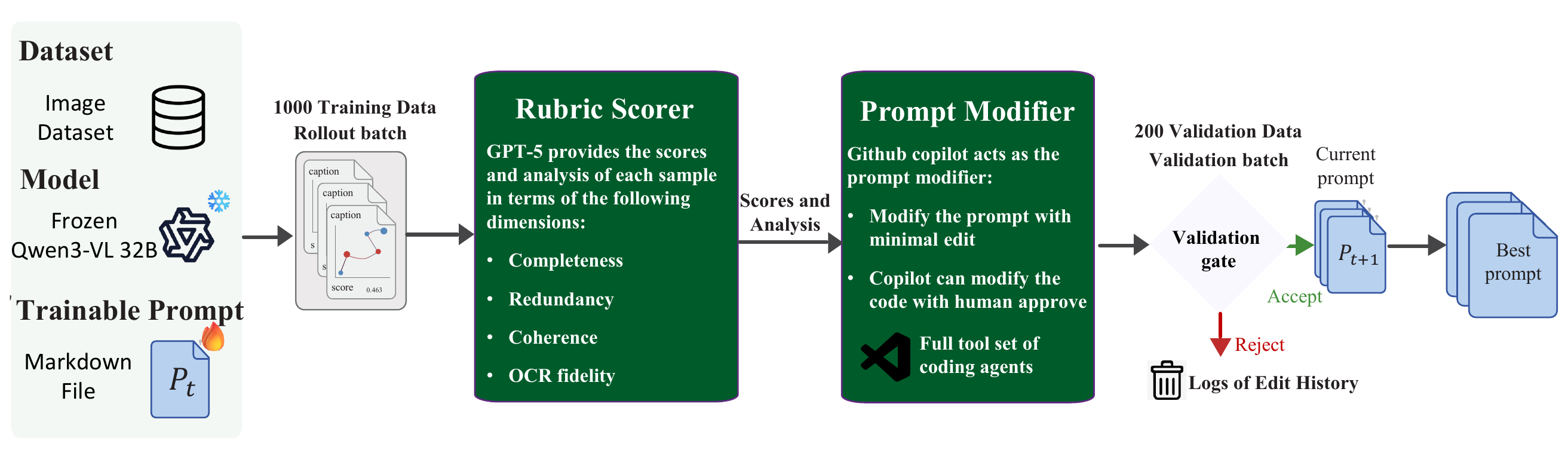} 
    \caption{\textbf{Iterative prompt-optimization pipeline for image recaptioning.} A frozen Qwen3-VL-32B generates captions for a rollout batch, which are evaluated by a GPT-5 Rubric Scorer for completeness, redundancy, coherence, and OCR fidelity. GitHub Copilot then proposes minimal prompt or code revisions, which are evaluated on a held-out validation batch and accepted only after human review.}
    \label{fig:prompt_optimization_pipeline}
\end{figure}

\paragraph{Scaled image-caption corpus.}
We construct approximately 350M image--caption pairs from two complementary sources. The first contains 85M images from the LLaVA-OneVision-1.5 mid-training corpus~\citep{li2024llavaonevision15}, while the second is obtained by filtering approximately one billion publicly available image--text pairs~\citep{zhang2026mageflow} to retain 265M high-quality images. Images from both sources are recaptioned using the optimized pipeline illustrated in Fig.~\ref{fig:prompt_optimization_pipeline}. Compared with short web alt-text, the resulting dense captions provide substantially richer descriptions of foreground and background objects, attributes, counts, spatial relationships, actions, and global scene context. For documents, charts, tables, posters, and screenshots, the captions additionally preserve rendered text, layout structure, and visual--textual relationships, with visible text retained verbatim in its original language whenever possible. Together, these data provide the primary early visual-language supervision for \magevl, covering object recognition, attributes, OCR, document understanding, chart interpretation, and spatial relationships before instruction tuning.

\paragraph{Caption-prompt optimization.}
Driven by the recent rise of the \textbf{AI4AI} paradigm—where automated systems and agentic workflows are deployed to iteratively scale and refine data quality~\cite{ai4ai_scale2026}. We proactively designed and implemented a closed-loop data optimization pipeline over six months ago during the early development of our framework.

The motivation is that the quality of recaptioned data depends strongly on the system prompt used by the captioning model. We therefore develop an iterative prompt optimization pipeline that combines multi-agent evaluation with human-in-the-loop verification, as illustrated in Fig.~\ref{fig:prompt_optimization_pipeline}. The optimized prompt ($P_{t+1}$) is subsequently used with a frozen Qwen3-VL-32B \cite{qwen3vl} to generate the large-scale dense-caption corpus.

We initialize the optimization process with a rollout batch of 1,000 training images uniformly sampled from ten representative visual domains, with 100 images from each domain. These domains cover diverse content distributions, including natural scenes, people, objects, documents, charts, rendered text, artworks, screenshots, and other visually structured content. Initial captions are generated using the baseline LLaVA-OneVision-1.5 recaptioning prompt~\cite{li2024llavaonevision15}.

For each optimization iteration, a rubric-scoring agent powered by GPT-5 evaluates the generated captions along four dimensions:
\begin{itemize}
    \item \textbf{Completeness:} whether the caption covers the major objects, attributes, actions, spatial relationships, and contextual information visible in the image;
    \item \textbf{Redundancy:} whether the caption contains repeated, verbose, or semantically duplicated descriptions;
    \item \textbf{Coherence:} whether the caption is logically organized, grammatically consistent, and easy to follow;
    \item \textbf{OCR fidelity:} whether visible text is correctly transcribed, preserved in its original language, and associated with the appropriate visual region or document structure.
\end{itemize}

Based on the scores and diagnostic analysis, a prompt-refinement agent, implemented via GitHub Copilot equipped with a full toolset of coding agents, proposes modifications to the current prompt ($P_t$). Guided by the principle of making minimal edits, Copilot refines the prompt within a Markdown file. The proposed prompt is then evaluated through a validation gate using a separate validation batch of 200 images. A human reviewer oversees this process to approve or reject the modification. Revisions that fail the validation gate are rejected and archived into the logs of edit history, while approved revisions are accepted as the new baseline prompt ($P_{t+1}$) for the next iteration. We repeat this evaluate--refine--verify cycle for ten iterations, yielding a final prompt that balances visual coverage, language conciseness, structural coherence, and OCR preservation.

The optimized system prompt is then used for large-scale image captioning. As illustrated in \cref{subsec:caption_prompt_ablation}, captions generated with the optimized prompt consistently improve performance across OCR, document, chart, perception, and general visual-reasoning benchmarks. The complete recaptioning system prompt is provided in Appendix~\ref{app:caption_prompt}.

\paragraph{Image instruction data.}
In addition to dense image captions, we collect approximately 54M image-instruction samples. The filtered mixture contains approximately 44.3M samples derived from LLaVA-OneVision-1.5~\citep{li2024llavaonevision15} and FineVision~\citep{finevision25m}, 6.0M samples from OpenBee~\citep{zhang2026beehighqualitycorpusfullstack}, and 4.6M spatial and GUI instruction samples from LLaVA-OneVision-2~\citep{llavaonevision2}.
The mixture covers general visual question answering, OCR, documents, charts, STEM and mathematical reasoning, multi-image understanding, spatial grounding, GUI understanding, and structured multimodal reasoning. The spatial subset emphasizes object size, count, relative direction, distance, depth, and appearance order, while the GUI subset covers interface-element recognition, localization, and action-oriented reasoning.
During later video stages, subsets of these samples are interleaved with video-caption and video-instruction data. This mixed training preserves image reasoning and instruction-following capabilities as the temporal context is progressively extended.

\subsubsection{Video Data}
\label{sec:video_data}
Our video-caption corpus~\citep{llavaonevision2} contains approximately \textbf{7.95M unique samples}: 4.2M videos of up to 30 seconds, 2.7M videos spanning 30--60 seconds, 0.7M videos spanning 60--180 seconds, and 350K videos of approximately 10--15 minutes. For the 350K long-video subset, we divide each video into non-overlapping segments and estimate content density using codec-stream bit counts, retaining the segment with the highest bit count as the most visually dynamic interval. This subset is initially processed through standard frame sampling and later revisited using codec-stream tokenization, but is counted only once in the unique-sample total. We additionally incorporate LLaVA-Video-178K~\citep{zhang2025llavavideovideoinstructiontuning}, TimeLens~\citep{zhang2026timelensrethinkingvideotemporal}, VideoChat-Flash-Training-Data~\citep{videochatflash2025}, Molmo2-VideoTrack, and Molmo2-VideoPoint~\citep{clark2026molmo2} for video question answering, temporal grounding, tracking, point-based localization, and spatio-temporal reasoning.

\paragraph{Duration-stratified video captions.}
The 4.2M short videos are sampled at 1 fps with at most 30 frames and primarily describe local actions, brief interactions, and object-state changes. The 30--60-second and 60--180-second subsets use up to 60 and 90 frames, respectively, providing supervision for multi-step activities, scene transitions, event ordering, and longer action sequences. The 10--15-minute subset initially uses up to 384 sampled frames and captures long-range event progression, repeated actions, scene continuity, and cross-segment dependencies.
Caption granularity is adapted to video duration. Short clips receive compact, action-dense descriptions, whereas longer videos use segment-level captions with explicit temporal progression and cross-segment references. This supervision teaches the model how visual states evolve over time rather than treating a video as an unordered collection of frames.

\paragraph{Frame-sampled and codec-stream representations.}
All 7.95M video-caption samples are introduced through conventional frame sampling during the early and intermediate training stages. For codec-native adaptation, the 350K long-video subset is converted into temporally ordered codec windows using the patchifier in Fig.~\ref{fig:magevit_framework}. Anchor-frame patches are retained densely, whereas predicted frames contribute only patches selected by codec-derived motion and residual importance. We use configurations of up to 384 and 768 frames.
This conversion preserves the original caption supervision while changing the visual observation interface. Codec-stream tokenization allocates tokens according to spatio-temporal variation rather than assigning a dense patch grid to every sampled frame, enabling denser temporal coverage under a comparable visual-token budget.

\subsubsection{Streaming Event Data}
\label{sec:streaming_event_data}

We construct streaming supervision from our 180-second caption corpus and existing timestamped datasets, including MatchTime~\citep{rao2024matchtimeautomaticsoccergame}, LiveCC~\citep{chen2025livecclearningvideollm}, and StreamingVLM~\citep{xu2026streamingvlmrealtimeunderstandinginfinite}. All sources are converted into a unified real-time \emph{speak}/\emph{silent} format.

\begin{figure*}[t]
    \centering
    \includegraphics[width=\textwidth]{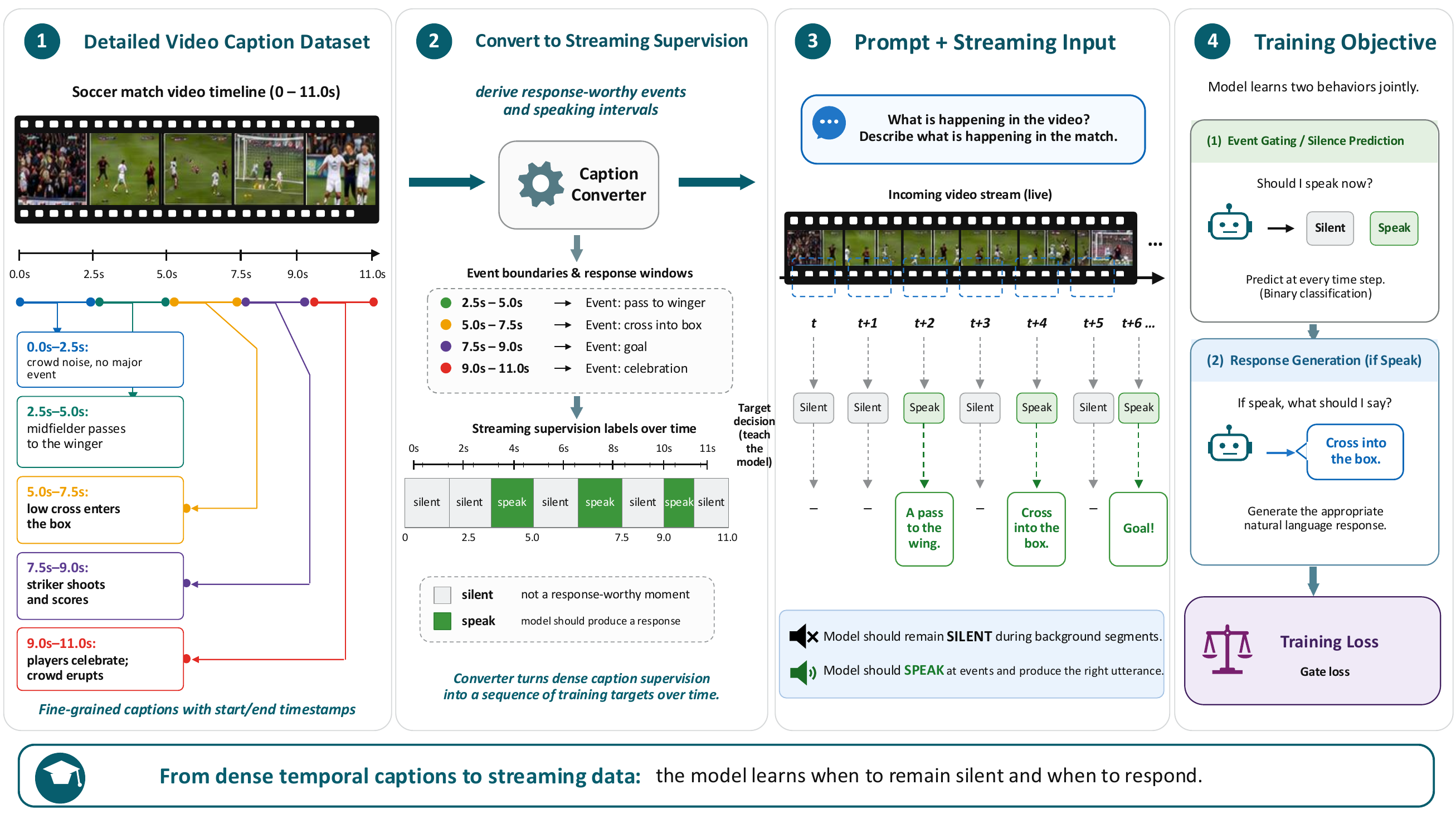}
    \caption{\textbf{Constructing proactive streaming supervision from timestamped video captions.} Timestamped captions are converted into gate and response targets over causal video prefixes. At each caption start timestamp, the target is \emph{speak} and the corresponding caption provides the response; all other candidate timestamps are labeled \emph{silent}. This supervision teaches the model when to remain silent and when to generate an event-conditioned response.}
    \label{fig:streaming_data}
\end{figure*}

Our streaming corpus contains \textbf{3.3M training samples}: 2.8M samples converted from our 180-second caption data and 0.5M samples derived from existing timestamped streaming datasets. Each sample consists of a causal video prefix, a binary \emph{speak}/\emph{silent} target, and (for positive instances) a corresponding language response. All sources are normalized into the same real-time annotation format.

As illustrated in \cref{fig:streaming_data}, an annotated video is represented as
\[
\mathcal{V}=\{(s_j,e_j,c_j)\}_{j=1}^{N},
\]
where $s_j$ and $e_j$ denote the start and end timestamps of the $j$-th segment, and $c_j$ describes the corresponding event, action, or speech. Following the event-gated formulation of StreamMind~\citep{ding2025streammindunlockingframerate}, a caption converter transforms these timestamped annotations into real-time gate and response targets.

For each caption $(s_j,e_j,c_j)$, the start timestamp $s_j$ is assigned a positive \emph{speak} label and paired with $c_j$ as the target response. Candidate timestamps that do not begin a caption receive a negative \emph{silent} label. Applying this conversion to the 180-second caption corpus produces 2,801,360 real-time training samples.

All examples satisfy a \emph{no-future-frame} constraint: the gate decision and target response depend only on observations available up to the current timestamp. Applying the same conversion to the existing timestamped datasets produces the remaining 543,943 samples and increases the diversity of events, response styles, and interaction timings.

\subsubsection{Data Curation and Balancing}
\label{sec:data_curation}
All data mixtures are filtered for corrupted files, unsafe content, low-quality samples, and near-duplicates. For caption data, we additionally remove captions that are too short to provide meaningful visual grounding, captions whose timestamps are inconsistent with the video duration, repeated or near-duplicate temporal descriptions, and samples flagged by caption--video consistency checks. For streaming supervision, we balance the \emph{speak}/\emph{silent} labels produced by the caption converter by subsampling redundant silent timestamps.

We balance the corpus along two additional axes. Domain balancing upweights regimes that are underrepresented in web data but important for evaluation and deployment, including OCR, documents, charts, STEM and mathematics, spatial reasoning, temporal grounding, and streaming events. Duration balancing controls the proportions of short, medium, long, and ultra-long videos at each training stage. Overall, the early stages inherit the open-data family of LLaVA-OneVision-style training, but place greater emphasis on scaled dense image captions, temporally precise video captions, codec-stream supervision, and calibrated cognition-gate targets.


\subsection{Progressive Training Curriculum}
\label{sec:magevl_training}

\magevl is trained through a progressive five-stage supervised curriculum. The five stages successively establish visual-language grounding, instruction following, temporal-context extension, codec-native long-context adaptation, and proactive streaming alignment. Together, they produce a single unified model that supports image understanding, offline video reasoning, and proactive interaction over continuous video streams. 

\paragraph{Stage 1: Multimodal alignment via captions.}
Stage~1 uses approximately 350M dense image captions and 4.2M short-video captions~\citep{llavaonevision2}. Dense image captions provide the dominant supervision, forcing the model to associate \magevit tokens with objects, attributes, OCR text, charts, documents, spatial relationships, and global scene context. Fine-grained short-video captions are included from the beginning so that the model encounters motion and state changes before the long-video curriculum. Unlike a projector-only alignment stage, this warm-up directly trains the model as a caption-capable multimodal language model.


\paragraph{Stage 2: Instruction tuning and short temporal grounding.}
Stage~2 uses approximately 54M image-instruction samples from LLaVA-OneVision-1.5~\citep{li2024llavaonevision15}, FineVision~\citep{finevision25m}, OpenBee~\citep{zhang2026beehighqualitycorpusfullstack}, and LLaVA-OneVision-2 spatial and GUI data~\citep{llavaonevision2}, together with 3.4M video captions spanning 30--180 seconds~\citep{llavaonevision2}. We increase the weight of multimodal instruction data, particularly OCR, document QA, chart reasoning, STEM and mathematics, multi-image reasoning, spatial grounding, and GUI understanding. Short-video captions and instructions remain active to provide local temporal grounding for actions and object-state transitions. Generic language-only or weakly visual SFT is assigned a low weight so that it does not dominate the visual grounding acquired in Stage~1. This stage transforms the model from a visual describer into a general multimodal assistant while retaining caption supervision as regularization.


\paragraph{Stage 3: Temporal-horizon expansion.}
Stage~3 retains 20M Stage-2 image-instruction samples and introduces medium- and long-video data from LLaVA-Video~\citep{zhang2025llavavideovideoinstructiontuning}, TimeLens~\citep{zhang2026timelensrethinkingvideotemporal}, VideoChat-Flash~\citep{videochatflash2025}, Molmo2~\citep{clark2026molmo2}, and our temporally structured captions. These data describe event order, scene transitions, repeated actions, and cross-segment dependencies. The goal is not simply to expose the model to more frames, but to teach it how visual states evolve over minutes and how later events depend on earlier context. Image SFT, spatial data, GUI data, and short-video samples remain in the mixture to preserve static perception and instruction-following capabilities. \textbf{We also employ AI4AI at this stage. Specifically, we employ AI-based diagnostics to pinpoint underperforming or missing capabilities, which are subsequently addressed by targeted human data supplementation. Meanwhile, these diagnostics dynamically guide the selection of the optimal resolution and frame count.}


\paragraph{Stage 4: Codec-native long-context adaptation.}
Stage~4 uses 350K long-video captions, 4M spatial samples~\citep{llavaonevision2}, Molmo2 tracking and pointing data~\citep{clark2026molmo2}, and 40M retained image-instruction samples. After the model has learned to reason over extended temporal contexts, we convert a large fraction of video training into the codec-native representation used at inference time. Captioned videos are represented as rolling token windows containing dense anchor-frame evidence and sparse predicted-frame updates. This stage adapts the language model to the irregular visual sequences produced by \magevit, where token density follows visual change rather than fixed frame slots. Long and ultra-long videos benefit most from this representation because the model can observe more event boundaries under the same visual-token budget. Non-video instruction data remain in the mixture to preserve image, OCR, chart, spatial, and GUI capabilities.


\paragraph{Stage 5: Proactive streaming alignment via cognition-gate fine-tuning.}
In Stage~5, we introduce a lightweight cognition gate to enable proactive streaming interaction while keeping the base language model completely frozen. For each causal segment, the codec-extraction pipeline packs codec-selected patches into a compact canvas while preserving their original spatio-temporal coordinates. These features are processed by the event-preserving feature extractor (EPFE), whose recurrent state maintains a streaming perception memory $\mathcal{M}_{\mathrm{per}}$ used exclusively by the cognition gate for real-time turn-taking decisions. Crucially, when the gate issues a \emph{speak} decision, we do not employ complex long-range memory for language generation. Instead, the model invokes the frozen base VLM directly using a local sliding window of the most recent $N$ codec-assembled visual segments as input context to execute standard autoregressive response generation. We train the cognition gate on approximately 3.35M streaming samples: 2.8M converted from 180-second captions and 0.54M converted from MatchTime~\citep{rao2024matchtimeautomaticsoccergame}, LiveCC~\citep{chen2025livecclearningvideollm}, and StreamingVLM~\citep{xu2026streamingvlmrealtimeunderstandinginfinite}, as described in \cref{sec:streaming_event_data}.

Throughout this stage, the visual backbone, EPFE, and language model remain strictly frozen, and we solely fine-tune the cognition gate. At every candidate timestamp, the gate autoregressively evaluates the accumulated streaming memory $\mathcal{M}_{\mathrm{per}}$ and predicts one of two tokens: \emph{silent} or \emph{speak}. Caption start timestamps provide speak targets, while all other timestamps serve as silent targets. To address the inherent sparsity of speak events, we optimize a class-weighted token cross-entropy loss:
\[
\mathcal{L}_{\mathrm{gate}}
=
-\sum_t w_{g_t}\log p(g_t \mid g_{<t}, \mathcal{M}_{\mathrm{per}}),
\qquad g_t\in\{\text{silent},\text{speak}\}.
\]
When $g_t=\text{speak}$, the recent $N$-segment codec visual canvas is passed directly to the frozen language model for autoregressive generation. Restricting optimization to the cognition gate adds proactive response timing while preserving the image and video understanding capabilities acquired during the preceding stages.
\section{Experiments}
\label{sec:experiments}

\subsection{Experimental Setup}
\label{sec:exp_setup}

\begin{table}[t]
    \centering
    \small
    \setlength{\tabcolsep}{5pt}
    \caption{\textbf{Image and video representation quality of \magevit.} Image classification is evaluated using linear probes with 256 visual tokens per image. Video recognition is evaluated using attentive probes with a fixed budget of 4096 visual tokens per video. Chunk-wise evaluation uniformly samples 16 frames with 256 tokens per frame, whereas codec-based evaluation distributes the same budget over 64 frames using sparse patch selection. Baseline video encoders are evaluated using their 16-frame representations. All values denote top-1 accuracy (\%).}
    \label{tab:summary_img_video}
    \begin{adjustbox}{max width=\textwidth}
    \begin{tabular}{l l | c | c c c c c}
    \toprule
    \multicolumn{8}{c}{\textit{Image Benchmarks}} \\
    \midrule
    Method & Backbone & Mode & DTD~\cite{bench_dtd} & CIFAR-10~\cite{bench_cifar10} & SUN397~\cite{bench_sun397} & Food-101~\cite{bench_food101} & ImageNet~\cite{bench_imagenet} \\
    \midrule
    SigLIP~\cite{zhai2023siglip}     & \textcolor{gray}{ViT-Large/16}  & \textcolor{gray}{Image} & 85.90 & 98.24 & 80.64 & 95.15          & 84.46          \\
    MetaCLIP2~\cite{model_metaclip2}  & \textcolor{gray}{ViT-Large/14}  & \textcolor{gray}{Image} & 85.48 & 98.76 & 79.75 & 93.93          & 82.74          \\
    AIMv2~\cite{model_aimv2}      & \textcolor{gray}{ViT-Large/14}  & \textcolor{gray}{Image} & 86.28 & 98.95 & 81.12 & 94.76          & 85.26          \\
    SigLIP2~\cite{tschannen2025siglip}    & \textcolor{gray}{ViT-Large/16}  & \textcolor{gray}{Image} & 85.90 & 98.31 & \best 82.36 & \best 95.85 & \best 85.92 \\
    DINOv3~\cite{model_dinov3}     & \textcolor{gray}{ViT-Large/14}  & \textcolor{gray}{Image} & \best 86.81 & \second 99.17 & 80.11 & 94.96          & \second 85.38          \\
    MoonViT~\cite{model_moonvit}    & \textcolor{gray}{ViT-SO400M/14} & \textcolor{gray}{Image} & 82.77 & 96.81 & 77.83 & 88.56          & 77.18          \\
    OV-Encoder~\cite{ovencoder2026} & \textcolor{gray}{ViT-Large/14}  & \textcolor{gray}{Image} & 85.48 & 99.06 & 80.60 & 94.79          & 84.54          \\
    \midrule
    \textbf{\magevit} & \textcolor{gray}{ViT-Large/16} & \textcolor{gray}{Image} & \second 85.96 & \best 99.33 & \second 82.01 & \second 95.60 & 85.69 \\
    \midrule
    \multicolumn{8}{c}{\textit{Video Benchmarks}} \\
    \midrule
    Method & Backbone & Mode & Diving-48~\cite{bench_diving48} & HMDB-51~\cite{bench_hmdb51} & Perception~\cite{bench_perceptiontest} & Charades-Ego~\cite{bench_charadesego} & K400~\cite{bench_kinetics400} \\
    \midrule
    SigLIP~\cite{zhai2023siglip}     & \textcolor{gray}{ViT-Large/16}  & \textcolor{gray}{Chunk} & 54.70 & 78.80 & 51.00 & 11.70 & 79.10 \\
    MetaCLIP2~\cite{model_metaclip2}  & \textcolor{gray}{ViT-Large/14}  & \textcolor{gray}{Chunk} & 42.10 & 78.20 & 51.10 & 11.20 & 84.00 \\
    AIMv2~\cite{model_aimv2}      & \textcolor{gray}{ViT-Large/14}  & \textcolor{gray}{Chunk} & 55.70 & 82.60 & 56.40 & 12.40 & 82.20 \\
    SigLIP2~\cite{tschannen2025siglip}    & \textcolor{gray}{ViT-Large/16}  & \textcolor{gray}{Chunk} & 62.30 & 84.61 & 58.47 & 13.75 & 83.93 \\
    DINOv3~\cite{model_dinov3}     & \textcolor{gray}{ViT-Large/14}  & \textcolor{gray}{Chunk} & 61.30 & 79.70 & \best 60.80 & \best 14.00 & 83.90 \\
    MoonViT~\cite{model_moonvit}    & \textcolor{gray}{ViT-SO400M/14} & \textcolor{gray}{Chunk} & 43.95 & 77.17 & 56.95 & 10.56 & 79.73 \\
    \multirow{2}{*}{OV-Encoder~\cite{ovencoder2026}} & \multirow{2}{*}{\textcolor{gray}{ViT-Large/14}} & \textcolor{gray}{Chunk} & 60.86 & 83.68 & 60.04 & 12.61 & \second 84.81 \\
     & & \textcolor{gray}{Codec} & \best 65.32 & 84.80 & \second 60.16 & 12.81 & \best 84.87 \\
    \midrule
    \multirow{2}{*}{\textbf{\magevit}} & \multirow{2}{*}{\textcolor{gray}{ViT-Large/16}} & \textcolor{gray}{Chunk} & 60.45 & \second 85.13 & 58.91 & \second 13.49 & \second 84.83 \\
     & & \textcolor{gray}{Codec} & \second 64.14 & \best 85.17 & 59.17 & 13.31 & 84.66 \\
    \bottomrule
    \end{tabular}%
    \end{adjustbox}
\end{table}

\paragraph{Models under evaluation.}
We evaluate \magevl-4B, which combines the from-scratch \magevit visual tokenizer (\cref{sec:magevit}) with a Qwen3-4B-Instruct-2507 language backbone~\citep{qwen3}. At inference time, videos are represented as codec-token streams on a shared $16{\times}16$ patch grid: anchor frames are encoded densely, whereas predicted frames contribute only patches selected by the codec-derived importance signal. We report three codec-canvas operating points with different accuracy--efficiency trade-offs. The default \emph{tc32} setting uses the largest visual budget and generally delivers the strongest performance; \emph{tc16} provides a balanced configuration with lower evaluation cost and competitive accuracy; and the lightweight \emph{tc8} setting prioritizes latency while retaining much of the higher-budget performance. Their nominal canvas budgets correspond to the visual workloads of 32, 16, and 8 uniformly sampled frames, respectively, enabling matched-budget comparisons with frame-sampling baselines.
Still images use single-image spatial patchification.

\paragraph{Baselines.}
Our primary comparison is Qwen3-VL-4B-Instruct~\citep{qwen3vl}, which uses a language model of the same parameter scale but a conventional dense visual front-end. We additionally compare with Phi-4-Multimodal-Instruct (Phi-4-MM, 5.6B)~\citep{abdin2024phi4} and Phi-4 reasoning-vision (Phi-4-R-V, 15B)~\citep{aneja2026phi4reasoningvision15btechnicalreport}. For the visual-representation study in \cref{sec:magevit_eval}, we compare \magevit with public contrastive and self-supervised encoders, including SigLIP~\citep{zhai2023siglip}, SigLIP2~\citep{tschannen2025siglip}, MetaCLIP2~\citep{model_metaclip2}, AIMv2~\citep{model_aimv2}, DINOv3~\citep{model_dinov3}, MoonViT~\citep{model_moonvit}, and OV-Encoder~\citep{ovencoder2026}.

\paragraph{Benchmarks.}
We organize the evaluation into four groups. First, representation quality (\cref{tab:summary_img_video}) directly evaluates \magevit using linear probes for image classification and attentive probes for video recognition. Second, image understanding (\cref{tab:understand_image}) covers document and chart understanding, OCR, general VQA, and 2D/3D spatial intelligence. Third, video understanding (\cref{tab:understand_video,tab:understand_video_tc8}) covers video QA, temporal grounding, video spatial reasoning, and referring-video tracking. Finally, proactive streaming is evaluated in \cref{sec:streaming_video_understanding} in terms of both response timing and response quality.

\paragraph{Protocol.}
All supported downstream benchmarks are evaluated using \texttt{lmms-eval}~\citep{zhang2025lmms} with the official prompt templates and scoring rules. We apply no model-specific prompt tuning, semantic rewriting, or manual output correction. For matched-budget video comparisons, the visual-input budget is controlled as described in \cref{sec:efficiency_analysis}. Temporal-grounding and tracking outputs are parsed using the official evaluation procedures. Wall-clock measurements in \cref{tab:understand_video_tc8} are collected on a single node with eight B200 GPUs.

\begin{figure*}[t]
    \centering
    \includegraphics[width=0.49\linewidth]
    {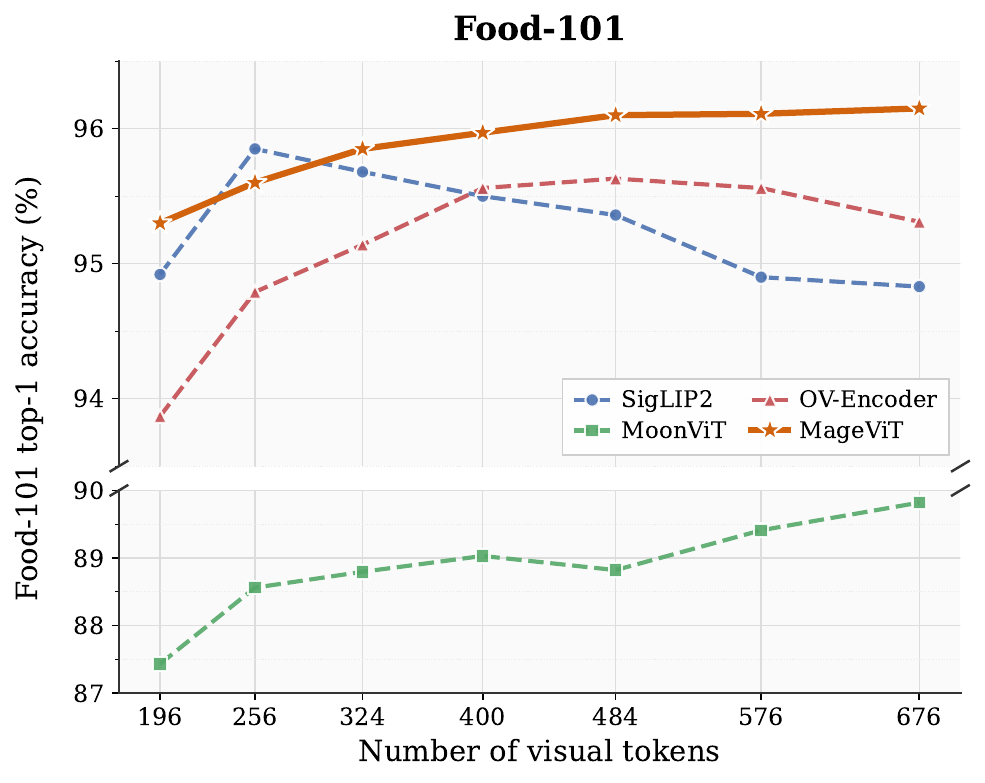}
    \hfill
    \includegraphics[width=0.49\linewidth]
    {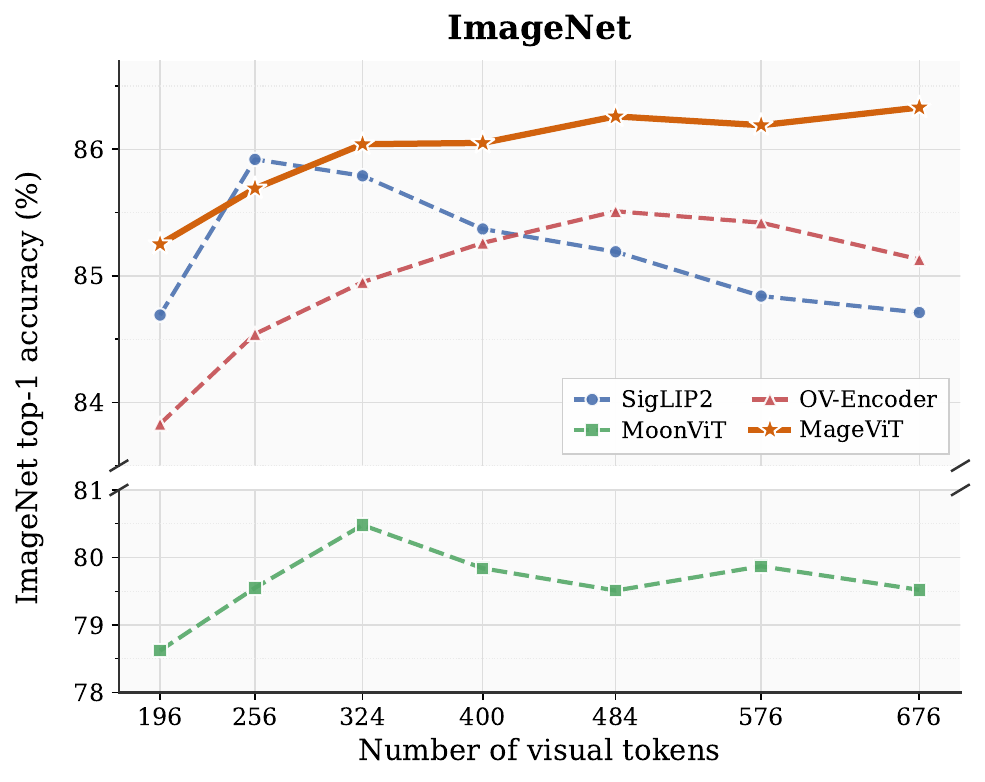}
    \caption{
    \textbf{Effect of the visual-token budget on image representation
    quality.} We evaluate frozen visual features on Food-101 and ImageNet while increasing the number of input visual tokens. \magevit, which is trained with variable-resolution inputs, consistently benefits from larger token budgets and achieves its best performance at the highest evaluated resolution. In contrast, the fixed-resolution baselines generally saturate or degrade after reaching their optimal token budget.
    }
    \label{fig:vit_variable_resolution}
\end{figure*}

\subsection{Evaluation of Mage-ViT}
\label{sec:magevit_eval}
We evaluate \magevit from three perspectives: representation quality under limited pre-training data, the effect of native-resolution training across visual-token budgets, and robustness to different codec families. Image representations are evaluated using linear probes, while video representations are evaluated using attentive probes. Unless otherwise specified, all visual encoders remain frozen.

\begin{table}[t]
\centering
\small
\setlength{\tabcolsep}{6pt}
\definecolor{canvascol}{gray}{0.45}
\newcommand{\cv}[1]{{\color{canvascol}#1}}
\caption{
\textbf{Cross-codec robustness of codec-guided patch selection.}
We compare the HEVC selector used during training with the neural
codec DCVC-RT at inference time, without codec-specific retraining.
We report the task score and the average token \cv{Canvas} per video.
DCVC-RT is constrained to use no more tokens than HEVC.}
\label{tab:dcvc_vs_hevc}
\begin{adjustbox}{max width=\textwidth}
\begin{tabular}{l c c c c}
\toprule
& \multicolumn{2}{c}{Traditional: HEVC~\cite{sullivan2012hevc}} & \multicolumn{2}{c}{Neural: DCVC-RT~\cite{jia2025towards}} \\
\cmidrule(lr){2-3} \cmidrule(lr){4-5}
Benchmark & Score & \cv{Canvas} & Score & \cv{Canvas} \\
\midrule
\multicolumn{5}{l}{\textit{Video QA}} \\
MV-Bench~\cite{bench_mvbench}             & 65.1 & \cv{34.0} & 66.5 & \cv{30.5} \\
NextQA~\cite{bench_nextqa}               & 83.1 & \cv{32.7} & 82.6 & \cv{32.1} \\
TempCompass~\cite{bench_tempcompass}          & 62.3 & \cv{34.0} & 61.7 & \cv{31.1} \\
VideoMME~\cite{bench_videomme}             & 64.0 & \cv{33.2} & 64.3 & \cv{30.2} \\
LongVideoBench~\cite{bench_longvideobench}       & 61.3 & \cv{32.6} & 60.0 & \cv{29.3} \\
LVBench~\cite{bench_lvbench}              & 41.8 & \cv{32.2} & 43.5 & \cv{29.6} \\
MLVU-dev~\cite{bench_mlvu}             & 68.7 & \cv{33.6} & 66.1 & \cv{29.9} \\
\midrule
\multicolumn{5}{l}{\textit{Temporal grounding}} \\
Charades~\cite{bench_charades_sta}             & 31.4 & \cv{33.7} & 31.6 & \cv{32.0} \\
Timelens-Charades~\cite{zhang2025timelens}    & 50.7 & \cv{33.5} & 50.5 & \cv{31.9} \\
Timelens-ActivityNet~\cite{zhang2025timelens} & 45.4 & \cv{33.7} & 44.2 & \cv{31.4} \\
Timelens-QVHighlight~\cite{zhang2025timelens} & 57.4 & \cv{34.3} & 57.5 & \cv{30.7} \\
\midrule
\multicolumn{5}{l}{\textit{Spatial reasoning}} \\
VSI-Bench~\cite{yang2024vsi}            & 64.3 & \cv{32.8} & 63.7 & \cv{31.3} \\
\midrule
\textbf{Avg}         & \textbf{58.0} & \cv{\textbf{33.4 (100\%)}} & \textbf{57.7} & \cv{\textbf{30.8 (92\%)}} \\
\bottomrule
\end{tabular}
\end{adjustbox}
\end{table}

\subsubsection{Learning Strong Visual Representations from Limited Data}
\label{sec:magevit_data_efficiency}

As summarized in \cref{tab:summary_img_video}, we comprehensively evaluate \magevit on both standard image recognition benchmarks via linear probing and video action/event recognition via attentive probing. Across both modalities, \magevit demonstrates strong representation capacities without requiring massive web-scale pre-training datasets.

\paragraph{Image representation performance.}
On standard 2D image benchmarks, \magevit delivers performance that consistently rivals or exceeds top-tier visual encoders trained on billions of images. Specifically, \magevit achieves the highest top-1 accuracy on CIFAR-10 (\textbf{99.33\%}) and exhibits highly competitive performance on coarse-grained and fine-grained classification tasks, such as SUN397 (\textbf{82.01\%}), Food-101 (\textbf{95.60\%}), and ImageNet-1K (\textbf{85.69\%}). Notably, its performance closely matches SigLIP2 (\textbf{85.92\%} on ImageNet) and surpasses established strong backbones including MetaCLIP2, AIMv2, and OV-Encoder. Importantly, while OV-Encoder relies on a smaller $14\times14$ patch size, \magevit adopts a larger $16\times16$ patch grid, achieving comparable or superior representation quality with fewer visual tokens. This confirms that our pre-training scheme effectively constructs a highly transferable and semantically rich feature space for static visual concepts using significantly fewer visual tokens and pre-training samples.

\paragraph{Video representation performance and temporal adaptation.}
When extended to temporal video understanding, \magevit demonstrates superior temporal modeling and efficiency, particularly in motion-heavy and long-horizon scenarios:
\begin{itemize}[leftmargin=*]
    \item \textbf{Chunk-wise Evaluation:} Under the standard 16-frame uniform sampling protocol, \magevit achieves a top-1 accuracy of \textbf{85.13\%} on HMDB-51 and \textbf{84.83\%} on Kinetics-400 (K400), outperforming strong baselines such as SigLIP (\textbf{78.80\%} and \textbf{79.10\%}) and DINOv3 (\textbf{79.70\%} and \textbf{83.90\%}). On fine-grained action perception benchmarks like Diving-48 (\textbf{60.45\%}), \magevit significantly outperforms traditional image-centric models, highlighting its robust spatial-temporal feature extraction.
    \item \textbf{Codec-based Sparse Sampling:} When leveraging sparse patch selection across an expanded temporal context of 64 frames under the exact same visual token budget (4096 tokens), \magevit exhibits substantial performance boosts on motion-critical tasks. Specifically, the accuracy on Diving-48 jumps from \textbf{60.45\%} to \textbf{64.14\%} (+3.69\%), and HMDB-51 reaches a peak accuracy of \textbf{85.17\%}. This confirms that the spatio-temporal representations learned by \magevit naturally pair with codec-driven sparse tokenization, enabling the model to capture fine-grained temporal dynamics over longer video horizons without increasing computational or memory overhead.
\end{itemize}

\begin{promptbox}
\large
\textbf{\textit{Finding 1:}}
\textbf{Web-Scale Pre-training is Not Essential for VLM Visual Encoders.}
Despite being trained from scratch on a modest corpus, \magevit achieves visual representation quality competitive with flagship encoders like SigLIP2, proving that architectural and objective design can effectively substitute for web-scale pre-training data.
\end{promptbox}

\subsubsection{Native-Resolution Scaling across Visual-Token Budgets}
\label{sec:magevit_token_scaling}

To further evaluate how visual representations adapt to flexible input resolutions, we analyze performance across an increasing visual token budget (from 196 to 676 tokens per image) on Food-101 and ImageNet-1K, as shown in Fig.~\ref{fig:vit_variable_resolution}.

\paragraph{Monotonic resolution scaling vs. fixed-resolution degradation.}
Visual encoders exhibit fundamentally distinct scaling behaviors depending on their pre-training resolution strategies:
\begin{itemize}[leftmargin=*]
    \item \textbf{Fixed-Resolution Baselines (SigLIP2 \& OV-Encoder):} Because these models are pre-trained on fixed spatial grids, evaluating them under larger token budgets introduces severe positional distribution shifts. Consequently, SigLIP2 reaches its peak performance at 256 tokens on ImageNet ($\sim 85.9\%$) and continuously degrades as the token budget expands to 676 ($\sim 84.7\%$). Similarly, OV-Encoder experiences a clear performance drop beyond 484 tokens.
    \item \textbf{Variable-Resolution Encoders (MoonViT \& \magevit):} Encoders pre-trained with variable-resolution objectives naturally adapt to flexible spatial formats. While MoonViT avoids high-resolution collapse, its overall representation capacity remains low across all budget settings. In contrast, \magevit achieves the best of both worlds: its representation quality monotonically improves as the visual token budget increases—reaching peak accuracy at 676 tokens ($>96.1\%$ on Food-101 and $>86.3\%$ on ImageNet)—while consistently outperforming all baselines at every evaluated resolution.
\end{itemize}
This superior extrapolation capacity confirms that \magevit seamlessly accommodates dynamic visual sequence lengths, making it an ideal visual backbone for Large Multimodal Models (LMMs).

\begin{promptbox}
\large
\textbf{\textit{Finding 2:}}
\textbf{Variable-Resolution Pre-training Enables Continuous Monotonic Scaling with Resolution and Token Budgets.}
While fixed-resolution vision encoders suffer from sharp performance degradation when evaluated at higher resolution budgets, \magevit leverages variable-resolution pre-training to achieve continuous, monotonic quality gains without saturation.
\end{promptbox}

\begin{table*}[t]
\centering
\scriptsize
\setlength{\tabcolsep}{4pt}
\caption{\textbf{Image understanding and spatial intelligence} across STEM/math, document understanding, general VQA, perception/alignment, and 2D/3D scene geometry, embodied/viewpoint, cross-view, and relational spatial benchmarks. \colorbox{bestblue}{Dark blue} and \colorbox{secondblue}{light blue} background colors indicate the best and second-best results per row, respectively.}
\label{tab:understand_image}
\begin{adjustbox}{max width=\textwidth, max totalheight=0.9\textheight}
\begin{tabular}{l c c c c}
\toprule
Benchmark & \magevl-4B & Qwen3-VL-4B~\cite{qwen3vl} & Phi-4-MM-5.6B ~\cite{abdin2024phi4} & Phi-4-R-V-15B~\cite{aneja2026phi4reasoningvision15btechnicalreport} \\
\midrule
\multicolumn{5}{l}{\textit{Document understanding}} \\
DocVQA-val~\cite{bench_docvqa}         & \best{95.14}   & \second{94.69} & 92.79          & 76.20 \\
InfoVQA-val~\cite{bench_infographicvqa}& \best{80.33}   & \second{79.50} & 71.84          & 55.41 \\
AI2D w/ Mask~\cite{bench_ai2d}         & \best{83.16}   & 81.54          & 81.83          & \second{82.87} \\
AI2D w/o Mask~\cite{bench_ai2d}                          & 91.87          & \second{92.20} & 91.35          & \best{93.81} \\
ChartQA~\cite{bench_chartqa}           & \best{84.88}   & \second{83.96} & 83.76          & 83.40 \\
OCRBench~\cite{bench_ocrbench}         & \best{81.80}   & 81.60          & \second{81.70} & 73.90 \\
CC-OCR Doc~\cite{bench_cc_ocr}         & \second{32.25} & \best{39.69}   & 4.99           & 17.65 \\
MultiDocVQA-val~\cite{bench_mp_docvqa}  & \best{87.46}   & \second{87.21} & 46.84          & 58.35 \\
DUDE~\cite{bench_dude}                 & \second{46.44} & \best{50.98}   & 3.78           & 34.34 \\
WebSRC-val~\cite{bench_websrc}         & \second{92.80} & \best{95.40}   & 71.30          & 76.60 \\
ChartQAPro~\cite{bench_chartqapro}     & \best{32.57}   & \second{26.79} & 0.13           & 25.38 \\
TextVQA-val~\cite{bench_textvqa}       & \second{77.28} & \best{80.55}   & 39.93          & 76.06 \\
CharXiv-Description~\cite{bench_charxiv}    & \second{75.85} & 74.40          & 18.65          & \best{75.98} \\
CharXiv-Reason~\cite{bench_charxiv}                         & \second{35.20} & 26.10          & 0.10           & \best{36.10} \\
\midrule
\multicolumn{5}{l}{\textit{General VQA}} \\
MMBench-EN-dev~\cite{bench_mmbench}   & \second{84.02} & 83.25          & 65.81          & \best{84.19} \\
MMBench-CN-dev~\cite{bench_mmbench}                         & \best{82.04}   & \second{80.58} & 75.17          & 79.47 \\
RealWorldQA~\cite{bench_realworldqa}   & 70.46          & \best{70.85}   & 60.65          & \second{70.72} \\
MMStar~\cite{bench_mmstar}             & \best{67.32}   & \second{62.04} & 61.24          & 59.63 \\
MME-Perception~\cite{bench_mme}        & \best{1709.54} & \second{1703.50}& 1409.66       & 1590.21 \\
SeedBench (All)~\cite{bench_seedbench} & \best{76.78}   & \second{75.65} & 68.28          & 73.70 \\
SeedBench-Image~\cite{bench_seedbench}                        & \best{79.32}   & \second{78.87} & 74.48          & 77.97 \\
SeedBench2-Plus~\cite{bench_seedbench2plus}& \second{69.21}& \best{69.65}   & 68.38          & 69.48 \\
MMT-val~\cite{bench_mmtbench}          & 64.47          & \best{66.52}   & 59.19          & \second{64.89} \\
CV-Bench~\cite{cambrian}               & \best{87.79}   & \second{85.37} & 57.09          & 81.31 \\
MME-RealWorld~\cite{bench_mmerealworld}& \best{66.52}   & \second{63.20} & 32.45          & 57.80 \\
MME-RealWorld-CN~\cite{bench_mmerealworld}                       & \second{61.33} & \best{63.09}   & 21.19          & 46.07 \\
\midrule
\multicolumn{5}{l}{\textit{Spatial intelligence}} \\
CV-Bench-2D~\cite{tong2024cambrian}                            & \best{82.13}   & \second{81.00} & 56.12          & 80.11 \\
CV-Bench-3D~\cite{tong2024cambrian}                            & \best{94.75}   & \second{92.30} & 56.92          & 82.50 \\
BLINK~\cite{bench_blink}               & \best{65.11}   & \second{65.10} & 35.24          & 57.80 \\
EmbSpatial~\cite{bench_embspatial}     & \best{82.67}   & \second{77.50} & 41.51          & 72.67 \\
CRPE-Relation~\cite{bench_crpe}        & \second{76.12} & \best{77.70}   & 34.60          & 74.46 \\
CrossPoint~\cite{wang2025crosspoint}                             & \best{80.00}   & 26.90          & 12.20          & \second{47.73} \\
ERQA~\cite{bench_erqa}                 & 36.00          & \best{42.30}   & 30.75          & \second{40.25} \\
MMSI-Bench~\cite{bench_mmsibench}      & 28.20          & \best{31.00}   & \second{28.80} & 25.70 \\
SAT~\cite{bench_sat}                   & \second{67.33} & \best{69.30}   & 55.33          & 66.67 \\
\bottomrule
\end{tabular}
\end{adjustbox}
\end{table*}

\begin{table}[t]
\centering
\small
\setlength{\tabcolsep}{6pt}
\caption{\textbf{Video understanding and temporal grounding.} \magevl and Qwen3-VL-4B share the 4B Qwen3 LLM backbone and differ only in the visual front-end (Qwen3-VL: dense ViT; \magevl: \magevit); Phi-4-MM and Phi-4-R-V are reported for reference. \colorbox{bestblue}{Dark blue} and \colorbox{secondblue}{light blue} background colors indicate the best and second-best results per row, respectively.}
\label{tab:understand_video}
\begin{adjustbox}{max width=\textwidth}
\begin{tabular}{l c c c c}
\toprule
Benchmark & \magevl-4B & Qwen3-VL-4B~\cite{qwen3vl} & Phi-4-MM-5.6B ~\cite{abdin2024phi4} & Phi-4-R-V-15B~\cite{aneja2026phi4reasoningvision15btechnicalreport} \\
\midrule
\multicolumn{5}{l}{\textit{Video QA}} \\
MV-Bench~\cite{bench_mvbench}          & \second{65.1}  & \best{66.7}   & 44.9           & 49.2 \\
NextQA~\cite{bench_nextqa}            & \best{83.1}    & \second{79.8} & 54.1           & 69.0 \\
VideoMME~\cite{bench_videomme}        & \best{64.0}    & \second{59.7} & 44.7           & 55.3 \\
LongVideoBench~\cite{bench_longvideobench}& \best{61.3} & \second{57.7} & 41.14          & 51.2 \\
LVBench~\cite{bench_lvbench}          & \best{41.8}    & \second{39.2} & 25.31          & 34.4 \\
MLVU-dev~\cite{bench_mlvu}            & \best{68.7}    & \second{61.5} & 44.18          & 51.8 \\
VideoMME (w/ sub.)~\cite{bench_videomme}                    & \second{66.3}  & \best{70.2}   & 45.4           & 58.3 \\
VideoMME-V2~\cite{fu2026video}                           & \second{24.3}  & \best{24.4}   & 19.2           & 23.9 \\
VideoEval-Pro~\cite{bench_videoevalpro}& \best{45.2}   & \second{20.7} & 14.35          & 16.8 \\
JumpScore~\cite{llavaonevision2}                             & \best{45.60}   & \second{3.82} & 3.61           & 1.53 \\
MMOU-Test-Mini~\cite{goel2026mmou}                        & \second{39.3}  & 38.1          & 32.22          & \best{51.9} \\
\midrule
\multicolumn{5}{l}{\textit{Temporal grounding}} \\
Timelens-Charades~\cite{zhang2025timelens}                     & \best{50.7}    & \second{43.1} & 4.09           & 20.6 \\
Timelens-ActivityNet~\cite{zhang2025timelens}                  & \best{45.4}    & \second{28.4} & 2.03           & 23.0 \\
Timelens-QVHighlight~\cite{zhang2025timelens}                  & \best{57.4}    & \second{34.9} & 2.47           & 11.6 \\
\midrule
\multicolumn{5}{l}{\textit{Spatial reasoning}} \\
VSI-Bench~\cite{yang2024vsi}          & \best{64.3}    & \second{53.3} & 24.09          & 25.5 \\
\midrule
\multicolumn{5}{l}{\textit{Tracking (J\&F)}} \\
Ref-DAVIS17~\cite{bench_refdavis17}   & \best{25.83}   & \second{7.48} & 3.14           & 2.15 \\
MeViS-ValidU~\cite{bench_mevis}       & \best{22.55}   & 3.16          & \second{10.28} & 1.53 \\
ReasonVOS~\cite{bench_revos}          & \best{17.76}   & 9.66          & 9.50           & \second{9.77} \\
Ref-YT-VOS~\cite{bench_refytvos}      & \best{25.57}   & \second{5.28} & 8.64           & 3.85 \\
\bottomrule
\end{tabular}
\end{adjustbox}
\end{table}

\subsubsection{Robust Token Selection across Codec Families}
\label{sec:magevit_codec_robustness}

As introduced in \cref{sec:magevit_arch}, \magevit uses a codec-derived importance map to select informative patches from predicted frames. During training, this map is computed from HEVC~\citep{sullivan2012hevc} using motion-vector magnitude and residual energy. Since these signals are specific to the coding process of a traditional codec, an important question is whether the resulting token-selection strategy generalizes to fundamentally different codec families.

We evaluate this by replacing the HEVC-based selector at inference time with DCVC-RT~\citep{jia2025towards}, a neural video codec whose learned probability model estimates local coding cost through negative log-likelihood. No codec-specific retraining or adaptation is applied. As shown in \cref{tab:dcvc_vs_hevc}, the two selectors achieve nearly identical average performance across video QA, temporal grounding, and spatial reasoning benchmarks, while the neural-codec selector uses a smaller average token canvas. Performance also remains comparable across most individual tasks, with no consistent degradation after switching codec families.

These results indicate that codec-guided token selection does not depend on the particular motion-estimation, residual-coding, or probability model used by a codec. Although HEVC and DCVC-RT implement compression through different mechanisms, both assign greater coding cost to regions that are difficult to predict from temporal context. Motion-vector magnitude, residual energy, and neural-codec negative log-likelihood therefore provide different estimates of the same underlying signal: local temporal predictability.

Importantly, \magevit consumes only the resulting patch-importance map, rather than codec-specific syntax or latent representations. The visual tokenizer is therefore decoupled from the mechanism used to estimate coding difficulty. This establishes temporal predictability as a transferable criterion for allocating visual tokens and allows codec-guided tokenization to generalize across traditional and neural compression systems.

\subsection{Evaluation of Mage-VL}
\label{sec:image_video_eval}
\subsubsection{Image Understanding}

\Cref{tab:understand_image} reports model performance across three core capability dimensions: document understanding, which aligns with Microsoft's primary domain applications; general VQA, which evaluates general multimodal proficiency; and spatial intelligence, which represents a core strategic focus of \magevl.
Against the matched-LLM Qwen3-VL-4B control, \magevl is competitive on aggregate and clearly ahead on the categories that our caption-heavy curriculum targets. On document, OCR, and chart understanding, \magevl leads on the majority of benchmarks---including DocVQA ($95.14$), InfoVQA ($80.33$), ChartQA ($84.88$), OCRBench ($81.80$), and AI2D-with-mask ($83.16$)---consistent with the dense-recaptioning and verbatim-OCR emphasis of our data pipeline. On general VQA, the two models trade wins within small margins, with \magevl achieving top results on MMStar ($67.32$), MME-Perception ($1709.54$), and CV-Bench ($87.79$), while comprehensively surpassing the similarly-sized Phi-4-MM across the board. The clearest and most consistent gains appear in the spatial-intelligence block: \magevl improves over Qwen3-VL on CV-Bench-3D ($94.75$ vs.\ $92.30$), EmbSpatial ($82.67$ vs.\ $77.50$), and dramatically on cross-view point correspondence (CrossPoint $80.00$ vs.\ $26.90$), where codec-aligned patchification preserves the dense, geometrically-consistent structure that viewpoint matching requires. Despite using roughly a quarter of its parameters, \magevl also surpasses Phi-4-R-V (15B) on the large majority of these spatial benchmarks. 

\subsubsection{Video Understanding}

\Cref{tab:understand_video} reports video understanding under the same matched-LLM comparison, so that \magevl and Qwen3-VL-4B differ only in the visual front-end.
\magevl improves on most video-QA benchmarks, including VideoMME ($+4.3$), MLVU ($+7.2$), LongVideoBench ($+3.5$), LVBench ($+2.6$), NextQA ($+3.3$), and especially the long and realistic VideoEval-Pro ($+24.5$). Qwen3-VL remains stronger on MV-Bench, TempCompass, and MMVU, which lean more on dense per-frame appearance than on temporal localization. The gains concentrate in \emph{localization-heavy} regimes: on temporal grounding \magevl improves by $+7.6$, $+17.1$, and $+22.5$ on the Timelens Charades/ActivityNet/QVHighlight splits; on video spatial reasoning by $+11.0$ on VSI-Bench; and on referring-video tracking by roughly $+18$ to $+20$ J\&F points on Ref-DAVIS17, MeViS, and Ref-YouTube-VOS (and $+8.1$ on ReasonVOS).
 
Notably, these competitive long VideoQA capabilities are achieved \emph{without any} long VideoQA SFT; instead, \magevl relies solely on short video SFT alongside detailed video captions. The dense temporal-visual alignment established by video captions enables the base LLM to handle VideoQA tasks zero-shot, bypassing the need for heavy VideoQA instruction tuning. \Cref{tab:understand_video_tc8} further shows that the lightweight tc8 setting preserves most of these gains at a fraction of the visual-token cost, confirming that the advantage does not hinge on a large frame budget.


\begin{promptbox}
\large
\textbf{\textit{Finding 3:}}
\textbf{Dense Video Captions Bypass Long VideoQA SFT.}
Without explicit long VideoQA instruction tuning, fine-tuning solely on video detailed captions (alongside short video SFT) is sufficient for strong VideoQA capabilities. High-quality captioning establishes solid temporal-visual alignment, allowing the base LLM to generalize to long VideoQA zero-shot.
\end{promptbox}

\begin{table*}[t]
\centering
\scriptsize
\setlength{\tabcolsep}{3.5pt}
\caption{\textbf{Video performance and evaluation efficiency across token/codec budgets.}
Performance uses each benchmark's primary metric ($\uparrow$); evaluation efficiency is reported as wall-clock Time in seconds ($\downarrow$) on a single 8$\times$B200 node. Qwen3-VL and Phi-4-R-V use 32 frames, and their time excludes estimated video-loading time ($1.01N/8$ and $1.01N$ for $N$ samples, respectively), whereas Mage-VL reports full measured wall-clock time. \colorbox{bestblue}{Dark blue} and \colorbox{secondblue}{light blue} backgrounds indicate the best and second-best performance per row.}
\label{tab:understand_video_tc8}
\begin{adjustbox}{max width=\textwidth}
\begin{tabular}{l *{5}{c c}}
\toprule
& \multicolumn{2}{c}{\magevl-4B (tc8)} 
& \multicolumn{2}{c}{\magevl-4B (tc16)} 
& \multicolumn{2}{c}{\magevl-4B (tc32)} 
& \multicolumn{2}{c}{Qwen3-VL-4B~\cite{qwen3vl}} 
& \multicolumn{2}{c}{Phi-4-R-V-15B~\cite{aneja2026phi4reasoningvision15btechnicalreport}} \\
\cmidrule(lr){2-3} \cmidrule(lr){4-5} \cmidrule(lr){6-7} \cmidrule(lr){8-9} \cmidrule(lr){10-11}
Benchmark & Perf. ($\uparrow$) & Time (s) ($\downarrow$) & Perf. ($\uparrow$) & Time (s) ($\downarrow$) & Perf. ($\uparrow$) & Time (s) ($\downarrow$) & Perf. ($\uparrow$) & Time (s) ($\downarrow$) & Perf. ($\uparrow$) & Time (s) ($\downarrow$) \\
\midrule
\multicolumn{11}{l}{\textit{Video QA}} \\
MV-Bench~\cite{bench_mvbench}             & 65.1 & 893 & \second{65.5} & 972  & 65.1          & 1054 & \best{66.7}   & 1268 & 49.2 & 2923 \\
NextQA~\cite{bench_nextqa}               & 80.8 & 415 & \second{82.4} & 502  & \best{83.1}   & 642  & 79.8          & 1460 & 69.0 & 29024 \\
TempCompass~\cite{bench_tempcompass} & 61.5 & 389 & \second{62.4} & 510 & 62.3 & 729  & \best{72.7}   & 433  & 34.8 & 3518 \\
VideoMME~\cite{bench_videomme}             & 57.9 & 270 & \second{61.8} & 439  & \best{64.0}   & 534  & 59.7          & 463  & 55.3 & 3187 \\
LongVideoBench~\cite{bench_longvideobench}       & 56.2 & 279 & \best{63.1}   & 1308 & \second{61.3} & 345  & 57.7          & 365  & 51.2 & 2617 \\
LVBench~\cite{bench_lvbench}              & 38.9 & 213 & 38.9          & 266  & \best{41.8}   & 333  & \second{39.2} & 554  & 34.4 & 2692 \\
MLVU-dev~\cite{bench_mlvu}             & 63.2 & 296 & \second{65.6} & 326  & \best{68.7}   & 361  & 61.5          & 786  & 51.8 & 6326 \\
\midrule
\multicolumn{11}{l}{\textit{Temporal grounding}} \\
Charades~\cite{bench_charades_sta} & 34.1 & 534 & 33.2 & 417 & 31.4 & 729 & \best{45.9} & 707 & 0.3 & 1844 \\
Timelens-Charades~\cite{zhang2025timelens}    & 46.4 & 483 & \second{50.4} & 538  & \best{50.7}   & 623  & 43.1          & 643  & 20.6 & 3607 \\
Timelens-ActivityNet~\cite{zhang2025timelens} & 32.9 & 554 & \second{39.7} & 622  & \best{45.4}   & 785  & 28.4          & 766  & 23.0 & 4611 \\
Timelens-QVHighlight~\cite{zhang2025timelens} & 42.6 & 357 & \second{52.1} & 358  & \best{57.4}   & 421  & 34.9          & 402  & 11.6 & 1643 \\
\midrule
\multicolumn{11}{l}{\textit{Spatial reasoning}} \\
VSI-Bench~\cite{yang2024vsi}            & 55.7 & 329 & \second{61.1} & 368  & \best{64.3}   & 537  & 53.3          & 255  & 25.5 & 4360 \\
\bottomrule
\end{tabular}
\end{adjustbox}
\end{table*}

Beyond temporal grounding and video-level reasoning, we observe an intriguing positive transfer from dynamic video training to static spatial intelligence. 
While traditional paradigms often treat spatial reasoning and temporal video understanding as isolated capabilities, exposure to continuous video streams—rich in egocentric perspective shifts, camera trajectories, and object-environment interactions—provides a strong inductive bias for fine-grained 2D/3D geometric perception. 
Although rigorous single-variable ablations remain challenging due to the tightly coupled joint pre-training recipe, our empirical results across spatial benchmarks (e.g., CV-Bench and EmbSpatial) strongly support this cross-task synergy.

\begin{promptbox}
\large
\textbf{\textit{Finding 4:}}
\textbf{Dynamic Video Training Synergizes with Spatial Intelligence.}
Incorporating dynamic video sequences with rich viewpoint transitions and spatial interactions enhances 2D/3D spatial reasoning capabilities, allowing \magevl to achieve strong performance on static spatial benchmarks alongside temporal video tasks.
\end{promptbox}

\subsection{Streaming Video Understanding}
\label{sec:streaming_video_understanding}

To systematically evaluate \magevl in continuous video streams, we investigate both its \emph{response timing} (when to speak) and \emph{response quality} (what to say), accompanied by a qualitative case study on real-world sports broadcasts.

\paragraph{Response timing.} 
We evaluate our codec-native streaming model on the SoccerNet-Caption validation split following the StreamMind evaluation protocol~\citep{ding2025streammindunlockingframerate}. The decision gate is evaluated at every valid streaming position via an $\mathrm{argmax}$ over the \emph{silent}/\emph{speak} logits using exact canvas-position matching with zero temporal tolerance. 
We report TriggerAcc (fraction of correctly predicted gate positions) and TimVal (which balances speak and silence correctness), both macro-averaged across 98 match halves. Additionally, raw per-position F1, ROC-AUC, and PR-AUC are reported without temporal tolerance. 

For reference, baseline numbers for StreamMind are directly taken from its original publication~\citep{ding2025streammindunlockingframerate}. Notably, StreamMind was explicitly trained in-distribution on SoccerNet-Caption, whereas \magevl achieves superior event-triggering performance under strict, zero-tolerance canvas-position matching without domain-specific over-tuning. Note that JoyAI is evaluated at 1~Hz with a relaxed $\pm1$-second window.

As presented in Table~\ref{tab:response_timing}, JoyAI-VL-Interaction exhibits a severe bias toward predicting silence. Due to the extreme class imbalance in SoccerNet-Caption (where silent frames dominate), JoyAI achieves an artificially inflated TriggerAcc ($97.98\%$) but suffers a severe degradation on precision-sensitive and balanced metrics (e.g., $19.25\%$ TimVal and $3.55\%$ F1). 
In contrast, \magevl significantly outperforms StreamMind—despite StreamMind being trained \emph{in-distribution} on SoccerNet-Caption—achieving $79.21\%$ TriggerAcc and a state-of-the-art $55.54\%$ TimVal. This superior performance across ROC-AUC ($83.14\%$) and PR-AUC ($9.30\%$) demonstrates that \magevl achieves well-calibrated, timely response triggering without collapsing into trivial silent predictions.

\begin{table}[t]
\centering
\small
\setlength{\tabcolsep}{6pt}
\caption{\textbf{Response timing evaluation on SoccerNet-Caption~\cite{mkhallati2023soccernet}.} Metrics include TriggerAcc, TimVal, per-position F1, ROC-AUC, and PR-AUC. Best results per column are in \textbf{bold}.}
\label{tab:response_timing}
\begin{tabular}{l c c c c c}
\toprule
Method & TriggerAcc & TimVal & F1 & ROC-AUC & PR-AUC \\
\midrule
StreamMind~\cite{ding2025streammindunlockingframerate} & 52.18 & 47.36 & -- & -- & -- \\
JoyAI-VL-Interaction-9B~\cite{joyai2026vlinteraction}     & \textbf{97.98} & 19.25 & 3.55 & 56.26 & 1.68 \\
Mage-VL-4B                                      & 79.21 & \textbf{55.54} & \textbf{16.35} & \textbf{83.14} & \textbf{9.30} \\
\bottomrule
\end{tabular}
\end{table}

\paragraph{Response quality.} 
Beyond timing accuracy, we evaluate the streaming perception quality of \magevl on OVO-Bench~\citep{li2025ovobench} under the recent-window protocol established by SimpleStream~\citep{shen2026simplestream}, where queries are answered using the four most recent frames sampled at 1\,fps. 
As detailed in Table~\ref{tab:main_ovo}, \magevl achieves $79.84\%$ average accuracy on Real-Time Visual Perception and $48.15\%$ on Backward Tracing, establishing a new state-of-the-art overall OVO-Bench score of $64.00\%$ among streaming architectures. Crucially, this competitive real-time perception capability is attained without requiring streaming-specific fine-tuning or heavy external memory modules.

\begin{table*}[t]
\centering
\caption{\textbf{Results on OVO-Bench.}
Baseline results and table structure are adapted from SimpleStream~\citep{shen2026simplestream}.
OVO-Bench reports per-task accuracy under Real-Time Visual Perception and Backward Tracing.
$\dagger$: Qwen2.5-VL-7B + HERMES (4K tokens).
OCR, ACR, ATR, STU, FPD, and OJR denote Optical Character Recognition, Action Recognition,
Attribute Recognition, Spatial Understanding, Future Prediction, and Object Recognition;
EPM, ASI, and HLD denote Episodic Memory, Action Sequence Identification, and Hallucination
Detection. ``Avg.'' is the mean of the Real-Time and Backward category averages. 
\colorbox{bestblue}{Dark blue} and \colorbox{secondblue}{light blue} backgrounds highlight the best and second-best results across all models, respectively.}
\label{tab:main_ovo}
\renewcommand{\arraystretch}{1.08}
\setlength{\tabcolsep}{3pt}
\resizebox{\textwidth}{!}{%
\begin{tabular}{lc|cccccc|c|ccc|c|c}
\toprule
\multirow{3}{*}{Model} & \multirow{3}{*}{\#Frames}
& \multicolumn{12}{c}{\textit{OVO-Bench}~\citep{li2025ovobench}} \\
\cmidrule(lr){3-14}
& & \multicolumn{7}{c|}{\textit{Real-Time Visual Perception}}
& \multicolumn{4}{c|}{\textit{Backward Tracing}}
& \multirow{2}{*}{Avg.} \\
\cmidrule(lr){3-9} \cmidrule(lr){10-13}
& & OCR & ACR & ATR & STU & FPD & OJR & Avg.
& EPM & ASI & HLD & Avg. & \\
\midrule
Human & --
& 94.0 & 92.6 & 94.8 & 92.7 & 91.1 & 94.0 & 93.2
& 92.6 & 93.0 & 91.4 & 92.3 & 92.77 \\
\midrule
\multicolumn{14}{c}{\textit{Offline Video LLMs}} \\
\midrule
Qwen2.5-VL-7B~\cite{bai2025qwen25vl} & 1\,fps
& 67.8 & 55.1 & 67.2 & 42.1 & 66.3 & 60.9 & 59.9
& 51.5 & 58.8 & 23.7 & 44.7 & 52.28 \\
LLaVA-Video-7B~\cite{model_llavavideo} & 64
& 69.1 & 58.7 & 68.8 & 49.4 & 74.3 & 59.8 & 63.5
& 56.2 & 57.4 & 7.5 & 40.4 & 51.95 \\
Qwen3-VL-4B~\cite{qwen3vl} & 64
& 77.9 & \second{69.7} & \second{77.6} & \second{60.1} & 76.2 & \best{75.5} & \second{72.8}
& \best{63.0} & \second{66.2} & 30.1 & \best{53.1} & \second{63.00} \\
\midrule
\multicolumn{14}{c}{\textit{Online / Streaming Video LLMs}} \\
\midrule
VideoLLM-online-8B~\cite{chen2024videollmonline} & 2\,fps
& 8.1 & 23.9 & 12.1 & 14.0 & 45.5 & 21.2 & 20.8
& 22.2 & 18.8 & 12.2 & 17.7 & 19.26 \\
Flash-VStream-7B~\cite{zhang2024flashvstream} & 1\,fps
& 24.2 & 29.4 & 28.5 & 33.7 & 25.7 & 28.8 & 28.4
& 39.1 & 37.2 & 5.9 & 27.4 & 27.90 \\
Dispider-7B~\cite{qian2025dispider} & 1\,fps
& 57.7 & 49.5 & 62.1 & 44.9 & 61.4 & 51.6 & 54.6
& 48.5 & 55.4 & 4.3 & 36.1 & 45.35 \\
TimeChat-Online-7B~\cite{model_timechatonline} & 1\,fps
& 75.2 & 46.8 & 70.7 & 47.8 & 69.3 & 61.4 & 61.9
& 55.9 & 59.5 & 9.7 & 41.7 & 51.80 \\
StreamForest-7B~\cite{model_streamforest} & 1\,fps
& 68.5 & 53.2 & 71.6 & 47.8 & 65.4 & 60.9 & 61.2
& \second{58.9} & 64.9 & 32.3 & 52.0 & 56.60 \\
Streamo-7B~\cite{model_streamo} & 1\,fps
& 79.2 & 57.8 & 75.0 & 49.4 & 64.4 & 70.1 & 66.0
& 54.6 & 52.0 & 31.7 & 46.1 & 56.05 \\
HERMES-7B$^\dagger$~\cite{model_hermes} & 1\,fps
& \second{85.2} & 64.2 & 71.6 & 53.4 & \second{74.3} & 65.2 & 69.0
& 48.5 & 62.2 & \best{37.6} & 49.4 & 59.20 \\
JoyAI-VL-Interaction-9B~\citep{joyai2026vlinteraction} & 1\,fps
& 72.5 & 61.5 & 75.0 & 59.0 & \best{78.2} & 64.1 & 68.4
& 62.0 & \best{70.9} & 12.9 & 48.6 & 58.50 \\
\textbf{\magevl-4B} & 1\,fps
& \best{94.63} & \best{83.49} & \best{79.31} & \best{74.16} & 70.30 & \second{77.17} & \best{79.84}
& 50.17 & 61.49 & \second{32.80} & \second{48.15} & \best{64.00} \\
\bottomrule
\end{tabular}%
}
\end{table*}

\begin{figure*}[h]
  \centering
  \includegraphics[width=\textwidth]{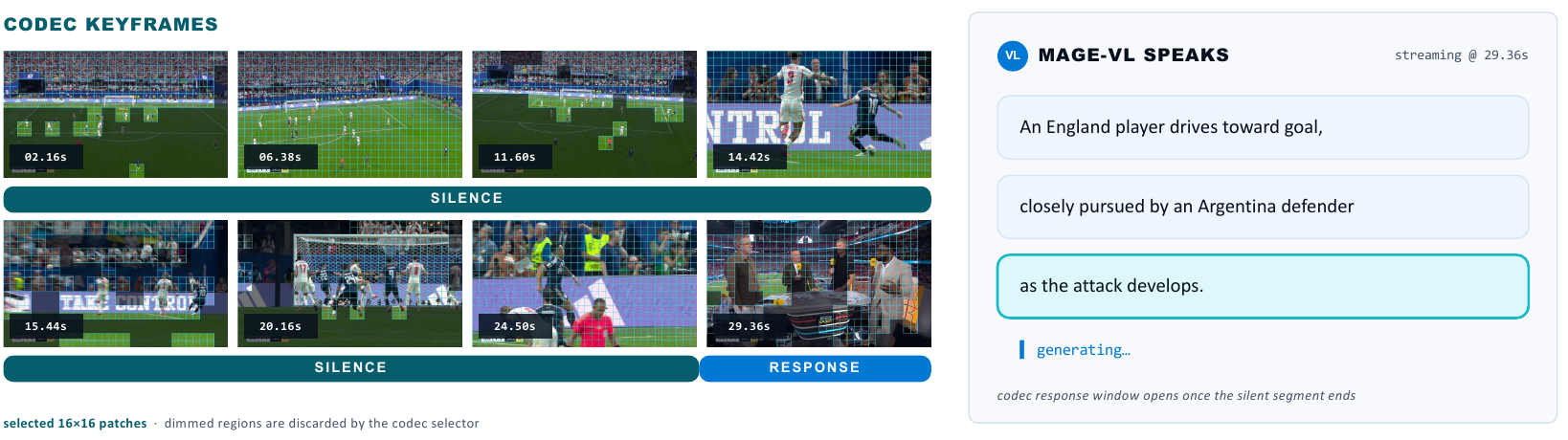}
  \caption{\textbf{Codec-native proactive streaming on a 2026 World Cup broadcast (England vs. Argentina).}
  Eight frames from a causal match window visualize the $16{\times}16$ patches retained by the codec selector; dimmed regions are discarded as temporally predictable. The event gate remains silent through routine updates and triggers the language decoder at $29.36\text{s}$ when a response-worthy offensive play is observed. The speech bubble shows the real-time generated response.}
  \label{fig:streaming_case}
\end{figure*}

\paragraph{Qualitative illustration.} 
Fig.~\ref{fig:streaming_case} presents a qualitative case study on a live 2026 World Cup broadcast between England and Argentina outside the SoccerNet evaluation split. As visualized, our codec front-end dynamically retains spatial patches ($16{\times}16$) associated with salient motion and novel visual details while suppressing temporally predictable background regions. Throughout routine gameplay ($02.16\text{s}$--$24.50\text{s}$), the event gate stays silent. At $29.36\text{s}$, upon detecting an active offensive play—where an England player drives toward the goal under close pursuit by an Argentina defender—the event gate triggers the language decoder to proactively generate a real-time commentary (\emph{``An England player drives toward goal, closely pursued by an Argentina defender as the attack develops.''}). This demonstrates the seamless synergy between codec-native sparse perception and event-driven response generation in demanding live streaming scenarios.

\subsection{Efficiency Analysis}
\label{sec:efficiency_analysis}

To understand how codec-native tokenization alters the trade-off between visual budget and downstream accuracy, we analyze both the scaling trends under matched visual input budgets and the wall-clock evaluation efficiency.

\paragraph{Matched-budget setup.}
We evaluate \emph{codec-input efficiency} by comparing downstream performance under an equivalent nominal visual input capacity to the vision encoder. 
For the uniform baseline, \emph{frame-$N$} denotes Qwen3-VL-4B evaluated on $N \in \{4, 8, 16, 32, 64\}$ uniformly sampled RGB frames. For \magevl, \emph{tc-$N$} denotes a target budget of $N$ codec canvases, constructed via adaptive temporal grouping and salient patch packing over an $8\times$ denser source-frame sequence ($8N$ raw frames). Following LLaVA-OneVision-2~\cite{llavaonevision2}, both paradigms share identical vision architectures ($16{\times}16$ patch size, max 150K pixels per input canvas). Consequently, frame-$N$ and tc-$N$ subject the vision backbone to the same nominal token workload, but tc-$N$ adaptively concentrates representation capacity onto motion-salient regions across a significantly wider temporal horizon.

\begin{figure*}[t]
  \centering
  \includegraphics[width=\textwidth]{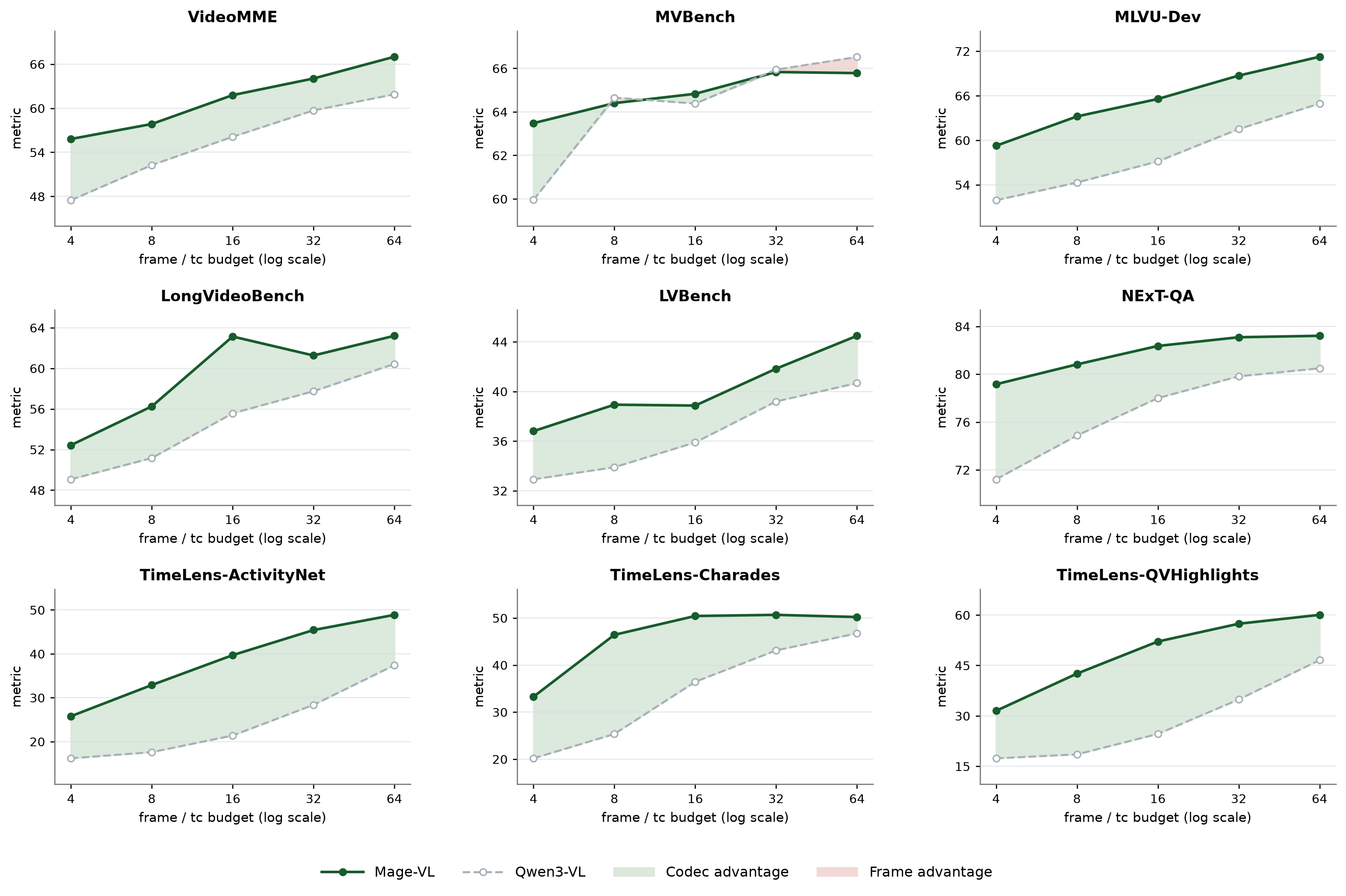}
  \caption{\textbf{Codec inputs versus uniform frame sampling across input budgets.}
  Frame-$N$ uses $N$ uniformly sampled RGB frames, whereas tc-$N$ targets $N$ codec canvases constructed from $8N$ source frames. Shaded regions highlight the performance gain of codec-native tokenization.}
  \label{fig:tc_vs_frame_scaling}
\end{figure*}

\paragraph{Accuracy-budget scaling.}
As illustrated in Fig.~\ref{fig:tc_vs_frame_scaling}, codec-native tokenization establishes a consistently superior accuracy-efficiency frontier over uniform sampling across long-video and fine-grained temporal tasks. 
Crucially, the performance advantage (\emph{green shaded area}) is most pronounced in low-to-medium budget regimes ($N \le 16$). On dense temporal localization benchmarks such as TimeLens-ActivityNet, TimeLens-Charades, and TimeLens-QVHighlights, sparse uniform sampling frequently misses crucial short-duration events, whereas tc-$N$ retains vital temporal evidence even at $N=4$. 
On long-video QA benchmarks (e.g., VideoMME, LVBench, and MLVU-Dev), tc-$N$ steadily outscales uniform sampling across all budgets ($4 \le N \le 64$). 
The only notable boundary case occurs on MVBench, where both representations converge at high frame budgets, as its short clip structures offer fewer temporally redundant regions for codec compression to exploit. Overall, codec tokenization acts as a dynamic capacity reallocator—pruning temporal redundancies to maximize information density per visual token.

\paragraph{Pareto efficiency in wall-clock time.}
Table~\ref{tab:understand_video_tc8} further highlights the practical wall-clock inference advantages of \magevl. 
By compressing redundant video content prior to vision encoding, \magevl achieves superior accuracy while drastically reducing latency. For instance, on VideoMME, \magevl under \emph{tc8} achieves $57.9\%$ accuracy in just $270\,\mathrm{s}$ wall-clock time—outperforming Qwen3-VL ($59.7\%$ at $463\,\mathrm{s}$) in compute time while matching its accuracy at \emph{tc16} ($61.8\%$ at $439\,\mathrm{s}$). On NExT-QA~\cite{xiao2021next}, \magevl (\emph{tc8}) cuts the evaluation wall-clock time from $1460\,\mathrm{s}$ down to $415\,\mathrm{s}$ ($3.5\times$ speedup) while achieving higher accuracy ($80.8\%$ vs. $79.8\%$). These results confirm that codec-native visual representation delivers genuine end-to-end inference speedups alongside strong empirical accuracy gains.

\begin{promptbox}
\large
\textbf{\textit{Finding 5:}}
\textbf{Codec-Native Inputs Significantly Boost Video Representation Efficiency.}
By dynamically packing motion-salient patches across extended temporal horizons, codec-native tokenization establishes a superior accuracy-efficiency frontier over uniform frame sampling, achieving substantially higher performance under matched token budgets and up to $3.5\times$ wall-clock inference speedups.
\end{promptbox}

\section{Discussions}

\subsection{AI4AI Data Pipeline}
\label{subsec:caption_prompt_ablation}

To evaluate the effectiveness of the prompt optimization pipeline, we conduct an ablation study by training our model on the optimized dataset, replacing the standard LLaVA-OV~1.5 full midtrain data. Performance is measured across nine standard multimodal benchmarks. As summarized in Table~\ref{tab:caption_results}, the optimized caption pipeline leads to consistent improvements across all evaluated tasks.

\begin{table}[t]
\centering
\caption{Downstream performance comparison before and after image captioning pipeline optimization.}
\label{tab:caption_results}
\vspace{0.5em}
\begin{adjustbox}{max width=\textwidth}
\begin{tabular}{lccccccccc}
\toprule
\textbf{Configuration} & \textbf{AI2D}~\cite{bench_ai2d} & \textbf{ChartQA}~\cite{bench_chartqa} & \textbf{InfoVQA}~\cite{bench_infographicvqa} & \textbf{MMStar}~\cite{bench_mmstar} & \textbf{OCRBench}~\cite{bench_ocrbench} & \textbf{MMBench-CN}~\cite{bench_mmbench} & \textbf{DocVQA}~\cite{bench_docvqa} & \textbf{CV-Bench}~\cite{cambrian} & \textbf{Realworld QA}~\cite{bench_realworldqa} \\
\midrule
Before Optimization & 82.74 & 84.80 & 74.26 & 62.91 & 76.90 & 78.78 & 93.66 & 78.51 & 69.41 \\
After Optimization  & \textbf{82.97} & \textbf{86.32} & \textbf{79.88} & \textbf{63.43} & \textbf{80.70} & \textbf{80.15} & \textbf{94.74} & \textbf{79.68} & \textbf{73.20} \\
\midrule
Gain ($\Delta$)     & +0.23 & +1.52 & +5.62 & +0.52 & +3.80 & +1.37 & +1.08 & +1.17 & +3.79 \\
\bottomrule
\end{tabular}%
\end{adjustbox}
\end{table}

The evaluation results show consistent performance gains across multiple capabilities:
\begin{enumerate}
    \item \textbf{Text and Document Comprehension:} The most notable gains are observed on document- and text-heavy benchmarks, such as InfoVQA (+5.62), OCRBench (+3.80), ChartQA (+1.52), and DocVQA (+1.08). This aligns with the explicit OCR quality metric incorporated in the evaluation sub-agent, which encourages the prompt optimization process to retain dense visual text details in the captions.
    \item \textbf{Generalization Capabilities:} Performance gains also extend to general visual reasoning and perception benchmarks, including RealWorldQA (+3.79) and MMBench-CN (+1.37). This indicates that multi-aspect feedback helps improve caption quality globally without leading to narrow task-specific overfitting.
\end{enumerate}

\begin{promptbox}
\large
\textbf{\textit{Finding 6:}}
\textbf{AI-Driven Data Pipeline Optimization Enhances Downstream Performance.}
Iterative dataset optimization guided by multi-dimensional LLM rubric scoring consistently yields global quality gains across diverse benchmarks. When linguistic prompt tuning hits a ceiling, the agent further extends to harness-level code modifications—such as video timestamp overlays—enabling effective code-prompt co-design.
\end{promptbox}

\paragraph{Emergent Behavior in Video Caption Pipeline.} During the iteration of the video captioning pipeline, the evaluation sub-agent identified a recurring error pattern: the caption model GPT-4.1 frequently struggled with precise temporal localization and timestamp grounding when relying solely on generic text prompts. Modifying the prompt text alone proved insufficient to resolve this bottleneck due to the challenge of implicit temporal alignment across video frames. To address this limitation, the pipeline established a synergy between harness code and prompted model skills. Instead of requiring the model to infer time implicitly through text instructions, the pipeline harness code modifies the input by drawing sequential timestamp overlays directly onto the sampled video frames prior to model encoding. The prompt (skill) then guides the model to utilize its visual OCR recognition capabilities to explicitly anchor events to these rendered timestamps. By converting implicit temporal tracking into a visual text-grounding task, this code-prompt co-design helps reduce temporal hallucinations and improves caption consistency across extended video sequences.

\paragraph{AI-Guided Training Diagnostics \& Recipe Optimization.} Through iterative AI-driven diagnostics, we systematically identified critical data and architectural bottlenecks, which directly guided our training configurations. First, the evaluation agent diagnosed a distinct vulnerability in the model's precise temporal localization capabilities; to remedy this, we strategically supplemented our dataset with high-quality temporal grounding and localization data. Second, scaling investigations revealed that training with a larger number of frames consistently boosts long-video performance, whereas incorporating dedicated long-video SFT data yielded no noticeable improvements, and scaling the spatial resolution beyond 384 pixels offered only marginal gains. Consequently, we adopted a spatial resolution of 384 and set the temporal length of Stage~3 to 384 frames. Finally, we discovered that increasing the RoPE base frequency ($\theta$) to 8M was highly beneficial to stabilizing and enhancing long-context sequence modeling over extended timelines. Together, these AI-derived diagnostics uniquely determined our final training length, hyperparameters, and data recipes.

\begin{table}[t]
\centering
\scriptsize
\setlength{\tabcolsep}{5pt}
\renewcommand{\arraystretch}{1.08}
\caption{
\textbf{Effectiveness of Skipping Vision SFT before Multimodal RL.} 
We compare our proposed \textit{Zero-Vision SFT + RL} pipeline against the standard \textit{OV1.5 Quick Start + RL} baseline across 24 multimodal benchmarks. 
$\Delta$ denotes the net improvement of Zero-Vision RL over Quick Start RL ($(\text{B}) - (\text{C})$). 
Bold text indicates superior performance between the two post-RL models.
}
\label{tab:sft_init_rl_optimized}
\begin{adjustbox}{max width=\linewidth}
\begin{tabular}{@{}l c cc c c@{}}
\toprule
 & \multicolumn{2}{c}{\textbf{Zero-Vision Paradigm}} & \multicolumn{1}{c}{\textbf{Baseline Paradigm}} & \\
\cmidrule(lr){2-3} \cmidrule(lr){4-4}
\textbf{Benchmark} & \textbf{(A) SFT Only} & \textbf{(B) + RL (Ours)} & \textbf{(C) QS + RL} & \textbf{$\Delta$ (B vs. C)} \\
\midrule

\multicolumn{5}{@{}l}{\textbf{\textit{General VQA}}} \\
MMStar~\cite{bench_mmstar} & 48.73 & \textbf{54.80} & 50.33 & $+4.47$ \\
MMBench-EN~\cite{bench_mmbench} & 81.10 & \textbf{83.69} & 81.25 & $+2.44$ \\
MMBench-CN~\cite{bench_mmbench} & 79.16 & 81.59 & \textbf{82.91} & $-1.32$ \\
MME-RealWorld-CN~\cite{bench_mmerealworld} & 34.32 & 43.40 & \textbf{43.82} & $-0.42$ \\
MME-RealWorld-EN~\cite{bench_mmerealworld} & 45.25 & 47.55 & \textbf{49.31} & $-1.76$ \\
SeedBench-Image~\cite{bench_seedbench} & 75.13 & \textbf{75.86} & 73.78 & $+2.08$ \\
SeedBench-2-Plus~\cite{bench_seedbench2plus} & 59.95 & \textbf{62.85} & 60.25 & $+2.60$ \\
CV-Bench~\cite{cambrian} & 68.76 & \textbf{73.88} & 69.14 & $+4.74$ \\
RealWorldQA~\cite{bench_realworldqa} & 59.08 & 55.69 & \textbf{62.74} & $-7.05$ \\
\rowcolor{gray!8} \textit{Category Avg.} & 61.28 & \textbf{64.37} & 63.73 & $+0.64$ \\

\midrule
\multicolumn{5}{@{}l}{\textbf{\textit{Reasoning}}} \\
MathVista-Mini~\cite{bench_mathvista} & 47.10 & \textbf{53.50} & 47.20 & $+6.30$ \\
WeMath-Loose~\cite{bench_wemath} & 33.90 & \textbf{54.86} & 42.57 & $+12.29$ \\
MathVision-Mini~\cite{bench_mathvision} & 22.70 & \textbf{31.58} & 30.59 & $+0.99$ \\
MathVision-Test~\cite{bench_mathvision} & 25.13 & \textbf{38.26} & 31.22 & $+7.04$ \\
MathVerse~\cite{bench_mathverse} & 34.31 & \textbf{45.48} & 33.47 & $+12.01$ \\
MMMU-Val~\cite{bench_mmmu} & 47.11 & \textbf{54.22} & 47.00 & $+7.22$ \\
DynaMath-Worst~\cite{bench_dynamath} & 16.37 & \textbf{22.36} & 14.37 & $+7.99$ \\
MMMU-Pro-Standard~\cite{bench_mmmu_pro} & 32.43 & \textbf{39.25} & 31.39 & $+7.86$ \\
MMMU-Pro-Vision~\cite{bench_mmmu_pro} & 25.72 & \textbf{32.49} & 22.95 & $+9.54$ \\
\rowcolor{gray!8} \textit{Category Avg.} & 31.64 & \textbf{41.33} & 33.42 & $+7.92$ \\

\midrule
\multicolumn{5}{@{}l}{\textbf{\textit{OCR \& Chart}}} \\
CharXiv-Description~\cite{bench_charxiv} & 72.13 & \textbf{74.48} & 47.18 & $+27.30$ \\
CharXiv-Reason~\cite{bench_charxiv} & 25.30 & \textbf{29.20} & 23.80 & $+5.40$ \\
ChartQA~\cite{bench_chartqa} & 60.92 & \textbf{72.84} & 71.08 & $+1.76$ \\
OCRBench~\cite{bench_ocrbench} & \textbf{63.40} & 61.20 & 57.60 & $+3.60$ \\
\rowcolor{gray!8} \textit{Category Avg.} & 55.44 & \textbf{59.43} & 49.92 & $+9.52$ \\

\midrule
\multicolumn{5}{@{}l}{\textbf{\textit{Others}}} \\
AI2D~\cite{bench_ai2d} & 65.90 & 71.99 & \textbf{73.84} & $-1.85$ \\
PixmoCount~\cite{deitke2024molmo} & 38.01 & \textbf{41.76} & 27.15 & $+14.61$ \\
\rowcolor{gray!8} \textit{Category Avg.} & 51.96 & \textbf{56.88} & 50.50 & $+6.38$ \\

\midrule\midrule
\textbf{Overall Average} & 48.41 & \textbf{54.28} & 48.96 & \textbf{$+5.33$} \\
\textbf{Win Count (vs. QS RL)} & -- & \textbf{19 / 24} & 5 / 24 & -- \\
\textbf{Optimal RL Steps} & -- & \textbf{2,175} & 4,415 & \textit{-50.7\% steps} \\
\bottomrule
\end{tabular}
\end{adjustbox}
\end{table}

\subsection{Zero-Vision SFT: Unlocking Agentic Capabilities}
\label{subsec:zero_vision_sft}

\paragraph{Motivation: Rethinking Visual Post-Training.}
Despite its strong architectural foundation, our primary model, Mage-VL, currently exhibits a capability gap on complex agentic tasks compared to Qwen3-VL. This gap largely stems from computational resource constraints, the lack of joint text-multimodal mixed pre-training, and omitted RL post-training. To address these limitations under tight compute budgets, two prevailing assumptions often act as bottlenecks: (1) explicit visual-language SFT (Image SFT) is universally considered indispensable before post-training, and (2) Reinforcement Learning (RL) is widely believed to be ineffective on small-data VLM regimes (such as the LLaVA-OV scale). Contrary to these common beliefs, we hypothesize that explicit visual SFT during the instruction-tuning phase may actually induce catastrophic forgetting of intrinsic textual reasoning capabilities, thereby creating a performance ceiling for subsequent RL exploration. Drawing inspiration from Kimi-2.5~\cite{team2026kimi}, we explore whether replacing visual SFT with high-quality, pure-text reasoning instruction tuning post-midtraining can unlock potent multimodal RL capabilities.

\paragraph{Experimental Pipeline and Fine-Grained Empirical Results.}
We validate our hypothesis using the open-source \textbf{LLaVA-OV-1.5 Quick Start dataset}. Training all models from scratch, we compare two distinct pipelines. The standard baseline follows a three-stage workflow: foundational midtraining on LLaVA-OV-1.5 Quick Start data (Stage 1), standard visual instruction tuning (Stage 2 SFT), and OpenMMReasoner RL~\cite{zhang2026openmmreasoner} (Stage 3). In contrast, our proposed \textit{Zero-Vision SFT (+ RL)} strategy bypasses visual SFT entirely: in Stage 1, we combine LLaVA-OV-1.5 Quick Start midtraining with pure-text math and code instruction data from \texttt{nvidia/Nemotron-Pretraining-SFT-v1}~\cite{nvidia_nemotron_pretraining_sft_v1} under an equivalent token budget to standard visual SFT, before directly applying OpenMMReasoner RL (Stage 2 RL) without any visual instruction tuning.

As detailed in Table~\ref{tab:sft_init_rl_optimized}, replacing visual SFT with pure-text Zero-Vision SFT in Stage 1 prior to multimodal RL yields impressive performance gains across 24 multimodal image benchmarks, achieving an overall average accuracy of \textbf{54.28\%} vs. \textbf{48.96\%} ($+5.33\%$ absolute margin) and winning in \textbf{19 out of 24} tasks. Crucially, Zero-Vision RL reaches optimal convergence at only \textbf{2,175 steps}, reducing required RL iterations by over \textbf{50.7\%} compared to the Quick Start baseline (4,415 steps).

Analyzing category-level metrics demonstrates that preserving pure-text reasoning capabilities pays off significantly. In \textbf{Reasoning} tasks ($+7.92\%$ avg. gain), Zero-Vision RL surges across all 9 benchmarks, most notably on WeMath-Loose (\textbf{54.86\%} vs. 42.57\%, $+12.29\%$), MathVerse (\textbf{45.48\%} vs. 33.47\%, $+12.01\%$), and MMMU-Pro-Vision (\textbf{32.49\%} vs. 22.95\%, $+9.54\%$). For \textbf{OCR \& Chart} understanding ($+9.52\%$ avg. gain), direct multimodal RL effectively aligns textual logic with dense visual structures, yielding massive jumps on CharXiv-Desc (\textbf{74.48\%} vs. 47.18\%, $+27.30\%$) and ChartQA (\textbf{72.84\%} vs. 71.08\%). Even in \textbf{General VQA} ($+0.64\%$ avg. gain) and \textbf{Others} ($+6.38\%$ avg. gain), direct RL grounds perceptual features effectively without visual SFT, achieving noticeable improvements on MMStar (\textbf{54.80\%} vs. 50.33\%), CV-Bench (\textbf{73.88\%} vs. 69.14\%), and counting tasks like PixmoCount (\textbf{41.76\%} vs. 27.15\%, $+14.61\%$).

\begin{promptbox}
\large
\textbf{Finding 7:}
\textbf{Bypassing Visual SFT Unlocks Multimodal RL and Bridges the Agentic Gap.}
Challenging the common belief that visual SFT is essential and that RL fails on small-data VLMs, replacing image SFT with high-quality text SFT unlocks potent multimodal RL capabilities. It provides a compute-efficient technical path to address current agentic limitations.
\end{promptbox}

\section{Conclusion and Limitations}
\label{sec:conclusion}
We presented \magevl, a lightweight 4B-parameter streaming vision--language model centered on its custom-designed ViT-encoder, \magevit. Built upon the principle of codec-aligned sparsity, \magevit is trained from scratch as a codec-agnostic visual front-end that dynamically extracts motion- and entropy-rich patches across traditional (e.g., HEVC) and neural codecs (e.g., DCVC). Integrated with a causal language decoder and an event-driven System 1 gate, \magevl provides a unified architecture for offline multimodal reasoning, fine-grained spatial grounding, and proactive streaming interaction.

Empirically, \magevl-4B matches flagship baselines on static image understanding while achieving uniform improvements across video QA, temporal grounding, and 2D/3D spatial reasoning benchmarks, accompanied by up to $3.5\times$ wall-clock inference speedups over dense frame sampling. Furthermore, our systematic empirical studies yield seven foundational findings, covering visual pre-training data efficiency, resolution scaling, codec-driven system acceleration, VideoQA SFT redundancy, motion-spatial synergy, AI-driven data pipeline optimization, and Zero-Vision SFT for multimodal RL.

\paragraph{Limitations and Future Work.}
Despite its strong visual and temporal performance, \magevl currently exhibits limitations in complex agentic workflows and mathematical reasoning. These gaps primarily stem from a shortage of high-quality text data in our current training mixture, as well as the omission of RL post-training. In future iterations, we plan to adopt a Zero-Vision SFT alongside RL to bridge these agentic and mathematical capability gaps, while extending \magevl to native audio-visual stream fusion and embodied edge applications.

\section*{Contributor List}
\label{sec:contributor}

\noindent\textbf{Contributors:} Senqiao Yang$^{*}$, Kaichen Zhang$^{*}$, Zhaoyang Jia$^{*}$, Jinghao Guo$^{*}$, Yifei Shen$^{*\dagger}$, Xinjie Zhang$^{*\dagger}$, Xiaoyi Zhang, Haoqing Wang, Xiao Li, Peng Zhang, Xiang An, Yin Xie, Zhening Liu, Xun Guo, Jiahao Li, Shicheng Zheng, Jinglu Wang, Zongyu Guo, Wenxuan Xie, Zihan Zheng, Yuxuan Luo, Bin Li, Yan Lu

\vspace{0.5em}
\noindent{\footnotesize $^{*}$Equal Contribution. \quad $^{\dagger}$Project Leads (yshenaw@connect.ust.hk, xinjiezhang@microsoft.com).}

\clearpage
\appendix

\section{Recaptioning System Prompt}
\label{app:caption_prompt}

This appendix provides the complete system prompt used for large-scale image recaptioning with Qwen3-VL-32B. The original prompt of LLaVA-OV is following:
\begin{CJK*}{UTF8}{gbsn}

\begin{tcblisting}{
  enhanced,
  breakable,
  colback=promptbg,
  colframe=promptborder,
  arc=3pt,
  boxrule=0.8pt,
  left=8pt,
  right=8pt,
  top=4pt,
  bottom=4pt,
  fonttitle=\bfseries\sffamily,
  title={LLaVA-OneVision captioning system prompt},
  listing only,
  listing engine=listings,
  listing options={
    basicstyle=\fontsize{6pt}{7pt}\selectfont\ttfamily,
    breaklines=true,
    breakatwhitespace=false,
    columns=fullflexible,
    keepspaces=true,
    showstringspaces=false,
    tabsize=2,
    escapeinside={(*@}{@*)}
  }
}
Generate a single coherent, detailed caption for the image that covers all applicable dimensions below.
Please integrate all relevant dimensions into one comprehensive caption rather than addressing each dimension separately.
Ensure the description is visually grounded - every detail should be verifiable from the image itself.


Pre-defined Dimensions:

0. **World Knowledge**:
Provide relevant factual background related to the image content (historical facts, cultural significance, scientific context).
Use specific names, places, and concepts rather than vague references.
Example: "The Eiffel Tower" instead of "a famous landmark".

1. **Character Name**:
If identifiable characters appear from movies, TV shows, anime, comics, literature, games, or virtual idols, state their names.

2. **Scene Description**:
Provide an overview of the image, identifying key objects, people, and interactions.
Classify and describe each element with specific attributes: size, color, material, texture.

3. **Actions and Interactions**:
Describe actions taking place and how people/objects interact.
Detail dynamic elements (e.g., running, jumping, flying, waving) and their state.

4. **Context and Environment**:
Describe the setting: location (indoor/outdoor), time of day, weather, background elements.
Explain how the environment contributes to the overall scene and mood.

5. **Emotion and Sentiment**:
If people are present, describe emotional states based on body language and facial expressions.
What mood or tone does the image convey (e.g., happiness, sadness, tension, peace)?

6. **Relationships and Spatial Arrangement**:
Explain how objects and people are positioned relative to one another (next to, above, to the right of).
Consider foreground, background, and overall spatial composition.

7. **Color and Texture**:
Describe the color palette and texture details (smooth, rough, soft).
How do color and texture contribute to the atmosphere or style?

8. **Symbolism or Abstract Interpretation**:
If relevant, interpret symbolic or abstract elements.
What deeper meanings or metaphors can be inferred? How do they relate to broader themes?

9. **Lighting and Shadows**:
Observe lighting conditions (sunlight, artificial light) and shadow/reflection patterns.
Note light intensity and its contribution to mood or focal points.

10. **Details and Fine Elements**:
Focus on intricate details (wrinkles in clothing, surface textures, distinct features).
These elements may carry significant meaning or provide vivid precision.

11. **Perspective and Composition**:
Describe viewpoint (aerial, eye-level, side view) and composition (symmetry, balance, focal point).
How does perspective affect viewer perception?

12. **Time and Season**:
Infer time of day or season from visual cues (light quality, weather, clothing).
Examples: winter snow scene, summer beach, autumn forest.

13. **Target Audience**:
Consider if the image suggests a specific target audience or purpose.
Adjust complexity and terminology accordingly (e.g., technical vs. general audience).

14. **OCR (Text in Image)**:
If text appears, transcribe it in the original language and provide English translation in parentheses.
Example: "(*@书本@*) (book)". Explain the meaning within context.

15. **Person Description**:
Describe physical features (age, gender, hairstyle, clothing), movements, and expressions.
Explain relationship to the surrounding environment.
Use singular pronouns (he/she) for single individuals rather than "they".

16. **Mathematics**:
Describe mathematical concepts: geometric shapes, equations, numeric values, relationships.
For graphs, describe axes, scales, and key points.
Explain how mathematical operations are visualized.

17. **Information Extraction**:
Extract textual and contextual information.
For documents, transcribe content accurately.
For GUI or structured data, describe layout, labels, and functionality.

18. **Planning**:
Identify sequences or logical arrangements.
For processes, explain steps in correct order.
For puzzles or games, provide rules and possible solutions.

19. **Coding**:
If code appears, output it line by line first.
Explain syntax, function, and purpose.
For debugging tasks, identify issues or provide equivalent code in other languages.

20. **Perception**:
Provide detailed perception-based descriptions.
Identify object attributes (color, shape, size) and spatial relationships.
For facial analysis or pose estimation, include expressions, poses, and physical traits.

21. **Metrics**:
Evaluate image quality, authenticity, and adherence to caption content.
For comparative tasks, provide constructive feedback or preference reasoning.

22. **Science**:
Explain scientific content or phenomena depicted.
Include details on experiments, natural phenomena, or theoretical concepts with relevant terminology.


Output Requirements:
- Generate a single, coherent caption integrating all applicable dimensions
- Do NOT format as separate sections (e.g., **Scene Description**, **Actions**, etc.)
- Omit dimensions that are not applicable to the image
- Maintain a detailed, comprehensive description
- Ensure all details are visually verifiable from the image
\end{tcblisting}

\end{CJK*}

The prompt is obtained through the iterative optimization procedure described in \cref{sec:image_data}, with the goal of improving visual coverage, reducing redundant descriptions, maintaining coherent organization, and preserving rendered text with high OCR fidelity. The final prompt is used to generate the approximately 350M image--caption pairs employed in our image-caption training stage.
\begin{tcblisting}{
  enhanced,
  breakable,
  colback=promptbg,
  colframe=promptborder,
  arc=3pt,
  boxrule=0.8pt,
  left=8pt,
  right=8pt,
  top=4pt,
  bottom=4pt,
  fonttitle=\bfseries\sffamily,
  title={Optimized image captioning system prompt},
  listing only,
  listing engine=listings,
  listing options={
    basicstyle=\fontsize{6pt}{7pt}\selectfont\ttfamily,
    breaklines=true,
    breakatwhitespace=false,
    columns=fullflexible,
    keepspaces=true,
    showstringspaces=false,
    tabsize=2,
    escapeinside={(*@}{@*)}
  }
}
Write a single, detailed, and strictly objective caption for this image. Describe only what is directly visible. Be assertive and precise--state facts, not possibilities.

Writing style: Write in natural, flowing prose as a single coherent paragraph. Integrate details organically into sentences rather than listing attributes one by one. Use varied sentence structures--combine related details within the same sentence using clauses, appositives, and conjunctions. Transition smoothly between subjects (e.g., from foreground to background, from one object to the next) so the description reads like a cohesive narrative, not a feature inventory.

BAD (mechanical listing): "The dog has brown fur. The dog has floppy ears. The dog's eyes are dark. The dog is sitting. The dog is on a wooden floor."
GOOD (flowing prose): "A brown dog with floppy ears and dark eyes sits on a wooden floor, its front paws resting side by side."

BAD (repetitive structure): "The shirt is blue. The pants are black. The shoes are white. The hat is red."
GOOD (integrated): "The person wears a blue shirt tucked into black pants, paired with white shoes and a red hat."

Cover the following aspects where applicable (do NOT use headers, bullet points, or labels):

- **Objects & Attributes**: Key objects, people, and elements with observable attributes: color, shape, size, material, texture, quantity.
- **Fine Details**: Intricate or small-scale details that add precision--wrinkles in fabric, scratches on surfaces, visible stitching, watermarks, tags, labels, logos, serial numbers, or distinctive marks.
- **Text & Data (OCR)**: Transcribe any visible text exactly as it appears in its original language. If the text is not in English, provide an English translation in parentheses immediately after (e.g., "Libro (book)", "Sortie (Exit)"). Briefly explain the meaning or role of the text within the context of the image (e.g., a sign, a label, a title, a watermark). Reproduce key data from tables, charts, and labels accurately, including column headers, row labels, and numerical values.
- **Documents & GUI**: For screenshots, documents, or structured interfaces, describe the layout, navigation elements, buttons, menus, input fields, and content hierarchy. Transcribe visible text content accurately.
- **Spatial Layout**: Positions, arrangements, and spatial relationships between elements. State each spatial relationship only once.
- **Actions & Poses**: Observable body positions and actions factually (e.g., "mouth open, eyes wide" not "expressing surprise"). Never infer emotions, intentions, or mental states.
- **People**: Describe age range, gender, hairstyle, clothing, posture, and physical features factually. For facial expressions, state only the physical configuration (e.g., "corners of the mouth turned upward") without labeling emotions. Use singular pronouns (he/she) for single individuals rather than "they."
- **Setting & Environment**: Indoor/outdoor, background elements, weather, time of day, and season only when indicated by visible cues (e.g., snow on ground -> winter; long shadows and orange sky -> late afternoon).
- **Color Palette & Lighting**: Colors, lighting direction/type (sunlight, artificial, diffused), shadow patterns, and reflections factually.
- **Perspective & Composition**: Viewpoint (top-down, eye-level, side view, aerial) and compositional structure (symmetry, focal point, leading lines, depth of field).
- **Identifiable Entities**: Use specific names for recognizable people, places, brands, characters, or landmarks. Always prefer precise identification over generic descriptions.
  Examples: "The Eiffel Tower" instead of "a tall metal tower"; "Pikachu" instead of "a yellow cartoon creature"; "a Nike swoosh logo" instead of "a checkmark-shaped logo"; "Barack Obama" instead of "a man in a suit."
  This applies to fictional characters from anime, comics, movies, TV shows, games, and virtual idols--state their names when recognizable.
- **Sequential & Multi-panel Content**: For images showing step-by-step processes, numbered panels, comic strips, or before/after comparisons, describe each step or panel in order, noting panel numbers or labels if present.
- **Code & Programming**: If code is visible, transcribe it line by line. Describe the programming language, syntax structure, and visible function/variable names. Note any syntax highlighting colors.
- **Mathematics & Graphs**: For equations, render them in LaTeX. For graphs and charts, describe axes (labels, units, scale), data points, trends, and legends. For geometric figures, describe shapes, angles, and measurements.
- **Scientific Content**: For scientific images (experiments, specimens, diagrams, phenomena), use appropriate technical terminology. Describe apparatus, measurements, labels, and observable processes.

Strict rules--violation of any rule makes the caption unacceptable:

1. NEVER start with "The image" in any form (e.g., "The image shows/displays/features/presents/depicts/is/contains"). Instead, begin directly with the subject: "A dog...", "Two women...", "An aerial view of...", "A close-up of...".
2. NEVER use any form of these words anywhere: "suggest/suggesting/suggests", "indicate/indicating/indicates", "imply/implying/implies", "evoke/evoking/evokes", "convey/conveying/conveys", "hint/hinting/hints at", "allude/alluding/alludes". State the visual fact directly without inferring purpose or meaning.
   BAD: "green leaves suggesting spring" -> GOOD: "green leaves"
   BAD: "open mouth indicating speech" -> GOOD: "open mouth"
   BAD: "a pattern suggesting movement" -> GOOD: "a curved, flowing pattern"
   BAD: "shadows suggesting depth" -> GOOD: "shadows fall across the surface"
   BAD: "colors hinting at sunset" -> GOOD: "orange and pink tones in the sky"
3. NEVER use "creating", "creates", or "created" to describe visual effects or impressions. Use concrete verbs instead: casts, produces, forms, results in, yields, adds, gives, lends.
   BAD: "red light creating a warm tone" -> GOOD: "red light casts a warm-toned glow"
   BAD: "lines creating depth" -> GOOD: "lines converge toward a vanishing point"
   BAD: "contrast creating emphasis" -> GOOD: "the high contrast between the dark background and bright subject draws focus"
   BAD: "bokeh creating a soft background" -> GOOD: "bokeh dots fill the out-of-focus background"
   BAD: "shadows creating texture" -> GOOD: "shadows accentuate the surface texture"
4. NEVER use "appears to", "appears", "seems to", "seems", "likely", "possibly", "presumably", "probably", or "perhaps"--state what IS visible as fact.
   BAD: "The surface appears rough" -> GOOD: "The surface is rough"
   BAD: "appears to be a cat" -> GOOD: "a cat"
5. NEVER use "overall" anywhere in the caption. Do not summarize the scene at the end; instead, continue describing visible details.
6. NEVER add emotional, atmospheric, or aesthetic commentary. Banned words include: "warm" (except for color temperature like "warm-toned lighting"), "serene", "calm", "peaceful", "gentle", "intimate", "heartfelt", "elegant", "cozy", "dramatic", "striking", "vibrant mood", "whimsical", "enchanting", "playful", "inviting", "charming", "delicate", "lively", "cheerful", "magical", "dynamic" (except physics), "graceful", "sense of", "gives a feeling of", "atmosphere", "mood".
7. NEVER describe what is absent. Only describe what IS present and visible. Banned phrases: "no visible", "no text", "no people", "no other", "no additional", "no distinct", "no further", "nothing else", "without any", "lacks". Simply omit anything not present.
   BAD: "No text is visible. No people are present. No other objects are seen."
   BAD: "no additional decorations or objects in the visible area"
   BAD: "no distinct shadows are cast"
   GOOD: (simply omit--say nothing about absent elements)
8. NEVER summarize or restate information already described. Each fact appears exactly once.
9. Every statement must be visually verifiable from the image alone.
10. Omit any aspect that does not apply.
11. Do NOT write in a mechanical, checklist style. Avoid sequences of short, isolated sentences that each describe one attribute. Instead, merge related details into richer, compound sentences.

Before finalizing, perform this mandatory self-check word by word:
- Scan for: "suggesting", "indicating", "implying", "hinting", "creating", "creates" -> replace with concrete verbs
- Scan for: "appears", "seems", "likely", "possibly", "overall" -> delete or rewrite as fact
- Scan for: "no visible", "no other", "no additional", "no distinct" -> delete the entire sentence
- Scan for: "warm", "gentle", "calm", "delicate", "striking", "elegant" -> delete the adjective
- Check the first word: if "The image" -> rewrite to start with the subject
- Check for 3+ consecutive short sentences -> combine them

Examples:

Example 1 (Table image):
A table titled "Table 2. Variation in content of lipids, EPA and DHA in livers from cod fish during 21 months" spans the full width of the frame against a white background, with the title set in bold black serif text centered above the grid. The header row is filled in solid red with three column labels in white bold text--"Total lipid content (g/100g)", "EPA (% of total fatty acids)", and "DHA (% of total fatty acids)"--separated by thin black vertical borders. Five data rows sit below, each beginning with a fish species name in the leftmost column: Ling, Saithe, Hake, Cod, and Monkfish, followed by their corresponding numerical ranges--Ling: 51-66, 5.1-7.7, 10.2-12.9; Saithe: 48-76, 4.4-9.0, 10.5-13.3; Hake: 42-67, 5.5-7.7, 11.7-13.0; Cod: 50-64, 6.0-8.6, 12.3-14.4; Monkfish: 31-43, 4.6-6.3, 10.8-13.1. All numerical values are center-aligned in black regular-weight serif font, and thin black borders outline every cell. The top-down perspective captures the complete table with uniform row heights and consistent cell spacing.

Example 2 (Outdoor scene with people):
A crowded outdoor market extends along a narrow cobblestone street, with vendor stalls on both sides sheltered under canvas tarps in faded blue, green, and beige. In the foreground, a woman wearing a long olive-green coat and brown leather boots reaches toward a pile of red and yellow apples arranged in tiered wooden crates, while the vendor--a man in a gray wool cap and dark blue apron over a plaid flannel shirt--extends a small brown paper bag toward her with his right hand. To her left, a child in a red puffy jacket and jeans holds a half-eaten pretzel in one hand and grips the woman's coat with the other, standing at hip height beside her. Further down the street, clusters of shoppers move between stalls stacked with root vegetables, hanging sausages, and loaves of bread on wire racks, their forms becoming progressively blurred with distance. Strings of bare Edison-style light bulbs hang in parallel rows above the stalls at roughly three-meter intervals, and late-afternoon sunlight enters from the upper right, casting long shadows to the left across the cobblestones and illuminating the stone facades of three- and four-story residential buildings lining both sides of the street. The eye-level perspective is centered in the street, with the converging lines of stalls and overhead lights leading toward a clock tower with a copper-green spire at the far end.

Example 3 (Animal close-up):
A chimpanzee with dense black fur sits on a thick horizontal tree branch, its left hand gripping the rough bark while its right hand holds a partially peeled banana near its open mouth, the lower lip pulled down to reveal a row of flat white teeth. The chimp's dark gray face is creased with deep wrinkles around the eyes, and it has a broad, flat nose with flared nostrils, small rounded ears set high on the head and partially obscured by tufts of coarse fur, and sparse white hairs along its chin and lower jaw. Its dark brown eyes, with visible round pupils, are directed downward toward the banana. Behind the chimpanzee, layers of out-of-focus green foliage--broad tropical leaves, thin vertical stems, and patches of brown bark from adjacent trees--fill the frame in varying shades of emerald and olive. A thin shaft of sunlight cuts diagonally from the upper left through the canopy, catching the fur along the chimp's right shoulder in a bright highlight while the rest of the body remains in dappled shade. The eye-level perspective places the chimpanzee slightly left of center, with the branch running horizontally across the lower third of the frame and the dense canopy occupying the upper two-thirds.

Now write the caption for the provided image:
\end{tcblisting}

\begin{tcblisting}{
  enhanced,
  breakable,
  colback=promptbg,
  colframe=promptborder,
  arc=3pt,
  boxrule=0.8pt,
  left=8pt,
  right=8pt,
  top=4pt,
  bottom=4pt,
  fonttitle=\bfseries\sffamily,
  title={Video captioning system prompt},
  listing only,
  listing engine=listings,
  listing options={
    basicstyle=\fontsize{6pt}{7pt}\selectfont\ttfamily,
    breaklines=true,
    breakatwhitespace=false,
    columns=fullflexible,
    keepspaces=true,
    showstringspaces=false,
    tabsize=2,
    escapeinside={(*@}{@*)}
  }
}
# Video Description Prompt & Requirements

Please describe the video in detail, including but not limited to the user pre-defined dimensions. Please make sure your description is visually grounded for the user-provided video, namely the user can find visual cues in the video for your generated video caption.

> **IMPORTANT:** Your response must be in English plain text format only. Do not use any markdown formatting such as `**bold**`, `*italic*`, `### headers`, or any other markdown symbols. Use only regular text without any special formatting characters.

---

## User Pre-defined Dimensions

### 0. Context and Environment
Describe the setting of the image, including:
* **Location:** Indoor or outdoor
* **Time & Weather:** Time of day, weather conditions
* **Background Elements:** Sky, buildings, roads, etc.
* **Impact:** How the environment contributes to the overall scene and whether the setting enhances the mood or theme.

### 1. Main Subject of the Video
Identify the primary subject of the video (e.g., a male athlete, a running lion, a plane taking off, a flying bird, a swimming fish, a female dancer performing a dance). 
* *Note:* If the main subject is a character, you **must** specify his or her name (including characters from movies, TV shows, anime, comics, literature, video games, virtual idols, or virtual characters).

### 2. Actions and Interactions
Describe any named actions occurring in the video (e.g., butterfly stroke, pole vault, jazz dance). Specify who is performing these actions and how they interact with other objects or people.

### 3. Motion Detail Description
For dynamic elements in the video:
* **State:** Running, jumping, flying, waving, swimming, cooking, etc.
* **Direction of Motion:** Left to right, front to back, clockwise/counterclockwise rotation, etc.
* **Speed of Movement:** Sudden acceleration, gradual deceleration, etc.
* **Sequence of Actions:** Step-by-step chronological order (e.g., *first open the refrigerator, then take out the food, next wash the food, and finally cook it*).

### 4. Background Changes
If the environment or scene in the video changes, provide a detailed description of how the background evolves over time.
* *Example 1:* At around the 30-second mark in the latter half of the video, the weather shifts from sunny to rainy.
* *Example 2:* Around the 20-second mark in the middle of the video, the street gradually transitions from noisy and crowded to quiet and empty.
* *Example 3:* At approximately 50 seconds, gentle ripples begin to appear on an otherwise calm sea surface, escalating by 80 seconds into turbulent, stormy waves.

### 5. Highlight Moments
If the video contains significant or impactful events (e.g., a soccer goal, a person falling, a tiger hunting prey, a boxer knocking down an opponent), specify the exact time each event occurs and provide a detailed description.
* *Example 1:* At approximately 15 seconds, the player wearing a white jersey, number 5, kicks the soccer ball into the goal.
* *Example 2:* At around 3 seconds, the boxer in red shorts knocks down his opponent, who is dressed in black.

### 6. First-Person Perspective
If the video is shot from a first-person point of view (POV):
* Clearly state the main activity being performed (e.g., cooking, painting, playing soccer).
* Provide a detailed chronological description of the subject's actions and their timing (e.g., *at around 2 seconds, the viewer opens a laptop; at 5 seconds, they navigate to a video website; at 8 seconds, they open a can of beverage*).

---

## Response Generation Rules

* **Output Style:** Do **not** generate your response split by dimension (e.g., do not write `**Context and Environment**: ...`). Provide **one single overall image/video caption** that naturally integrates all dimensions into a cohesive text.

### Critical Formatting Requirements

1. **Structure:** The format of your answer should be a **single continuous paragraph**.
2. **Plain Text Only:** Use **ONLY plain text** -- strictly NO markdown formatting whatsoever (no `**`, `*`, `###`, etc.).
3. **No Special Symbols:** Do not use any special characters for emphasis or formatting.
4. **Tone & Flow:** Write in natural, flowing sentences without any markup symbols.
\end{tcblisting}

\newpage
\FloatBarrier  
\bibliography{references}

\begin{thebibliography}{177}
\providecommand{\natexlab}[1]{#1}
\providecommand{\url}[1]{\texttt{#1}}
\expandafter\ifx\csname urlstyle\endcsname\relax
  \providecommand{\doi}[1]{doi: #1}\else
  \providecommand{\doi}{doi: \begingroup \urlstyle{rm}\Url}\fi

\bibitem[{Anthropic}(2026)]{anthropic2026claude5}
{Anthropic}.
\newblock Claude fable 5 \& claude mythos 5 system card.
\newblock Technical report, Anthropic, 6 2026.
\newblock URL \url{https://www.anthropic.com/system-cards}.

\bibitem[{Google DeepMind}(2025)]{deepmind2025gemini3pro}
{Google DeepMind}.
\newblock Gemini 3 pro model card.
\newblock Technical report, Google DeepMind, 11 2025.
\newblock URL \url{https://storage.googleapis.com/deepmind-media/Model-Cards/Gemini-3-Pro-Model-Card.pdf}.

\bibitem[{OpenAI}(2026)]{openai2026gpt55}
{OpenAI}.
\newblock Gpt-5.5 system card.
\newblock Technical report, OpenAI, 4 2026.
\newblock URL \url{https://openai.com/index/gpt-5-5-system-card/}.

\bibitem[Moravec(1988)]{moravec1988mind}
Hans Moravec.
\newblock \emph{Mind Children: The Future of Robot and Human Intelligence}.
\newblock Harvard University Press, 1988.

\bibitem[Bai et~al.(2025{\natexlab{a}})Bai, Cai, Chen, Chen, Chen, Cheng, Deng, Ding, Gao, Ge, Ge, Guo, Huang, Huang, Huang, Hui, Jiang, Li, Li, Li, Li, Lin, Lin, Liu, Liu, Liu, Liu, Liu, Liu, Lu, Luo, Lv, Men, Meng, Ren, Ren, Song, Sun, Tang, Tu, Wan, Wang, Wang, Wang, Wang, Xie, Xu, Xu, Xu, Yang, Yang, Yang, Yang, Yu, Zhang, Zhang, Zhang, Zheng, Zhong, Zhou, Zhou, Zhou, Zhu, and Zhu]{qwen3vl}
Shuai Bai, Yuxuan Cai, Ruizhe Chen, Keqin Chen, Xionghui Chen, Zesen Cheng, Lianghao Deng, Wei Ding, Chang Gao, Chunjiang Ge, Wenbin Ge, Zhifang Guo, Qidong Huang, Jie Huang, Fei Huang, Binyuan Hui, Shutong Jiang, Zhaohai Li, Mingsheng Li, Mei Li, Kaixin Li, Zicheng Lin, Junyang Lin, Xuejing Liu, Jiawei Liu, Chenglong Liu, Yang Liu, Dayiheng Liu, Shixuan Liu, Dunjie Lu, Ruilin Luo, Chenxu Lv, Rui Men, Lingchen Meng, Xuancheng Ren, Xingzhang Ren, Sibo Song, Yuchong Sun, Jun Tang, Jianhong Tu, Jianqiang Wan, Peng Wang, Pengfei Wang, Qiuyue Wang, Yuxuan Wang, Tianbao Xie, Yiheng Xu, Haiyang Xu, Jin Xu, Zhibo Yang, Mingkun Yang, Jianxin Yang, An~Yang, Bowen Yu, Fei Zhang, Hang Zhang, Xi~Zhang, Bo~Zheng, Humen Zhong, Jingren Zhou, Fan Zhou, Jing Zhou, Yuanzhi Zhu, and Ke~Zhu.
\newblock {Qwen3-VL} technical report.
\newblock \emph{arXiv preprint arXiv:2511.21631}, 2025{\natexlab{a}}.

\bibitem[Bai et~al.(2025{\natexlab{b}})Bai, Chen, Liu, Wang, Ge, Song, Dang, Wang, Wang, Tang, et~al.]{bai2025qwen25vl}
Shuai Bai, Keqin Chen, Xuejing Liu, Jialin Wang, Wenbin Ge, Sibo Song, Kai Dang, Peng Wang, Shijie Wang, Jun Tang, et~al.
\newblock {Qwen2.5-VL} technical report.
\newblock \emph{arXiv:2502.13923}, 2025{\natexlab{b}}.

\bibitem[Zhu et~al.(2025)Zhu, Wang, Chen, Liu, Ye, Gu, Duan, Tian, Su, Shao, Gao, Cui, Cao, Liu, Xu, Li, Wang, Lv, Chen, Li, He, Jiang, Luo, Wang, He, Shi, Zhang, Shao, He, Xiong, Qu, Sun, Jiao, Wu, Zhang, Deng, Ge, Chen, Wang, Dou, Lu, Zhu, Lu, Lin, Qiao, Dai, and Wang]{internvl3}
Jinguo Zhu, Weiyun Wang, Zhe Chen, Zhaoyang Liu, Shenglong Ye, Lixin Gu, Yuchen Duan, Hao Tian, Weijie Su, Jie Shao, Zhangwei Gao, Erfei Cui, Yue Cao, Yangzhou Liu, Weiye Xu, Hao Li, Jiahao Wang, Han Lv, Dengnian Chen, Songze Li, Yinan He, Tan Jiang, Jiapeng Luo, Yi~Wang, Conghui He, Botian Shi, Xingcheng Zhang, Wenqi Shao, Junjun He, Yingtong Xiong, Wenwen Qu, Peng Sun, Penglong Jiao, Lijun Wu, Kaipeng Zhang, Huipeng Deng, Jiaye Ge, Kai Chen, Limin Wang, Min Dou, Lewei Lu, Xizhou Zhu, Tong Lu, Dahua Lin, Yu~Qiao, Jifeng Dai, and Wenhai Wang.
\newblock {InternVL3}: Exploring advanced training and test-time recipes for open-source multimodal models.
\newblock \emph{arXiv:2504.10479}, 2025.

\bibitem[Li et~al.(2025{\natexlab{a}})Li, Zhang, Guo, Zhang, Li, Zhang, Zhang, Li, Liu, and Li]{li2024llavaonevision15}
Bo~Li, Yuanhan Zhang, Dong Guo, Renrui Zhang, Feng Li, Hao Zhang, Kaichen Zhang, Yanwei Li, Ziwei Liu, and Chunyuan Li.
\newblock {LLaVA-OneVision-1.5}: A family of fully open vision--language models with the llava-onevision-1.5 mid-training and instruct data.
\newblock \emph{arXiv preprint}, 2025{\natexlab{a}}.

\bibitem[Yang et~al.(2025{\natexlab{a}})Yang, Wen, Ding, et~al.]{Keye-VL-1.5}
Biao Yang, Bin Wen, Boyang Ding, et~al.
\newblock Kwai keye-vl 1.5 technical report.
\newblock \emph{arXiv:2509.01563}, 2025{\natexlab{a}}.

\bibitem[Tong et~al.(2024{\natexlab{a}})Tong, Brown, Wu, Woo, Middepogu, Akula, Yang, Yang, Iyer, Pan, et~al.]{cambrian}
Shengbang Tong, Ellis Brown, Penghao Wu, Sanghyun Woo, Manoj Middepogu, Sai~Charitha Akula, Jihan Yang, Shusheng Yang, Adithya Iyer, Xichen Pan, et~al.
\newblock {Cambrian-1}: A fully open, vision-centric exploration of multimodal {LLMs}.
\newblock In \emph{NeurIPS}, 2024{\natexlab{a}}.

\bibitem[{Kimi Team}(2025{\natexlab{a}})]{kimi2025vl}
{Kimi Team}.
\newblock Kimi-vl technical report.
\newblock \emph{arXiv preprint arXiv:2504.07491}, 2025{\natexlab{a}}.

\bibitem[Tschannen et~al.(2025)Tschannen, Gritsenko, Wang, Naeem, Alabdulmohsin, Parthasarathy, Evans, Beyer, Xia, Mustafa, et~al.]{tschannen2025siglip}
Michael Tschannen, Alexey Gritsenko, Xiao Wang, Muhammad~Ferjad Naeem, Ibrahim Alabdulmohsin, Nikhil Parthasarathy, Talfan Evans, Lucas Beyer, Ye~Xia, Basil Mustafa, et~al.
\newblock Siglip 2: Multilingual vision-language encoders with improved semantic understanding, localization, and dense features.
\newblock \emph{arXiv preprint arXiv:2502.14786}, 2025.

\bibitem[Sterling and Laughlin(2015)]{sterling2015principles}
Peter Sterling and Simon Laughlin.
\newblock \emph{Principles of neural design}.
\newblock MIT press, 2015.

\bibitem[Niven and Laughlin(2008)]{niven2008energy}
Jeremy~E Niven and Simon~B Laughlin.
\newblock Energy limitation as a selective pressure on the evolution of sensory systems.
\newblock \emph{Journal of Experimental Biology}, 211\penalty0 (11):\penalty0 1792--1804, 2008.

\bibitem[Gollisch and Meister(2010)]{gollisch2010eye}
Tim Gollisch and Markus Meister.
\newblock Eye smarter than scientists believed: neural computations in circuits of the retina.
\newblock \emph{Neuron}, 65\penalty0 (2):\penalty0 150--164, 2010.

\bibitem[Goodale and Milner(1992)]{goodale1992separate}
Melvyn~A Goodale and A~David Milner.
\newblock Separate visual pathways for perception and action.
\newblock \emph{Trends in neurosciences}, 15\penalty0 (1):\penalty0 20--25, 1992.

\bibitem[Kahneman(2011)]{kahneman2011thinking}
Daniel Kahneman.
\newblock \emph{Thinking, fast and slow}.
\newblock Macmillan, 2011.

\bibitem[Tang et~al.(2026)Tang, An, Yan, Xie, Qin, Yang, Shen, Zhang, Li, Feng, Chen, Tan, Hu, Zhang, Li, Feng, Liu, Ge, and Deng]{ovencoder2026}
Feilong Tang, Xiang An, Yunyao Yan, Yin Xie, Bin Qin, Kaicheng Yang, Yifei Shen, Yuanhan Zhang, Chunyuan Li, Shikun Feng, Changrui Chen, Huajie Tan, Ming Hu, Manyuan Zhang, Bo~Li, Ziyong Feng, Ziwei Liu, Zongyuan Ge, and Jiankang Deng.
\newblock {OneVision-Encoder}: Codec-aligned sparsity as a foundational principle for multimodal intelligence.
\newblock \emph{arXiv preprint arXiv:2602.08683}, 2026.

\bibitem[Li et~al.(2021)Li, Li, and Lu]{li2021dcvc}
Jiahao Li, Bin Li, and Yan Lu.
\newblock Deep contextual video compression ({DCVC}).
\newblock \emph{NeurIPS}, 2021.

\bibitem[An et~al.(2026)An, Xie, Tang, Yan, Tan, Zhu, Li, Li, Liu, Deng, et~al.]{llavaonevision2}
Xiang An, Yin Xie, Feilong Tang, Yunyao Yan, Huajie Tan, Didi Zhu, Chunyuan Li, Bo~Li, Ziwei Liu, Jiankang Deng, et~al.
\newblock {LLaVA-OneVision-2}: Towards next-generation perceptual intelligence.
\newblock \emph{arXiv preprint arXiv:2605.25979}, 2026.

\bibitem[Abouelenin et~al.(2025)Abouelenin, Ashfaq, Atkinson, Awadalla, Bach, Bao, Benhaim, Cai, Chaudhary, Chen, Chen, Chen, Chen, Chen, Chen, Chen, Dai, Dai, Fan, Gao, Gao, Garg, Goswami, Hao, Hendy, Hu, Jin, Khademi, Kim, Kim, Lee, Li, Li, Liang, Lin, Lin, Liu, Liu, Lopez, Luo, Madan, Mazalov, Mitra, Mousavi, Nguyen, Pan, Perez-Becker, Platin, Portet, Qiu, Ren, Ren, Roy, Shang, Shen, Singhal, Som, Song, Sych, Vaddamanu, Wang, Wang, Wang, Wu, Xu, Xu, Yang, Yang, Yu, Zabir, Zhang, Zhang, Zhang, and Zhou]{abdin2024phi4}
Abdelrahman Abouelenin, Atabak Ashfaq, Adam Atkinson, Hany Awadalla, Nguyen Bach, Jianmin Bao, Alon Benhaim, Martin Cai, Vishrav Chaudhary, Congcong Chen, Dong Chen, Dongdong Chen, Junkun Chen, Weizhu Chen, Yen-Chun Chen, Yi-ling Chen, Qi~Dai, Xiyang Dai, Ruchao Fan, Mei Gao, Min Gao, Amit Garg, Abhishek Goswami, Junheng Hao, Amr Hendy, Yuxuan Hu, Xin Jin, Mahmoud Khademi, Dongwoo Kim, Young~Jin Kim, Gina Lee, Jinyu Li, Yunsheng Li, Chen Liang, Xihui Lin, Zeqi Lin, Mengchen Liu, Yang Liu, Gilsinia Lopez, Chong Luo, Piyush Madan, Vadim Mazalov, Arindam Mitra, Ali Mousavi, Anh Nguyen, Jing Pan, Daniel Perez-Becker, Jacob Platin, Thomas Portet, Kai Qiu, Bo~Ren, Liliang Ren, Sambuddha Roy, Ning Shang, Yelong Shen, Saksham Singhal, Subhojit Som, Xia Song, Tetyana Sych, Praneetha Vaddamanu, Shuohang Wang, Yiming Wang, Zhenghao Wang, Haibin Wu, Haoran Xu, Weijian Xu, Yifan Yang, Ziyi Yang, Donghan Yu, Ishmam Zabir, Jianwen Zhang, Li~Lyna Zhang, Yunan Zhang, and Xiren Zhou.
\newblock {Phi-4-Mini} technical report: Compact yet powerful multimodal language models via mixture-of-loras.
\newblock \emph{arXiv preprint arXiv:2503.01743}, 2025.

\bibitem[Aneja et~al.(2026)Aneja, Harrison, Joshi, LaBonte, Langford, and Salinas]{aneja2026phi4reasoningvision15btechnicalreport}
Jyoti Aneja, Michael Harrison, Neel Joshi, Tyler LaBonte, John Langford, and Eduardo Salinas.
\newblock Phi-4-reasoning-vision-15b technical report, 2026.
\newblock URL \url{https://arxiv.org/abs/2603.03975}.

\bibitem[Bergen and Adelson(1991)]{bergen1991plenoptic}
James~R Bergen and Edward~H Adelson.
\newblock The plenoptic function and the elements of early vision.
\newblock \emph{Computational models of visual processing}, 1\penalty0 (8):\penalty0 3, 1991.

\bibitem[Shen et~al.(2026{\natexlab{a}})Shen, Li, and Zhang]{shen2026skillopt}
Yifei Shen, Bo~Li, and Xinjie Zhang.
\newblock Skillopt-lite: Better and faster agent self-evolution via one line of vibe.
\newblock \emph{arXiv preprint arXiv:2607.03451}, 2026{\natexlab{a}}.

\bibitem[Dosovitskiy et~al.(2021)Dosovitskiy, Beyer, Kolesnikov, Weissenborn, Zhai, Unterthiner, Dehghani, Minderer, Heigold, Gelly, Uszkoreit, and Houlsby]{dosovitskiy2021vit}
Alexey Dosovitskiy, Lucas Beyer, Alexander Kolesnikov, Dirk Weissenborn, Xiaohua Zhai, Thomas Unterthiner, Mostafa Dehghani, Matthias Minderer, Georg Heigold, Sylvain Gelly, Jakob Uszkoreit, and Neil Houlsby.
\newblock An image is worth 16x16 words: Transformers for image recognition at scale.
\newblock In \emph{International Conference on Learning Representations (ICLR)}, 2021.

\bibitem[Radford et~al.(2021)Radford, Kim, Hallacy, Ramesh, Goh, Agarwal, Sastry, Askell, Mishkin, Clark, Krueger, and Sutskever]{radford2021clip}
Alec Radford, Jong~Wook Kim, Chris Hallacy, Aditya Ramesh, Gabriel Goh, Sandhini Agarwal, Girish Sastry, Amanda Askell, Pamela Mishkin, Jack Clark, Gretchen Krueger, and Ilya Sutskever.
\newblock Learning transferable visual models from natural language supervision.
\newblock \emph{ICML}, 2021.

\bibitem[Zhai et~al.(2023)Zhai, Mustafa, Kolesnikov, and Beyer]{zhai2023siglip}
Xiaohua Zhai, Basil Mustafa, Alexander Kolesnikov, and Lucas Beyer.
\newblock Sigmoid loss for language image pre-training.
\newblock \emph{ICCV}, 2023.

\bibitem[Oquab et~al.(2024)Oquab, Darcet, Moutakanni, Vo, Szafraniec, Khalidov, Fernandez, Haziza, Massa, El-Nouby, et~al.]{oquab2024dinov2}
Maxime Oquab, Timoth{\'e}e Darcet, Th{\'e}o Moutakanni, Huy Vo, Marc Szafraniec, Vasil Khalidov, Pierre Fernandez, Daniel Haziza, Francisco Massa, Alaaeldin El-Nouby, et~al.
\newblock {DINOv2}: Learning robust visual features without supervision.
\newblock \emph{Transactions on Machine Learning Research (TMLR)}, 2024.

\bibitem[Rao et~al.(2021)Rao, Zhao, Liu, Lu, Zhou, and Hsieh]{rao2021dynamicvit}
Yongming Rao, Wenliang Zhao, Benlin Liu, Jiwen Lu, Jie Zhou, and Cho-Jui Hsieh.
\newblock {DynamicViT}: Efficient vision transformers with dynamic token sparsification.
\newblock In \emph{NeurIPS}, 2021.

\bibitem[Meng et~al.(2022)Meng, Li, Chen, Lan, Wu, Jiang, and Lim]{meng2022adavit}
Lingchen Meng, Hengduo Li, Bor-Chun Chen, Shiyi Lan, Zuxuan Wu, Yu-Gang Jiang, and Ser-Nam Lim.
\newblock {AdaViT}: Adaptive vision transformers for efficient image recognition.
\newblock In \emph{CVPR}, 2022.

\bibitem[Bolya et~al.(2023)Bolya, Fu, Dai, Zhang, Feichtenhofer, and Hoffman]{bolya2023tome}
Daniel Bolya, Cheng-Yang Fu, Xiaoliang Dai, Peizhao Zhang, Christoph Feichtenhofer, and Judy Hoffman.
\newblock Token merging: Your {ViT} but faster.
\newblock In \emph{ICLR}, 2023.

\bibitem[Chen et~al.(2024{\natexlab{a}})Chen, Zhao, Liu, Bai, Lin, Zhou, and Chang]{chen2024fastv}
Liang Chen, Haozhe Zhao, Tianyu Liu, Shuai Bai, Junyang Lin, Chang Zhou, and Baobao Chang.
\newblock An image is worth 1/2 tokens after layer 2: Plug-and-play inference acceleration for large vision-language models.
\newblock In \emph{European Conference on Computer Vision (ECCV)}, 2024{\natexlab{a}}.

\bibitem[Shang et~al.(2025)Shang, Cai, Xu, Lee, and Yan]{shang2024prumerge}
Yuzhang Shang, Mu~Cai, Bingxin Xu, Yong~Jae Lee, and Yan Yan.
\newblock {LLaVA-PruMerge}: Adaptive token reduction for efficient large multimodal models.
\newblock In \emph{IEEE/CVF International Conference on Computer Vision (ICCV)}, 2025.

\bibitem[Bertasius et~al.(2021)Bertasius, Wang, and Torresani]{bertasius2021timesformer}
Gedas Bertasius, Heng Wang, and Lorenzo Torresani.
\newblock Is space-time attention all you need for video understanding?
\newblock In \emph{International Conference on Machine Learning (ICML)}, 2021.

\bibitem[Arnab et~al.(2021)Arnab, Dehghani, Heigold, Sun, Lu{\v{c}}i{\'c}, and Schmid]{arnab2021vivit}
Anurag Arnab, Mostafa Dehghani, Georg Heigold, Chen Sun, Mario Lu{\v{c}}i{\'c}, and Cordelia Schmid.
\newblock {ViViT}: A video vision transformer.
\newblock In \emph{IEEE/CVF International Conference on Computer Vision (ICCV)}, 2021.

\bibitem[Liu et~al.(2022)Liu, Ning, Cao, Wei, Zhang, Lin, and Hu]{liu2022videoswin}
Ze~Liu, Jia Ning, Yue Cao, Yixuan Wei, Zheng Zhang, Stephen Lin, and Han Hu.
\newblock Video {Swin} transformer.
\newblock In \emph{IEEE/CVF Conference on Computer Vision and Pattern Recognition (CVPR)}, 2022.

\bibitem[Tong et~al.(2022)Tong, Song, Wang, and Wang]{tong2022videomae}
Zhan Tong, Yibing Song, Jue Wang, and Limin Wang.
\newblock {VideoMAE}: Masked autoencoders are data-efficient learners for self-supervised video pre-training.
\newblock In \emph{Advances in Neural Information Processing Systems (NeurIPS)}, 2022.

\bibitem[Li et~al.(2023{\natexlab{a}})Li, Wang, and Jia]{li2023llamavid}
Yanwei Li, Chengyao Wang, and Jiaya Jia.
\newblock {LLaMA-VID}: An image is worth 2 tokens in large language models.
\newblock \emph{arXiv preprint arXiv:2311.17043}, 2023{\natexlab{a}}.

\bibitem[Jin et~al.(2024{\natexlab{a}})Jin, Takanobu, Zhang, Cao, and Yuan]{jin2024chatunivi}
Peng Jin, Ryuichi Takanobu, Wancai Zhang, Xiaochun Cao, and Li~Yuan.
\newblock {Chat-UniVi}: Unified visual representation empowers large language models with image and video understanding.
\newblock In \emph{IEEE/CVF Conference on Computer Vision and Pattern Recognition (CVPR)}, 2024{\natexlab{a}}.

\bibitem[Xu et~al.(2024)Xu, Gao, Gan, Chen, Lai, Gang, Kang, and Dehghan]{xu2024slowfastllava}
Mingze Xu, Mingfei Gao, Zhe Gan, Hong-You Chen, Zhengfeng Lai, Haiming Gang, Kai Kang, and Afshin Dehghan.
\newblock {SlowFast-LLaVA}: A strong training-free baseline for video large language models.
\newblock \emph{arXiv preprint arXiv:2407.15841}, 2024.

\bibitem[Song et~al.(2024)Song, Chai, Wang, Zhang, Zhou, Wu, Chi, Guo, Ye, Zhang, Lu, Hwang, and Wang]{song2024moviechat}
Enxin Song, Wenhao Chai, Guanhong Wang, Yucheng Zhang, Haoyang Zhou, Feiyang Wu, Haozhe Chi, Xun Guo, Tian Ye, Yanting Zhang, Yan Lu, Jenq-Neng Hwang, and Gaoang Wang.
\newblock {MovieChat}: From dense token to sparse memory for long video understanding.
\newblock In \emph{IEEE/CVF Conference on Computer Vision and Pattern Recognition (CVPR)}, 2024.

\bibitem[Shen et~al.(2024)Shen, Xiong, Zhao, Wu, Chen, Zhu, Liu, Xiao, Varadarajan, Bordes, Liu, Xu, Kim, Soran, Krishnamoorthi, Elhoseiny, and Chandra]{shen2024longvu}
Xiaoqian Shen, Yunyang Xiong, Changsheng Zhao, Lemeng Wu, Jun Chen, Chenchen Zhu, Zechun Liu, Fanyi Xiao, Balakrishnan Varadarajan, Florian Bordes, Zhuang Liu, Hu~Xu, Hyunwoo~J. Kim, Bilge Soran, Raghuraman Krishnamoorthi, Mohamed Elhoseiny, and Vikas Chandra.
\newblock {LongVU}: Spatiotemporal adaptive compression for long video-language understanding.
\newblock \emph{arXiv preprint arXiv:2410.17434}, 2024.

\bibitem[Li et~al.(2025{\natexlab{b}})Li, Wang, Yu, Chen, Zhu, Sun, He, Wang, and Wang]{videochatflash2025}
Xinhao Li, Yi~Wang, Jiashuo Yu, Xiangyu Chen, Yinan Zhu, Haian Sun, Yali He, Yu~Wang, and Limin Wang.
\newblock {VideoChat-Flash}: Hierarchical compression for long-context video modeling.
\newblock \emph{arXiv preprint arXiv:2501.00574}, 2025{\natexlab{b}}.

\bibitem[Wu et~al.(2018)Wu, Zaheer, Hu, Manmatha, Smola, and Kr{\"a}henb{\"u}hl]{wu2018coviar}
Chao-Yuan Wu, Manzil Zaheer, Hexiang Hu, R.~Manmatha, Alexander~J. Smola, and Philipp Kr{\"a}henb{\"u}hl.
\newblock Compressed video action recognition.
\newblock In \emph{IEEE/CVF Conference on Computer Vision and Pattern Recognition (CVPR)}, 2018.

\bibitem[Jin et~al.(2024{\natexlab{b}})Jin, Sun, Xu, Xu, Chen, Jiang, Huang, Song, Liu, Zhang, Song, Gai, and Mu]{videolavit2024}
Yang Jin, Zhicheng Sun, Kun Xu, Kun Xu, Liwei Chen, Hao Jiang, Quzhe Huang, Chengru Song, Yuliang Liu, Di~Zhang, Yang Song, Kun Gai, and Yadong Mu.
\newblock {Video-LaVIT}: Unified video-language pre-training with decoupled visual-motional tokenization.
\newblock \emph{ICML}, 2024{\natexlab{b}}.

\bibitem[Alayrac et~al.(2022)Alayrac, Donahue, Luc, Miech, Barr, Hasson, Lenc, Mensch, Millican, Reynolds, et~al.]{alayrac2022flamingo}
Jean-Baptiste Alayrac, Jeff Donahue, Pauline Luc, Antoine Miech, Iain Barr, Yana Hasson, Karel Lenc, Arthur Mensch, Katherine Millican, Malcolm Reynolds, et~al.
\newblock Flamingo: a visual language model for few-shot learning.
\newblock In \emph{Advances in Neural Information Processing Systems (NeurIPS)}, 2022.

\bibitem[Li et~al.(2023{\natexlab{b}})Li, Li, Savarese, and Hoi]{li2023blip2}
Junnan Li, Dongxu Li, Silvio Savarese, and Steven Hoi.
\newblock {BLIP-2}: Bootstrapping language-image pre-training with frozen image encoders and large language models.
\newblock In \emph{International Conference on Machine Learning (ICML)}, 2023{\natexlab{b}}.

\bibitem[Liu et~al.(2023)Liu, Li, Wu, and Lee]{liu2023llava}
Haotian Liu, Chunyuan Li, Qingyang Wu, and Yong~Jae Lee.
\newblock Visual instruction tuning.
\newblock \emph{NeurIPS}, 2023.

\bibitem[Zhu et~al.(2023)Zhu, Chen, Shen, Li, and Elhoseiny]{zhu2023minigpt4}
Deyao Zhu, Jun Chen, Xiaoqian Shen, Xiang Li, and Mohamed Elhoseiny.
\newblock {MiniGPT-4}: Enhancing vision-language understanding with advanced large language models.
\newblock \emph{arXiv preprint arXiv:2304.10592}, 2023.

\bibitem[Dai et~al.(2023)Dai, Li, Li, Tiong, Zhao, Wang, Li, Fung, and Hoi]{dai2023instructblip}
Wenliang Dai, Junnan Li, Dongxu Li, Anthony Meng~Huat Tiong, Junqi Zhao, Weisheng Wang, Boyang Li, Pascale Fung, and Steven Hoi.
\newblock {InstructBLIP}: Towards general-purpose vision-language models with instruction tuning.
\newblock In \emph{Advances in Neural Information Processing Systems (NeurIPS)}, 2023.

\bibitem[Liu et~al.(2024{\natexlab{a}})Liu, Li, Li, and Lee]{liu2024llava15}
Haotian Liu, Chunyuan Li, Yuheng Li, and Yong~Jae Lee.
\newblock Improved baselines with visual instruction tuning.
\newblock In \emph{IEEE/CVF Conference on Computer Vision and Pattern Recognition (CVPR)}, 2024{\natexlab{a}}.

\bibitem[Bai et~al.(2023)Bai, Bai, Yang, Wang, Tan, Wang, Lin, Zhou, and Zhou]{bai2023qwenvl}
Jinze Bai, Shuai Bai, Shusheng Yang, Shijie Wang, Sinan Tan, Peng Wang, Junyang Lin, Chang Zhou, and Jingren Zhou.
\newblock {Qwen-VL}: A versatile vision-language model for understanding, localization, text reading, and beyond.
\newblock \emph{arXiv preprint arXiv:2308.12966}, 2023.

\bibitem[Chen et~al.(2024{\natexlab{b}})Chen, Wu, Wang, Su, Chen, Xing, Zhong, Zhang, Zhu, Lu, et~al.]{chen2024nnvl}
Zhe Chen, Jiannan Wu, Wenhai Wang, Weijie Su, Guo Chen, Sen Xing, Muyan Zhong, Qinglong Zhang, Xizhou Zhu, Lewei Lu, et~al.
\newblock {InternVL}: Scaling up vision foundation models and aligning for generic visual-linguistic tasks.
\newblock \emph{CVPR}, 2024{\natexlab{b}}.

\bibitem[Lu et~al.(2024{\natexlab{a}})Lu, Liu, Zhang, Wang, Dong, Liu, Sun, Ren, Li, Yang, et~al.]{lu2024deepseekvl}
Haoyu Lu, Wen Liu, Bo~Zhang, Bingxuan Wang, Kai Dong, Bo~Liu, Jingxiang Sun, Tongzheng Ren, Zhuoshu Li, Hao Yang, et~al.
\newblock {DeepSeek-VL}: Towards real-world vision-language understanding.
\newblock \emph{arXiv preprint arXiv:2403.05525}, 2024{\natexlab{a}}.

\bibitem[Agrawal et~al.(2024)Agrawal, Antoniak, Hanna, Bout, Chaplot, Chudnovsky, Costa, De~Monicault, Garg, Gervet, et~al.]{agrawal2024pixtral}
Pravesh Agrawal, Szymon Antoniak, Emma~Bou Hanna, Baptiste Bout, Devendra Chaplot, Jessica Chudnovsky, Diogo Costa, Baudouin De~Monicault, Saurabh Garg, Theophile Gervet, et~al.
\newblock Pixtral 12{B}.
\newblock \emph{arXiv preprint arXiv:2410.07073}, 2024.

\bibitem[Deitke et~al.(2024)Deitke, Clark, Lee, Tripathi, Yang, Park, Salehi, Muennighoff, Lo, Soldaini, et~al.]{deitke2024molmo}
Matt Deitke, Christopher Clark, Sangho Lee, Rohun Tripathi, Yue Yang, Jae~Sung Park, Mohammadreza Salehi, Niklas Muennighoff, Kyle Lo, Luca Soldaini, et~al.
\newblock Molmo and {PixMo}: Open weights and open data for state-of-the-art vision-language models.
\newblock \emph{arXiv preprint arXiv:2409.17146}, 2024.

\bibitem[Team et~al.(2026)Team, Bai, Bai, Bao, Cai, Cao, Charles, Che, Chen, Chen, et~al.]{team2026kimi}
Kimi Team, Tongtong Bai, Yifan Bai, Yiping Bao, SH~Cai, Yuan Cao, Y~Charles, HS~Che, Cheng Chen, Guanduo Chen, et~al.
\newblock Kimi k2. 5: Visual agentic intelligence.
\newblock \emph{arXiv preprint arXiv:2602.02276}, 2026.

\bibitem[Abdin et~al.(2024)Abdin, Jacobs, Awan, Aneja, Awadallah, Awadalla, Bach, Bahree, Bakhtiari, Bao, et~al.]{abdin2024phi3}
Marah Abdin, Sam~Ade Jacobs, Ammar~Ahmad Awan, Jyoti Aneja, Ahmed Awadallah, Hany Awadalla, Nguyen Bach, Amit Bahree, Arash Bakhtiari, Jianmin Bao, et~al.
\newblock {Phi-3} technical report: A highly capable language model locally on your phone.
\newblock \emph{arXiv preprint arXiv:2404.14219}, 2024.

\bibitem[Deng et~al.(2025)]{deng2025bagel}
Chaorui Deng et~al.
\newblock {BAGEL}: A unified multimodal foundation model for image understanding, generation, and editing.
\newblock \emph{arXiv preprint}, 2025.

\bibitem[Cui et~al.(2025)]{cui2025emu35nativemultimodalmodels}
Yuying Cui et~al.
\newblock {Emu3.5}: Native multimodal models.
\newblock \emph{arXiv preprint}, 2025.

\bibitem[Wang et~al.(2025{\natexlab{a}})Wang, Zhao, Zhang, Cao, Zhan, Duan, Lu, Fu, Chen, Zhao, et~al.]{wang2025ovis}
Guo-Hua Wang, Shanshan Zhao, Xinjie Zhang, Liangfu Cao, Pengxin Zhan, Lunhao Duan, Shiyin Lu, Minghao Fu, Xiaohao Chen, Jianshan Zhao, et~al.
\newblock Ovis-u1 technical report.
\newblock \emph{arXiv preprint arXiv:2506.23044}, 2025{\natexlab{a}}.

\bibitem[Zhang et~al.(2025{\natexlab{a}})Zhang, Guo, Zhao, Fu, Duan, Hu, Chng, Wang, Chen, Xu, Luo, and Zhang]{zhang2025unified}
Xinjie Zhang, Jintao Guo, Shanshan Zhao, Minghao Fu, Lunhao Duan, Jiakui Hu, Yong~Xien Chng, Guo-Hua Wang, Qing-Guo Chen, Zhao Xu, Weihua Luo, and Kaifu Zhang.
\newblock Unified multimodal understanding and generation models: Advances, challenges, and opportunities.
\newblock \emph{arXiv preprint arXiv:2505.02567}, 2025{\natexlab{a}}.

\bibitem[Li et~al.(2023{\natexlab{c}})Li, He, Wang, Li, Wang, Luo, Wang, Wang, and Qiao]{li2023videochat}
KunChang Li, Yinan He, Yi~Wang, Yizhuo Li, Wenhai Wang, Ping Luo, Yali Wang, Limin Wang, and Yu~Qiao.
\newblock {VideoChat}: Chat-centric video understanding.
\newblock \emph{arXiv preprint arXiv:2305.06355}, 2023{\natexlab{c}}.

\bibitem[Maaz et~al.(2023)Maaz, Rasheed, Khan, and Khan]{maaz2023videochatgpt}
Muhammad Maaz, Hanoona Rasheed, Salman Khan, and Fahad~Shahbaz Khan.
\newblock {Video-ChatGPT}: Towards detailed video understanding via large vision and language models.
\newblock \emph{arXiv preprint arXiv:2306.05424}, 2023.

\bibitem[Zhang et~al.(2023)Zhang, Li, and Bing]{zhang2023videollama}
Hang Zhang, Xin Li, and Lidong Bing.
\newblock {Video-LLaMA}: An instruction-tuned audio-visual language model for video understanding.
\newblock \emph{arXiv preprint arXiv:2306.02858}, 2023.

\bibitem[Lin et~al.(2023)Lin, Ye, Zhu, Cui, Ning, Jin, and Yuan]{lin2023videollava}
Bin Lin, Yang Ye, Bin Zhu, Jiaxi Cui, Munan Ning, Peng Jin, and Li~Yuan.
\newblock {Video-LLaVA}: Learning united visual representation by alignment before projection.
\newblock \emph{arXiv preprint arXiv:2311.10122}, 2023.

\bibitem[Li et~al.(2024{\natexlab{a}})Li, Zhang, Guo, Zhang, Li, Zhang, Zhang, Li, Liu, and Li]{li2024llavaonevision}
Bo~Li, Yuanhan Zhang, Dong Guo, Renrui Zhang, Feng Li, Hao Zhang, Kaichen Zhang, Yanwei Li, Ziwei Liu, and Chunyuan Li.
\newblock {LLaVA-OneVision}: Easy visual task transfer.
\newblock \emph{arXiv preprint arXiv:2408.03326}, 2024{\natexlab{a}}.

\bibitem[Zhang et~al.(2024{\natexlab{a}})Zhang, Zhang, Li, Zeng, Yang, Zhang, Wang, Tan, Li, and Liu]{zhang2024longva}
Peiyuan Zhang, Kaichen Zhang, Bo~Li, Guangtao Zeng, Jingkang Yang, Yuanhan Zhang, Ziyue Wang, Haoran Tan, Chunyuan Li, and Ziwei Liu.
\newblock Long context transfer from language to vision.
\newblock \emph{arXiv preprint arXiv:2406.16852}, 2024{\natexlab{a}}.

\bibitem[Chen et~al.(2024{\natexlab{c}})Chen, Lv, Wu, Lin, Song, Gao, Liu, Gao, Mao, and Shou]{chen2024videollmonline}
Joya Chen, Zhaoyang Lv, Shiwei Wu, Kevin~Qinghong Lin, Chenan Song, Difei Gao, Jia-Wei Liu, Ziteng Gao, Dongxing Mao, and Mike~Zheng Shou.
\newblock {VideoLLM-online}: Online video large language model for streaming video.
\newblock In \emph{IEEE/CVF Conference on Computer Vision and Pattern Recognition (CVPR)}, 2024{\natexlab{c}}.

\bibitem[Wang et~al.(2024{\natexlab{a}})Wang, Meng, Wang, Liang, Wei, Zhang, and Zhao]{wang2024mmduet}
Yueqian Wang, Xiaojun Meng, Yuxuan Wang, Jianxin Liang, Jiansheng Wei, Huishuai Zhang, and Dongyan Zhao.
\newblock {VideoLLM} knows when to speak: Enhancing time-sensitive video comprehension with video-text duet interaction format.
\newblock \emph{arXiv preprint arXiv:2411.17991}, 2024{\natexlab{a}}.

\bibitem[Qian et~al.(2025)Qian, Ding, Dong, Zhang, Zang, Cao, Lin, and Wang]{qian2025dispider}
Rui Qian, Shuangrui Ding, Xiaoyi Dong, Pan Zhang, Yuhang Zang, Yuhang Cao, Dahua Lin, and Jiaqi Wang.
\newblock {Dispider}: Enabling video llms with active real-time interaction via disentangled perception, decision, and reaction.
\newblock \emph{arXiv preprint arXiv:2501.03218}, 2025.

\bibitem[Ding et~al.(2025{\natexlab{a}})Ding, Wu, Yang, Jiang, Zhang, Bai, Chen, and Cao]{ding2025streammind}
Xin Ding, Hao Wu, Yifan Yang, Shiqi Jiang, Qianxi Zhang, Donglin Bai, Zhibo Chen, and Ting Cao.
\newblock Streammind: Unlocking full frame rate streaming video dialogue through event-gated cognition.
\newblock In \emph{Proceedings of the IEEE/CVF International Conference on Computer Vision}, pages 13448--13459, 2025{\natexlab{a}}.

\bibitem[{Video Understanding Team of JoyAI-VL @ Joy Future Academy, JD}(2026)]{joyai2026vlinteraction}
{Video Understanding Team of JoyAI-VL @ Joy Future Academy, JD}.
\newblock Joyai-vl-interaction: Real-time vision-language interaction intelligence.
\newblock Technical report, Joy Future Academy, JD, June 2026.

\bibitem[Zhang et~al.(2024{\natexlab{b}})Zhang, Wang, Tang, Liu, Feng, Dai, and Jin]{zhang2024flashvstream}
Haoji Zhang, Yiqin Wang, Yansong Tang, Yong Liu, Jiashi Feng, Jifeng Dai, and Xiaojie Jin.
\newblock {Flash-VStream}: Memory-based real-time understanding for long video streams.
\newblock \emph{arXiv preprint arXiv:2406.08085}, 2024{\natexlab{b}}.

\bibitem[Xu et~al.(2026)Xu, Xiao, Chen, He, Lu, and Han]{xu2026streamingvlmrealtimeunderstandinginfinite}
Ruyi Xu, Guangxuan Xiao, Yukang Chen, Liuning He, Yao Lu, and Song Han.
\newblock Streamingvlm: Real-time understanding for infinite video streams, 2026.
\newblock URL \url{https://arxiv.org/abs/2510.09608}.

\bibitem[Zhang et~al.(2024{\natexlab{c}})Zhang, Dong, Cao, Zang, Qian, Wei, Chen, Li, Niu, Ding, et~al.]{zhang2024ixc25omnilive}
Pan Zhang, Xiaoyi Dong, Yuhang Cao, Yuhang Zang, Rui Qian, Xilin Wei, Lin Chen, Yifei Li, Junbo Niu, Shuangrui Ding, et~al.
\newblock {InternLM-XComposer2.5-OmniLive}: A comprehensive multimodal system for long-term streaming video and audio interactions.
\newblock \emph{arXiv preprint arXiv:2412.09596}, 2024{\natexlab{c}}.

\bibitem[Chen et~al.(2025)Chen, Zeng, Lin, Li, Ma, and Shou]{chen2025livecclearningvideollm}
Joya Chen, Ziyun Zeng, Yiqi Lin, Wei Li, Zejun Ma, and Mike~Zheng Shou.
\newblock Livecc: Learning video llm with streaming speech transcription at scale, 2025.
\newblock URL \url{https://arxiv.org/abs/2504.16030}.

\bibitem[Rao et~al.(2024)Rao, Wu, Liu, Wang, and Xie]{rao2024matchtimeautomaticsoccergame}
Jiayuan Rao, Haoning Wu, Chang Liu, Yanfeng Wang, and Weidi Xie.
\newblock Matchtime: Towards automatic soccer game commentary generation, 2024.
\newblock URL \url{https://arxiv.org/abs/2406.18530}.

\bibitem[Lin et~al.(2024)Lin, Fang, Chen, Wan, Luo, Li, Liu, and Sun]{lin2024streamingbench}
Junming Lin, Zheng Fang, Chi Chen, Zihao Wan, Fuwen Luo, Peng Li, Yang Liu, and Maosong Sun.
\newblock {StreamingBench}: Assessing the gap for mllms to achieve streaming video understanding.
\newblock \emph{arXiv preprint arXiv:2411.03628}, 2024.

\bibitem[Li et~al.(2025{\natexlab{c}})Li, Niu, Miao, Ge, Zhou, He, Dong, Duan, Ding, Qian, Zhang, Zang, Cao, He, and Wang]{li2025ovobench}
Yifei Li, Junbo Niu, Ziyang Miao, Chunjiang Ge, Yuanhang Zhou, Qihao He, Xiaoyi Dong, Haodong Duan, Shuangrui Ding, Rui Qian, Pan Zhang, Yuhang Zang, Yuhang Cao, Conghui He, and Jiaqi Wang.
\newblock {OVO-Bench}: How far is your video-llms from real-world online video understanding?
\newblock In \emph{IEEE/CVF Conference on Computer Vision and Pattern Recognition (CVPR)}, 2025{\natexlab{c}}.

\bibitem[Sullivan et~al.(2012)Sullivan, Ohm, Han, and Wiegand]{sullivan2012hevc}
Gary~J. Sullivan, Jens-Rainer Ohm, Woo-Jin Han, and Thomas Wiegand.
\newblock Overview of the high efficiency video coding {(HEVC)} standard.
\newblock In \emph{IEEE Transactions on Circuits and Systems for Video Technology}, 2012.

\bibitem[Jia et~al.(2025)Jia, Li, Li, Xie, Qi, Li, and Lu]{jia2025towards}
Zhaoyang Jia, Bin Li, Jiahao Li, Wenxuan Xie, Linfeng Qi, Houqiang Li, and Yan Lu.
\newblock Towards practical real-time neural video compression.
\newblock In \emph{Proceedings of the Computer Vision and Pattern Recognition Conference}, pages 12543--12552, 2025.

\bibitem[Dao(2024)]{dao2023flashattention2}
Tri Dao.
\newblock Flash{A}ttention-2: Faster attention with better parallelism and work partitioning.
\newblock In \emph{International Conference on Learning Representations (ICLR)}, 2024.

\bibitem[Su et~al.(2024)Su, Lu, Pan, Murtadha, Wen, and Liu]{su2024rope}
Jianlin Su, Yu~Lu, Shengfeng Pan, Ahmed Murtadha, Bo~Wen, and Yunfeng Liu.
\newblock {RoFormer}: Enhanced transformer with rotary position embedding.
\newblock \emph{Neurocomputing}, 2024.

\bibitem[Schuhmann et~al.(2021)Schuhmann, Vencu, Beaumont, Kaczmarczyk, Mullis, Katta, Coombes, Jitsev, and Komatsuzaki]{schuhmann2021laion}
Christoph Schuhmann, Richard Vencu, Romain Beaumont, Robert Kaczmarczyk, Clayton Mullis, Aarush Katta, Theo Coombes, Jenia Jitsev, and Aran Komatsuzaki.
\newblock {LAION}-400m: Open dataset of {CLIP}-filtered 400 million image-text pairs.
\newblock \emph{arXiv preprint arXiv:2111.02114}, 2021.

\bibitem[Byeon et~al.(2022)Byeon, Park, Kim, Lee, Baek, and Kim]{byeon2022coyo}
Minwoo Byeon, Beomhee Park, Haecheon Kim, Sungjun Lee, Woonhyuk Baek, and Saehoon Kim.
\newblock {COYO}-700m: Image-text pair dataset.
\newblock \url{https://github.com/kakaobrain/coyo-dataset}, 2022.

\bibitem[Lauren\c{c}on et~al.(2023)Lauren\c{c}on, Saulnier, Tronchon, Bekman, Singh, Lozhkov, Wang, Karamcheti, Rush, Kiela, Cord, and Sanh]{laurenccon2023obelics}
Hugo Lauren\c{c}on, Lucile Saulnier, L\'{e}o Tronchon, Stas Bekman, Amanpreet Singh, Anton Lozhkov, Thomas Wang, Siddharth Karamcheti, Alexander~M. Rush, Douwe Kiela, Matthieu Cord, and Victor Sanh.
\newblock {OBELICS}: An open web-scale filtered dataset of interleaved image-text documents.
\newblock In \emph{Advances in Neural Information Processing Systems (NeurIPS)}, 2023.

\bibitem[Xie et~al.(2023)Xie, Cai, Li, Kong, Wu, Song, Morimitsu, Yao, Wang, Zhang, Leng, Zhang, Ji, and Deng]{xie2023ccmb}
Chunyu Xie, Heng Cai, Jincheng Li, Fanjing Kong, Xiaoyu Wu, Jianfei Song, Henrique Morimitsu, Lin Yao, Dexin Wang, Xiangzheng Zhang, Dawei Leng, Baochang Zhang, Xiangyang Ji, and Yafeng Deng.
\newblock {CCMB}: A large-scale chinese cross-modal benchmark.
\newblock \emph{arXiv preprint arXiv:2205.03860}, 2023.

\bibitem[Deng et~al.(2009)Deng, Dong, Socher, Li, Li, and Fei-Fei]{deng2009imagenet}
Jia Deng, Wei Dong, Richard Socher, Li-Jia Li, Kai Li, and Li~Fei-Fei.
\newblock {ImageNet}: A large-scale hierarchical image database.
\newblock In \emph{IEEE Conference on Computer Vision and Pattern Recognition (CVPR)}, pages 248--255, 2009.

\bibitem[Miech et~al.(2019)Miech, Zhukov, Alayrac, Tapaswi, Laptev, and Sivic]{miech2019howto100m}
Antoine Miech, Dimitri Zhukov, Jean-Baptiste Alayrac, Makarand Tapaswi, Ivan Laptev, and Josef Sivic.
\newblock {HowTo100M}: Learning a text-video embedding by watching hundred million narrated video clips.
\newblock In \emph{IEEE/CVF International Conference on Computer Vision (ICCV)}, 2019.

\bibitem[Chen et~al.(2024{\natexlab{d}})Chen, Siarohin, Menapace, Deyneka, Chao, Jeon, Fang, Lee, Ren, Yang, and Tulyakov]{chen2024panda70m}
Tsai-Shien Chen, Aliaksandr Siarohin, Willi Menapace, Ekaterina Deyneka, Hsiang-wei Chao, Byung~Eun Jeon, Yuwei Fang, Hsin-Ying Lee, Jian Ren, Ming-Hsuan Yang, and Sergey Tulyakov.
\newblock {Panda}-70m: Captioning 70m videos with multiple cross-modality teachers.
\newblock In \emph{IEEE/CVF Conference on Computer Vision and Pattern Recognition (CVPR)}, 2024{\natexlab{d}}.

\bibitem[{Qwen Team}(2025)]{qwen3}
{Qwen Team}.
\newblock {Qwen3} technical report.
\newblock \emph{arXiv preprint}, 2025.

\bibitem[Zhang et~al.(2026{\natexlab{a}})Zhang, Zhang, Zheng, Guo, Jia, Shen, Guo, Luo, Li, Xie, Pu, Zhang, Zhang, Guo, Bi, Gui, Liu, Wen, Zheng, Yang, Li, Wang, Li, and Lu]{zhang2026mageflow}
Xinjie Zhang, Peng Zhang, Shicheng Zheng, Jinghao Guo, Zhaoyang Jia, Yifei Shen, Xun Guo, Yuxuan Luo, Jiahao Li, Wenxuan Xie, Fanyi Pu, Xiaoyi Zhang, Kaichen Zhang, Zongyu Guo, Tianci Bi, Dongnan Gui, Zhening Liu, Zimo Wen, Zihan Zheng, Senqiao Yang, Xiao Li, Jinglu Wang, Bin Li, and Yan Lu.
\newblock Mage-flow: An efficient native-resolution foundation model for image generation and editing.
\newblock \emph{arXiv preprint arXiv:2607.19064}, 2026{\natexlab{a}}.

\bibitem[Team(2026)]{ai4ai_scale2026}
Microsoft~Mage Team.
\newblock Ai4ai at scale: A full-pipeline system for enhancing llm agentic capabilities.
\newblock Technical report, Microsoft, 2026.

\bibitem[Wiedmann et~al.(2025)Wiedmann, Zohar, Mahla, Wang, Li, Frere, von Werra, Roy~Gosthipaty, and Marafioti]{finevision25m}
Luis Wiedmann, Orr Zohar, Amir Mahla, Xiaohan Wang, Rui Li, Thibaud Frere, Leandro von Werra, Aritra Roy~Gosthipaty, and Andr\'es Marafioti.
\newblock {FineVision}: Open data is all you need.
\newblock \emph{arXiv preprint arXiv:2510.17269}, 2025.

\bibitem[Zhang et~al.(2026{\natexlab{b}})Zhang, Ni, Chen, Zhang, Rao, Peng, Lu, Hu, Guo, and Hu]{zhang2026beehighqualitycorpusfullstack}
Yi~Zhang, Bolin Ni, Xin-Sheng Chen, Heng-Rui Zhang, Yongming Rao, Houwen Peng, Qinglin Lu, Han Hu, Meng-Hao Guo, and Shi-Min Hu.
\newblock {Bee}: A high-quality corpus and full-stack suite to unlock advanced fully open {MLLMs}, 2026{\natexlab{b}}.
\newblock URL \url{https://arxiv.org/abs/2510.13795}.

\bibitem[Zhang et~al.(2025{\natexlab{b}})Zhang, Wu, Li, Li, Ma, Liu, and Li]{zhang2025llavavideovideoinstructiontuning}
Yuanhan Zhang, Jinming Wu, Wei Li, Bo~Li, Zejun Ma, Ziwei Liu, and Chunyuan Li.
\newblock {LLaVA-Video}: Video instruction tuning with synthetic data, 2025{\natexlab{b}}.
\newblock URL \url{https://arxiv.org/abs/2410.02713}.

\bibitem[Zhang et~al.(2026{\natexlab{c}})Zhang, Wang, Ge, Ge, Li, Shan, and Wang]{zhang2026timelensrethinkingvideotemporal}
Jun Zhang, Teng Wang, Yuying Ge, Yixiao Ge, Xinhao Li, Ying Shan, and Limin Wang.
\newblock {TimeLens}: Rethinking video temporal grounding with multimodal {LLMs}, 2026{\natexlab{c}}.
\newblock URL \url{https://arxiv.org/abs/2512.14698}.

\bibitem[Clark et~al.(2026)Clark, Zhang, Ma, Park, Tripathi, Lee, Salehi, Ren, Kim, Yang, et~al.]{clark2026molmo2}
Christopher Clark, Jieyu Zhang, Zixian Ma, Jae~Sung Park, Rohun Tripathi, Sangho Lee, Mohammadreza Salehi, Jason Ren, Chris~Dongjoo Kim, Yinuo Yang, et~al.
\newblock Molmo2: Open weights and data for vision-language models with video understanding and grounding.
\newblock In \emph{Proceedings of the IEEE/CVF Conference on Computer Vision and Pattern Recognition}, pages 28652--28668, 2026.

\bibitem[Ding et~al.(2025{\natexlab{b}})Ding, Wu, Yang, Jiang, Bai, Chen, and Cao]{ding2025streammindunlockingframerate}
Xin Ding, Hao Wu, Yifan Yang, Shiqi Jiang, Donglin Bai, Zhibo Chen, and Ting Cao.
\newblock Streammind: Unlocking full frame rate streaming video dialogue through event-gated cognition, 2025{\natexlab{b}}.
\newblock URL \url{https://arxiv.org/abs/2503.06220}.

\bibitem[Cimpoi et~al.(2013)Cimpoi, Maji, Kokkinos, Mohamed, and Vedaldi]{bench_dtd}
Mircea Cimpoi, Subhransu Maji, Iasonas Kokkinos, Sammy Mohamed, and Andrea Vedaldi.
\newblock Describing textures in the wild.
\newblock \emph{arXiv preprint arXiv:1311.3618}, 2013.

\bibitem[Krizhevsky(2009)]{bench_cifar10}
Alex Krizhevsky.
\newblock Learning multiple layers of features from tiny images.
\newblock Technical report, University of Toronto, 2009.

\bibitem[Xiao et~al.(2010)Xiao, Hays, Ehinger, Oliva, and Torralba]{bench_sun397}
Jianxiong Xiao, James Hays, Krista~A. Ehinger, Aude Oliva, and Antonio Torralba.
\newblock {SUN} database: Large-scale scene recognition from abbey to zoo.
\newblock In \emph{CVPR}, 2010.

\bibitem[Bossard et~al.(2014)Bossard, Guillaumin, and Van~Gool]{bench_food101}
Lukas Bossard, Matthieu Guillaumin, and Luc Van~Gool.
\newblock Food-101 -- mining discriminative components with random forests.
\newblock In \emph{ECCV}, 2014.

\bibitem[Russakovsky et~al.(2015)Russakovsky, Deng, Su, Krause, Satheesh, Ma, Huang, Karpathy, Khosla, Bernstein, Berg, and Fei-Fei]{bench_imagenet}
Olga Russakovsky, Jia Deng, Hao Su, Jonathan Krause, Sanjeev Satheesh, Sean Ma, Zhiheng Huang, Andrej Karpathy, Aditya Khosla, Michael Bernstein, Alexander~C. Berg, and Li~Fei-Fei.
\newblock Imagenet large scale visual recognition challenge.
\newblock \emph{International Journal of Computer Vision}, 2015.

\bibitem[Chuang et~al.(2025)Chuang, Li, Wang, Yeh, Lyu, Raghavendra, Glass, Huang, Weston, Zettlemoyer, Chen, Liu, Xie, Yih, Li, and Xu]{model_metaclip2}
Yung-Sung Chuang, Yang Li, Dong Wang, Ching-Feng Yeh, Kehan Lyu, Ramya Raghavendra, James Glass, Lifei Huang, Jason Weston, Luke Zettlemoyer, Xinlei Chen, Zhuang Liu, Saining Xie, Wen-tau Yih, Shang-Wen Li, and Hu~Xu.
\newblock Meta clip 2: A worldwide scaling recipe.
\newblock \emph{arXiv preprint arXiv:2507.22062}, 2025.

\bibitem[Fini et~al.(2024)Fini, Shukor, Li, Dufter, Klein, Haldimann, et~al.]{model_aimv2}
Enrico Fini, Mustafa Shukor, Xiujun Li, Philipp Dufter, Michal Klein, David Haldimann, et~al.
\newblock Multimodal autoregressive pre-training of large vision encoders.
\newblock \emph{arXiv preprint arXiv:2411.14402}, 2024.

\bibitem[Sim{\'e}oni et~al.(2025)Sim{\'e}oni, Vo, Seitzer, Baldassarre, Oquab, Jose, et~al.]{model_dinov3}
Oriane Sim{\'e}oni, Huy~V. Vo, Maximilian Seitzer, Federico Baldassarre, Maxime Oquab, Cijo Jose, et~al.
\newblock {DINOv3}.
\newblock \emph{arXiv preprint arXiv:2508.10104}, 2025.

\bibitem[{Kimi Team}(2025{\natexlab{b}})]{model_moonvit}
{Kimi Team}.
\newblock Kimi-vl technical report.
\newblock \emph{arXiv preprint arXiv:2504.07491}, 2025{\natexlab{b}}.

\bibitem[Li et~al.(2018)Li, Li, and Vasconcelos]{bench_diving48}
Yingwei Li, Yi~Li, and Nuno Vasconcelos.
\newblock {RESOUND}: Towards action recognition without representation bias.
\newblock In \emph{ECCV}, 2018.

\bibitem[Kuehne et~al.(2011)Kuehne, Jhuang, Garrote, Poggio, and Serre]{bench_hmdb51}
Hildegard Kuehne, Hueihan Jhuang, Est{\'i}baliz Garrote, Tomaso Poggio, and Thomas Serre.
\newblock {HMDB}: A large video database for human motion recognition.
\newblock In \emph{ICCV}, 2011.

\bibitem[P{\u{a}}tr{\u{a}}ucean et~al.(2023)P{\u{a}}tr{\u{a}}ucean, Smaira, Gupta, Recasens, et~al.]{bench_perceptiontest}
Viorica P{\u{a}}tr{\u{a}}ucean, Lucas Smaira, Ankush Gupta, Adri{\`a} Recasens, et~al.
\newblock Perception test: A diagnostic benchmark for multimodal video models.
\newblock In \emph{NeurIPS Datasets and Benchmarks Track}, 2023.

\bibitem[Sigurdsson et~al.(2018)Sigurdsson, Gupta, Schmid, Farhadi, and Alahari]{bench_charadesego}
Gunnar~A. Sigurdsson, Abhinav Gupta, Cordelia Schmid, Ali Farhadi, and Karteek Alahari.
\newblock Actor and observer: Joint modeling of first and third-person videos.
\newblock In \emph{CVPR}, 2018.

\bibitem[Kay et~al.(2017)Kay, Carreira, Simonyan, Zhang, Hillier, Vijayanarasimhan, et~al.]{bench_kinetics400}
Will Kay, Joao Carreira, Karen Simonyan, Brian Zhang, Chloe Hillier, Sudheendra Vijayanarasimhan, et~al.
\newblock The kinetics human action video dataset.
\newblock \emph{arXiv preprint arXiv:1705.06950}, 2017.

\bibitem[Zhang et~al.(2025{\natexlab{c}})Zhang, Li, Zhang, Pu, Cahyono, Hu, Liu, Zhang, Yang, Li, et~al.]{zhang2025lmms}
Kaichen Zhang, Bo~Li, Peiyuan Zhang, Fanyi Pu, Joshua~Adrian Cahyono, Kairui Hu, Shuai Liu, Yuanhan Zhang, Jingkang Yang, Chunyuan Li, et~al.
\newblock Lmms-eval: Reality check on the evaluation of large multimodal models.
\newblock In \emph{Findings of the Association for Computational Linguistics: NAACL 2025}, pages 881--916, 2025{\natexlab{c}}.

\bibitem[Li et~al.(2024{\natexlab{b}})Li, Wang, He, Li, Wang, et~al.]{bench_mvbench}
Kunchang Li, Yali Wang, Yinan He, Yizhuo Li, Yi~Wang, et~al.
\newblock {MVBench}: A comprehensive multi-modal video understanding benchmark.
\newblock In \emph{CVPR}, 2024{\natexlab{b}}.

\bibitem[Xiao et~al.(2021{\natexlab{a}})Xiao, Shang, Yao, and Chua]{bench_nextqa}
Junbin Xiao, Xindi Shang, Angela Yao, and Tat-Seng Chua.
\newblock {NExT-QA}: Next phase of question-answering to explaining temporal actions.
\newblock In \emph{CVPR}, 2021{\natexlab{a}}.

\bibitem[Liu et~al.(2024{\natexlab{b}})Liu, Li, Liu, Wang, Ren, et~al.]{bench_tempcompass}
Yuanxin Liu, Shicheng Li, Yi~Liu, Yuxiang Wang, Shuhuai Ren, et~al.
\newblock {TempCompass}: Do video {LLMs} really understand videos?
\newblock \emph{arXiv preprint arXiv:2403.00476}, 2024{\natexlab{b}}.

\bibitem[Fu et~al.(2024{\natexlab{a}})Fu, Dai, Luo, Li, Ren, et~al.]{bench_videomme}
Chaoyou Fu, Yuhan Dai, Yongdong Luo, Lei Li, Shuhuai Ren, et~al.
\newblock {Video-MME}: The first-ever comprehensive evaluation benchmark of multi-modal {LLMs} in video analysis.
\newblock \emph{arXiv preprint arXiv:2405.21075}, 2024{\natexlab{a}}.

\bibitem[Wu et~al.(2024)Wu, Li, Chen, and Li]{bench_longvideobench}
Haoning Wu, Dongxu Li, Bei Chen, and Junnan Li.
\newblock {LongVideoBench}: A benchmark for long-context interleaved video-language understanding.
\newblock \emph{arXiv preprint arXiv:2407.15754}, 2024.

\bibitem[Wang et~al.(2024{\natexlab{b}})Wang, He, Hong, Cheng, Zhang, et~al.]{bench_lvbench}
Weihan Wang, Zehai He, Wenyi Hong, Yean Cheng, Xiaohan Zhang, et~al.
\newblock {LVBench}: An extreme long video understanding benchmark.
\newblock \emph{arXiv preprint arXiv:2406.08035}, 2024{\natexlab{b}}.

\bibitem[Zhou et~al.(2024)Zhou, Shu, Zhao, Wu, Liang, et~al.]{bench_mlvu}
Junjie Zhou, Yan Shu, Bo~Zhao, Boya Wu, Zhengyang Liang, et~al.
\newblock {MLVU}: Benchmarking multi-task long video understanding.
\newblock \emph{arXiv preprint arXiv:2406.04264}, 2024.

\bibitem[Gao et~al.(2017)Gao, Sun, Yang, and Nevatia]{bench_charades_sta}
Jiyang Gao, Chen Sun, Zhenheng Yang, and Ram Nevatia.
\newblock {TALL}: Temporal activity localization via language query.
\newblock In \emph{ICCV}, 2017.

\bibitem[Zhang et~al.(2025{\natexlab{d}})Zhang, Wang, Ge, Ge, Li, Shan, and Wang]{zhang2025timelens}
Jun Zhang, Teng Wang, Yuying Ge, Yixiao Ge, Xinhao Li, Ying Shan, and Limin Wang.
\newblock Timelens: Rethinking video temporal grounding with multimodal llms.
\newblock \emph{arXiv preprint arXiv:2512.14698}, 2025{\natexlab{d}}.

\bibitem[Yang et~al.(2025{\natexlab{b}})Yang, Yang, Gupta, Han, Fei-Fei, and Xie]{yang2024vsi}
Jihan Yang, Shusheng Yang, Anjali Gupta, Rilyn Han, Li~Fei-Fei, and Saining Xie.
\newblock Thinking in space: How multimodal large language models see, remember, and recall spaces.
\newblock In \emph{CVPR}, 2025{\natexlab{b}}.

\bibitem[Mathew et~al.(2021{\natexlab{a}})Mathew, Karatzas, and Jawahar]{bench_docvqa}
Minesh Mathew, Dimosthenis Karatzas, and C.V. Jawahar.
\newblock {DocVQA}: A dataset for {VQA} on document images.
\newblock In \emph{WACV}, 2021{\natexlab{a}}.

\bibitem[Mathew et~al.(2021{\natexlab{b}})Mathew, Bagal, P{\'e}rez-Tito, Karatzas, Valveny, and Jawahar]{bench_infographicvqa}
Minesh Mathew, Viraj Bagal, Rub{\`e}n P{\'e}rez-Tito, Dimosthenis Karatzas, Ernest Valveny, and C.V. Jawahar.
\newblock {InfographicVQA}.
\newblock \emph{arXiv preprint arXiv:2104.12756}, 2021{\natexlab{b}}.

\bibitem[Kembhavi et~al.(2016)Kembhavi, Salvato, Kolve, Seo, Hajishirzi, and Farhadi]{bench_ai2d}
Aniruddha Kembhavi, Mike Salvato, Eric Kolve, Minjoon Seo, Hannaneh Hajishirzi, and Ali Farhadi.
\newblock A diagram is worth a dozen images.
\newblock \emph{arXiv preprint arXiv:1603.07396}, 2016.

\bibitem[Masry et~al.(2022)Masry, Long, Tan, Joty, and Hoque]{bench_chartqa}
Ahmed Masry, Do~Xuan Long, Jia~Qing Tan, Shafiq Joty, and Enamul Hoque.
\newblock {ChartQA}: A benchmark for question answering about charts with visual and logical reasoning.
\newblock In \emph{Findings of ACL}, 2022.

\bibitem[Liu et~al.(2024{\natexlab{c}})Liu, Li, Huang, Yang, Yu, Li, et~al.]{bench_ocrbench}
Yuliang Liu, Zhang Li, Mingxin Huang, Biao Yang, Wenwen Yu, Chunyuan Li, et~al.
\newblock {OCRBench}: On the hidden mystery of {OCR} in large multimodal models.
\newblock \emph{Science China Information Sciences}, 2024{\natexlab{c}}.

\bibitem[Yang et~al.(2024)Yang, Tang, Li, Wang, Wan, et~al.]{bench_cc_ocr}
Zhibo Yang, Jun Tang, Zhaohai Li, Pengfei Wang, Jianqiang Wan, et~al.
\newblock {CC-OCR}: A comprehensive and challenging {OCR} benchmark for evaluating large multimodal models in literacy.
\newblock \emph{arXiv preprint arXiv:2412.02210}, 2024.

\bibitem[Tito et~al.(2022)Tito, Karatzas, and Valveny]{bench_mp_docvqa}
Rub{\`e}n Tito, Dimosthenis Karatzas, and Ernest Valveny.
\newblock Hierarchical multimodal transformers for multi-page {DocVQA}.
\newblock \emph{arXiv preprint arXiv:2212.05935}, 2022.

\bibitem[Van~Landeghem et~al.(2023)Van~Landeghem, Tito, Borchmann, Pietruszka, et~al.]{bench_dude}
Jordy Van~Landeghem, Rub{\'e}n Tito, {\L}ukasz Borchmann, Micha{\l} Pietruszka, et~al.
\newblock Document understanding dataset and evaluation ({DUDE}).
\newblock In \emph{ICCV}, 2023.

\bibitem[Chen et~al.(2021)Chen, Zhao, Chen, Zhang, Ji, Luo, Xiong, and Yu]{bench_websrc}
Xingyu Chen, Zihan Zhao, Lu~Chen, Danyang Zhang, Jiabao Ji, Ao~Luo, Yuxuan Xiong, and Kai Yu.
\newblock {WebSRC}: A dataset for web-based structural reading comprehension.
\newblock In \emph{EMNLP}, 2021.

\bibitem[Masry et~al.(2025)Masry, Islam, Ahmed, Bajaj, et~al.]{bench_chartqapro}
Ahmed Masry, Mohammed~Saidul Islam, Mahir Ahmed, Aayush Bajaj, et~al.
\newblock {ChartQAPro}: A more diverse and challenging benchmark for chart question answering.
\newblock \emph{arXiv preprint arXiv:2504.05506}, 2025.

\bibitem[Singh et~al.(2019)Singh, Natarajan, Shah, Jiang, Chen, Batra, Parikh, and Rohrbach]{bench_textvqa}
Amanpreet Singh, Vivek Natarajan, Meet Shah, Yu~Jiang, Xinlei Chen, Dhruv Batra, Devi Parikh, and Marcus Rohrbach.
\newblock Towards {VQA} models that can read.
\newblock In \emph{CVPR}, 2019.

\bibitem[Wang et~al.(2024{\natexlab{c}})Wang, Xia, He, Chen, Liu, et~al.]{bench_charxiv}
Zirui Wang, Mengzhou Xia, Luxi He, Howard Chen, Yitao Liu, et~al.
\newblock {CharXiv}: Charting gaps in realistic chart understanding in multimodal {LLMs}.
\newblock In \emph{NeurIPS Datasets and Benchmarks Track}, 2024{\natexlab{c}}.

\bibitem[Liu et~al.(2024{\natexlab{d}})Liu, Duan, Zhang, Li, Zhang, et~al.]{bench_mmbench}
Yuan Liu, Haodong Duan, Yuanhan Zhang, Bo~Li, Songyang Zhang, et~al.
\newblock {MMBench}: Is your multi-modal model an all-around player?
\newblock In \emph{ECCV}, 2024{\natexlab{d}}.

\bibitem[{xAI}(2024)]{bench_realworldqa}
{xAI}.
\newblock {RealWorldQA}: A benchmark for real-world spatial understanding.
\newblock \url{https://x.ai/news/grok-1.5v}, 2024.

\bibitem[Chen et~al.(2024{\natexlab{e}})Chen, Li, Dong, Zhang, Zang, et~al.]{bench_mmstar}
Lin Chen, Jinsong Li, Xiaoyi Dong, Pan Zhang, Yuhang Zang, et~al.
\newblock Are we on the right way for evaluating large vision-language models?
\newblock In \emph{NeurIPS}, 2024{\natexlab{e}}.

\bibitem[Fu et~al.(2023)Fu, Chen, Shen, Qin, Zhang, Lin, et~al.]{bench_mme}
Chaoyou Fu, Peixian Chen, Yunhang Shen, Yulei Qin, Mengdan Zhang, Xu~Lin, et~al.
\newblock {MME}: A comprehensive evaluation benchmark for multimodal large language models.
\newblock \emph{arXiv preprint arXiv:2306.13394}, 2023.

\bibitem[Li et~al.(2024{\natexlab{c}})Li, Wang, Wang, Ge, Ge, and Shan]{bench_seedbench}
Bohao Li, Rui Wang, Guangzhi Wang, Yuying Ge, Yixiao Ge, and Ying Shan.
\newblock {SEED-Bench}: Benchmarking multimodal {LLMs} with generative comprehension.
\newblock In \emph{CVPR}, 2024{\natexlab{c}}.

\bibitem[Li et~al.(2024{\natexlab{d}})Li, Ge, Chen, Ge, Zhang, and Shan]{bench_seedbench2plus}
Bohao Li, Yuying Ge, Yi~Chen, Yixiao Ge, Ruimao Zhang, and Ying Shan.
\newblock {SEED-Bench-2-Plus}: Benchmarking multimodal large language models with text-rich visual comprehension.
\newblock \emph{arXiv preprint arXiv:2404.16790}, 2024{\natexlab{d}}.

\bibitem[Ying et~al.(2024)Ying, Meng, Wang, Li, Lin, et~al.]{bench_mmtbench}
Kaining Ying, Fanqing Meng, Jin Wang, Zhiqian Li, Han Lin, et~al.
\newblock {MMT-Bench}: A comprehensive multimodal benchmark for evaluating large vision-language models towards multitask {AGI}.
\newblock In \emph{ICML}, 2024.

\bibitem[Zhang et~al.(2025{\natexlab{e}})Zhang, Zhang, Tian, Fu, Zhang, et~al.]{bench_mmerealworld}
Yi-Fan Zhang, Huanyu Zhang, Haochen Tian, Chaoyou Fu, Shuangqing Zhang, et~al.
\newblock {MME-RealWorld}: Could your multimodal {LLM} challenge high-resolution real-world scenarios that are difficult for humans?
\newblock In \emph{ICLR}, 2025{\natexlab{e}}.

\bibitem[Tong et~al.(2024{\natexlab{b}})Tong, Brown, Wu, Woo, Middepogu, Akula, Yang, Yang, Iyer, Pan, et~al.]{tong2024cambrian}
Shengbang Tong, Ellis Brown, Penghao Wu, Sanghyun Woo, Manoj Middepogu, Sai~C Akula, Jihan Yang, Shusheng Yang, Adithya Iyer, Xichen Pan, et~al.
\newblock Cambrian-1: A fully open, vision-centric exploration of multimodal llms.
\newblock \emph{Advances in Neural Information Processing Systems}, 37:\penalty0 87310--87356, 2024{\natexlab{b}}.

\bibitem[Fu et~al.(2024{\natexlab{b}})Fu, Hu, Li, Feng, Wang, et~al.]{bench_blink}
Xingyu Fu, Yushi Hu, Bangzheng Li, Yu~Feng, Haoyu Wang, et~al.
\newblock {BLINK}: Multimodal large language models can see but not perceive.
\newblock In \emph{ECCV}, 2024{\natexlab{b}}.

\bibitem[Du et~al.(2024)Du, Wu, Li, Huang, and Wei]{bench_embspatial}
Mengfei Du, Binhao Wu, Zejun Li, Xuanjing Huang, and Zhongyu Wei.
\newblock {EmbSpatial-Bench}: Benchmarking spatial understanding for embodied tasks with large vision-language models.
\newblock In \emph{ACL}, 2024.

\bibitem[Wang et~al.(2024{\natexlab{d}})Wang, Ren, Luo, Li, Yan, et~al.]{bench_crpe}
Weiyun Wang, Yiming Ren, Haowen Luo, Tiantong Li, Chenxiang Yan, et~al.
\newblock The all-seeing project {V2}: Towards general relation comprehension of the open world.
\newblock In \emph{ECCV}, 2024{\natexlab{d}}.

\bibitem[Wang et~al.(2025{\natexlab{b}})Wang, Ji, Liu, Zhou, Yang, Tian, Qin, Liu, Tan, Chi, Ma, Zeng, and Zheng]{wang2025crosspoint}
Yipu Wang, Yuheng Ji, Yuyang Liu, Enshen Zhou, Ziqiang Yang, Yuxuan Tian, Ziheng Qin, Yue Liu, Huajie Tan, Cheng Chi, Zhiyuan Ma, Daniel~Dajun Zeng, and Xiaolong Zheng.
\newblock Towards cross-view point correspondence in vision-language models.
\newblock \emph{arXiv preprint arXiv:2512.04686}, 2025{\natexlab{b}}.

\bibitem[{Gemini Robotics Team}(2025)]{bench_erqa}
{Gemini Robotics Team}.
\newblock Gemini robotics: Bringing {AI} into the physical world.
\newblock Technical report, Google DeepMind, 2025.
\newblock arXiv:2503.20020.

\bibitem[Yang et~al.(2025{\natexlab{c}})Yang, Xu, Xie, Yang, et~al.]{bench_mmsibench}
Sihan Yang, Runsen Xu, Yiman Xie, Sizhe Yang, et~al.
\newblock {MMSI-Bench}: A benchmark for multi-image spatial intelligence.
\newblock \emph{arXiv preprint arXiv:2505.23764}, 2025{\natexlab{c}}.

\bibitem[Ray et~al.(2025)Ray, Duan, Tan, Bashkirova, et~al.]{bench_sat}
Arijit Ray, Jiafei Duan, Reuben Tan, Dina Bashkirova, et~al.
\newblock {SAT}: Dynamic spatial aptitude training for multimodal language models.
\newblock In \emph{COLM}, 2025.

\bibitem[Fu et~al.(2026)Fu, Yuan, Dong, Zhang, Shen, Hu, Li, Su, Long, Xie, et~al.]{fu2026video}
Chaoyou Fu, Haozhi Yuan, Yuhao Dong, Yi-Fan Zhang, Yunhang Shen, Xiaoxing Hu, Xueying Li, Jinsen Su, Chengwu Long, Xiaoyao Xie, et~al.
\newblock Video-mme-v2: Towards the next stage in benchmarks for comprehensive video understanding.
\newblock \emph{arXiv preprint arXiv:2604.05015}, 2026.

\bibitem[Ma et~al.(2025)Ma, Ren, Jia, Li, Nie, Zhang, and Chen]{bench_videoevalpro}
Wentao Ma, Weiming Ren, Yiming Jia, Zhuofeng Li, Ping Nie, Ge~Zhang, and Wenhu Chen.
\newblock {VideoEval-Pro}: Robust and realistic long video understanding evaluation.
\newblock \emph{arXiv preprint arXiv:2505.14640}, 2025.

\bibitem[Goel et~al.(2026)Goel, Ghosh, Agarwal, Anand, Jayakumar, Koroshinadze, Xu, Lyons, Case, Sapra, et~al.]{goel2026mmou}
Arushi Goel, Sreyan Ghosh, Vatsal Agarwal, Nishit Anand, Kaousheik Jayakumar, Lasha Koroshinadze, Yao Xu, Katie Lyons, James Case, Karan Sapra, et~al.
\newblock Mmou: A massive multi-task omni understanding and reasoning benchmark for long and complex real-world videos.
\newblock \emph{arXiv preprint arXiv:2603.14145}, 2026.

\bibitem[Khoreva et~al.(2018)Khoreva, Rohrbach, and Schiele]{bench_refdavis17}
Anna Khoreva, Anna Rohrbach, and Bernt Schiele.
\newblock Video object segmentation with language referring expressions.
\newblock In \emph{ACCV}, 2018.

\bibitem[Ding et~al.(2023)Ding, Liu, He, Jiang, and Loy]{bench_mevis}
Henghui Ding, Chang Liu, Shuting He, Xudong Jiang, and Chen~Change Loy.
\newblock {MeViS}: A large-scale benchmark for video segmentation with motion expressions.
\newblock In \emph{ICCV}, 2023.

\bibitem[Yan et~al.(2024)Yan, Wang, Yan, Jiang, Hu, et~al.]{bench_revos}
Cilin Yan, Haochen Wang, Shilin Yan, Xiaolong Jiang, Yao Hu, et~al.
\newblock {VISA}: Reasoning video object segmentation via large language models.
\newblock \emph{arXiv preprint arXiv:2407.11325}, 2024.

\bibitem[Seo et~al.(2020)Seo, Lee, and Han]{bench_refytvos}
Seonguk Seo, Joon-Young Lee, and Bohyung Han.
\newblock {URVOS}: Unified referring video object segmentation network with a large-scale benchmark.
\newblock In \emph{ECCV}, 2020.

\bibitem[Mkhallati et~al.(2023)Mkhallati, Cioppa, Giancola, Ghanem, and Van~Droogenbroeck]{mkhallati2023soccernet}
Hassan Mkhallati, Anthony Cioppa, Silvio Giancola, Bernard Ghanem, and Marc Van~Droogenbroeck.
\newblock {SoccerNet-Caption}: Dense video captioning for soccer broadcasts commentaries.
\newblock In \emph{IEEE/CVF Conference on Computer Vision and Pattern Recognition Workshops (CVPRW)}, 2023.

\bibitem[Shen et~al.(2026{\natexlab{b}})Shen, Tian, Yang, and Liu]{shen2026simplestream}
Yujiao Shen, Shulin Tian, Jingkang Yang, and Ziwei Liu.
\newblock A simple baseline for streaming video understanding.
\newblock \emph{arXiv preprint arXiv:2604.02317}, 2026{\natexlab{b}}.

\bibitem[Zhang et~al.(2024{\natexlab{d}})Zhang, Wu, Li, Li, Ma, Liu, and Li]{model_llavavideo}
Yuanhan Zhang, Jinming Wu, Wei Li, Bo~Li, Zejun Ma, Ziwei Liu, and Chunyuan Li.
\newblock Video instruction tuning with synthetic data.
\newblock \emph{Transactions on Machine Learning Research (TMLR)}, 2024{\natexlab{d}}.
\newblock arXiv:2410.02713.

\bibitem[Yao et~al.(2025)Yao, Li, Wei, Li, Ren, et~al.]{model_timechatonline}
Linli Yao, Yicheng Li, Yuancheng Wei, Lei Li, Shuhuai Ren, et~al.
\newblock {TimeChat-Online}: 80\% visual tokens are naturally redundant in streaming videos.
\newblock \emph{arXiv preprint arXiv:2504.17343}, 2025.

\bibitem[Zeng et~al.(2025)Zeng, Qiu, Zhang, Li, Wang, et~al.]{model_streamforest}
Xiangyu Zeng, Kefan Qiu, Qingyu Zhang, Xinhao Li, Jing Wang, et~al.
\newblock {StreamForest}: Efficient online video understanding with persistent event memory.
\newblock \emph{arXiv preprint arXiv:2509.24871}, 2025.

\bibitem[Xia et~al.(2026)Xia, Chen, Zhang, Sun, and Zhou]{model_streamo}
Jiaer Xia, Peixian Chen, Mengdan Zhang, Xing Sun, and Kaiyang Zhou.
\newblock Streaming video instruction tuning.
\newblock \emph{arXiv preprint arXiv:2512.21334}, 2026.

\bibitem[Faure et~al.(2025)Faure, Yeh, Chen, Su, Lai, and Hsu]{model_hermes}
Gueter~Josmy Faure, Jia-Fong Yeh, Min-Hung Chen, Hung-Ting Su, Shang-Hong Lai, and Winston~H. Hsu.
\newblock {HERMES}: temporal-coherent long-form understanding with episodes and semantics.
\newblock \emph{arXiv preprint arXiv:2408.17443}, 2025.

\bibitem[Xiao et~al.(2021{\natexlab{b}})Xiao, Shang, Yao, and Chua]{xiao2021next}
Junbin Xiao, Xindi Shang, Angela Yao, and Tat-Seng Chua.
\newblock Next-qa: Next phase of question-answering to explaining temporal actions.
\newblock In \emph{Proceedings of the IEEE/CVF conference on computer vision and pattern recognition}, pages 9777--9786, 2021{\natexlab{b}}.

\bibitem[Lu et~al.(2024{\natexlab{b}})Lu, Bansal, Xia, Liu, Li, Hajishirzi, Cheng, Chang, Galley, and Gao]{bench_mathvista}
Pan Lu, Hritik Bansal, Tony Xia, Jiacheng Liu, Chunyuan Li, Hannaneh Hajishirzi, Hao Cheng, Kai-Wei Chang, Michel Galley, and Jianfeng Gao.
\newblock {MathVista}: Evaluating mathematical reasoning of foundation models in visual contexts.
\newblock In \emph{ICLR}, 2024{\natexlab{b}}.

\bibitem[Qiao et~al.(2024)Qiao, Tan, Dong, Wu, Sun, et~al.]{bench_wemath}
Runqi Qiao, Qiuna Tan, Guanting Dong, Minhui Wu, Chong Sun, et~al.
\newblock {We-Math}: Does your large multimodal model achieve human-like mathematical reasoning?
\newblock \emph{arXiv preprint arXiv:2407.01284}, 2024.

\bibitem[Wang et~al.(2024{\natexlab{e}})Wang, Pan, Shi, Lu, Zhan, and Li]{bench_mathvision}
Ke~Wang, Junting Pan, Weikang Shi, Zimu Lu, Mingjie Zhan, and Hongsheng Li.
\newblock Measuring multimodal mathematical reasoning with {MATH-Vision} dataset.
\newblock \emph{arXiv preprint arXiv:2402.14804}, 2024{\natexlab{e}}.

\bibitem[Zhang et~al.(2024{\natexlab{e}})Zhang, Jiang, Zhang, Lin, Guo, et~al.]{bench_mathverse}
Renrui Zhang, Dongzhi Jiang, Yichi Zhang, Haokun Lin, Ziyu Guo, et~al.
\newblock {MathVerse}: Does your multi-modal {LLM} truly see the diagrams in visual math problems?
\newblock In \emph{ECCV}, 2024{\natexlab{e}}.

\bibitem[Yue et~al.(2024)Yue, Ni, Zhang, Zheng, Liu, et~al.]{bench_mmmu}
Xiang Yue, Yuansheng Ni, Kai Zhang, Tianyu Zheng, Ruoqi Liu, et~al.
\newblock {MMMU}: A massive multi-discipline multimodal understanding and reasoning benchmark for expert {AGI}.
\newblock In \emph{CVPR}, 2024.

\bibitem[Zou et~al.(2025)Zou, Guo, Yang, Zhang, Hu, and Zhang]{bench_dynamath}
Chengke Zou, Xingang Guo, Rui Yang, Junyu Zhang, Bin Hu, and Huan Zhang.
\newblock {DynaMath}: A dynamic visual benchmark for evaluating mathematical reasoning robustness of vision language models.
\newblock In \emph{ICLR}, 2025.

\bibitem[Yue et~al.(2025)Yue, Zheng, Ni, Wang, Zhang, et~al.]{bench_mmmu_pro}
Xiang Yue, Tianyu Zheng, Yuansheng Ni, Yubo Wang, Kai Zhang, et~al.
\newblock {MMMU-Pro}: A more robust multi-discipline multimodal understanding benchmark.
\newblock In \emph{ACL}, 2025.

\bibitem[Zhang et~al.(2026{\natexlab{d}})Zhang, Wu, Yang, Li, Hu, Wang, Li, and Bing]{zhang2026openmmreasoner}
Kaichen Zhang, Keming Wu, Zuhao Yang, Bo~Li, Kairui Hu, Bin Wang, Xingxuan Li, and Lidong Bing.
\newblock Openmmreasoner: Pushing the frontiers in multimodal reasoning with an open and general recipe.
\newblock In \emph{Proceedings of the IEEE/CVF Conference on Computer Vision and Pattern Recognition}, pages 19276--19286, 2026{\natexlab{d}}.

\bibitem[{NVIDIA}(2025)]{nvidia_nemotron_pretraining_sft_v1}
{NVIDIA}.
\newblock {Nemotron-Pretraining-SFT-v1 Dataset}.
\newblock \url{https://huggingface.co/datasets/nvidia/Nemotron-Pretraining-SFT-v1}, 2025.

\end{thebibliography}
\bibliographystyle{unsrtnat}

\end{document}